\documentclass[letterpaper]{article} %
\usepackage{aaai2026}  %
\usepackage{times}  %
\usepackage{helvet}  %
\usepackage{courier}  %
\usepackage[hyphens]{url}  %
\usepackage{graphicx} %
\usepackage{natbib}  %
\usepackage{caption} %
\usepackage{comment}
\usepackage{algorithm}
\usepackage{algorithmic}

\usepackage{amsmath,amsfonts,bm}

\def\eqref#1{equation~\ref{#1}}

\def\1{\bm{1}}

\def\ve{{\bm{e}}}

\def\vx{{\bm{x}}}

\def\ve{{\bm{e}}}

\DeclareMathAlphabet{\mathsfit}{\encodingdefault}{\sfdefault}{m}{sl}
\SetMathAlphabet{\mathsfit}{bold}{\encodingdefault}{\sfdefault}{bx}{n}

\newcommand\remiicml{}
\newcommand\remiaaai{}
\newcommand\remirebutal{}
\newcommand\remi{}
\newcommand\eloise{}

\usepackage{enumitem}
\usepackage{booktabs}
\usepackage{makecell}
\usepackage{stmaryrd}
\usepackage{subcaption}
\usepackage{xcolor}

\usepackage{newfloat}
\usepackage{listings}
\DeclareCaptionStyle{ruled}{labelfont=normalfont,labelsep=colon,strut=off} %
\floatstyle{ruled}
\newfloat{listing}{tb}{lst}{}
\floatname{listing}{Listing}
\nocopyright

\title{Benchmarking XAI Explanations with Human-Aligned Evaluations}
\author{
    Rémi Kazmierczak\textsuperscript{\rm 1}, Steve Azzolin\textsuperscript{\rm 2}, Eloïse Berthier\textsuperscript{\rm 1}, Anna Hedström\textsuperscript{\rm 3}, Patricia Delhomme\textsuperscript{\rm 4}, David Filliat\textsuperscript{\rm 1}, Nicolas Bousquet\textsuperscript{\rm 5}, Goran Frehse\textsuperscript{\rm 1}, Massimiliano Mancini\textsuperscript{\rm 2}, Baptiste Caramiaux\textsuperscript{\rm 6}, Andrea Passerini\textsuperscript{\rm 2}, Gianni Franchi\textsuperscript{\rm 1}\\
}
\affiliations{
    \textsuperscript{\rm 1}Unité d'Informatique et d'Ingénierie des Systèmes, ENSTA Paris, Institut Polytechnique de Paris, Palaiseau, France\\
    \textsuperscript{\rm 2}Department of Information Engineering and Computer Science, University of Trento, Trento, Italy\\
    \textsuperscript{\rm 3}Understandable Machine Intelligence Lab, TU Berlin, Berlin, Germany\\
    \textsuperscript{\rm 4}Laboratory of Applied Ergonomics and Psychology, Université Gustave Eiffel, Versailles, France\\
    \textsuperscript{\rm 5}SINCLAIR Laboratory, Palaiseau, France\\
    \textsuperscript{\rm 6}Institute of Intelligent Systems and Robotics, Sorbonne Université, Paris, France\\
    \texttt{\{remi.kazmierczak, eloise.berthier, goran.frehse, gianni.franchi\}@ensta-paris.fr}\\
    \texttt{\{steve.azzolin, massimiliano.mancini, andrea.passerini\}@unitn.it}\\
    \texttt{hedstroem.anna@gmail.com}\\
    \texttt{patricia.delhomme@univ-eiffel.fr}\\
    \texttt{nicolas.bousquet@edf.fr}\\
    \texttt{baptiste.caramiaux@sorbonne-universite.fr}\\
}

\usepackage{bibentry}

\begin{document}

\maketitle

\begin{abstract}

We introduce PASTA (Perceptual Assessment System for explanaTion of Artificial Intelligence), a novel human-centric framework for evaluating eXplainable AI (XAI) techniques in computer vision. Our first contribution is the creation of the \textbf{PASTA-dataset}, the first large-scale benchmark that spans a diverse set of models and both saliency-based and concept-based explanation methods. This dataset enables robust, comparative analysis of XAI techniques based on human judgment. Our second contribution is an automated, data-driven benchmark that predicts human preferences using the \textbf{PASTA-dataset}. This scoring called \textbf{PASTA-score} offers scalable, reliable, and consistent evaluation aligned with human perception. Additionally, our benchmark allows for comparisons between explanations across different modalities, an aspect previously unaddressed. We then propose to apply our scoring method to probe the interpretability of existing models and to build more human-interpretable XAI methods.
\end{abstract}

\begin{figure}[t!]
    \centering
    \includegraphics[width=1\columnwidth]{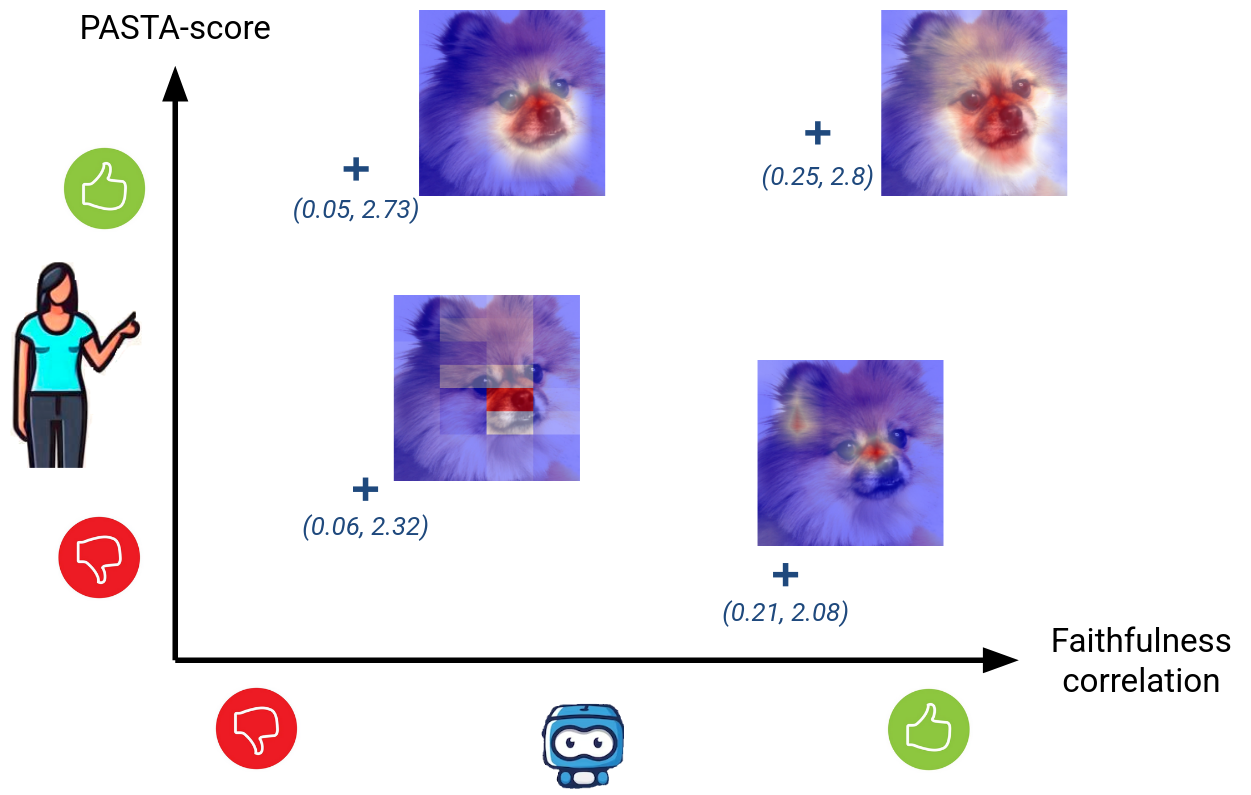}
\caption{\remiaaai{\textbf{PASTA automates the evaluation of human perception} of provided explanations by computing a PASTA-score. By integrating PASTA-score (y-axis) with existing faithfulness metrics (x-axis), we aim to foster the development of explanations that are not only aligned with the model's behavior but also comprehensible to human evaluators. Samples reported in the figure correspond to the label \textit{dog} for a ResNet50 classifier trained on PascalPART. XAI methods: top left: GradCAM; Top right: FullGrad; Bottom left: SHAP; Bottom right: AblationCAM}}
\label{fig:faith_comput}
\end{figure}
\section{Introduction}
As Deep Neural Networks (DNNs) are increasingly deployed in high-stakes domains such as law and medicine \citep{surden2021machine,litjens2017survey}, understanding their decision-making process has become essential \citep{bender2021dangers}. Their opacity often earns them the label ``black boxes'' \citep{castelvecchi2016can}, raising trust and accountability concerns in critical applications \citep{vereschak2024trust}. This has given rise to the field of eXplainable AI (XAI) \citep{gunning2019xai}.

A wide variety of XAI techniques have been proposed \citep{speith2022review,saeed2023explainable}, notably saliency-based methods \citep{muhammad2020eigen,bohle2024b}, which highlight relevant input features, and concept-based methods \citep{yan2023learning,diaz2022explainable}, which associate predictions with high-level semantic concepts. However, comparing such heterogeneous approaches remains an open problem.

Evaluating XAI methods is particularly challenging for two main reasons. First, the diversity of explanation types complicates the definition of a common evaluation framework. Second, the notion of a ``good explanation'' is inherently subjective. This creates a dichotomy between \emph{non-perceptual} evaluations—focused on model-centric metrics using toolkits like \texttt{OpenXAI}, \texttt{Quantus}, and \texttt{Xplique} \citep{agarwal2022openxai,hedstrom2023quantus,fel2022xplique}—and \emph{perceptual} evaluations, which assess human understanding. While the latter is often explored through anecdotal examples \citep{selvaraju2017grad,wang2020score}, user studies \citep{dawoud2023human,colin2022cannot}, or region-of-interest alignment \citep{liu2024human,arras2022clevr}, there is still a lack of standardization in assessing explanations from the human perspective \citep{nauta2023anecdotal}. 
\remiaaai{Yet, this dimension is crucial—explanations faithful to the model’s reasoning may still be unintelligible to human users, limiting their actionability. In this sense, our proposed PASTA-score allows for evaluating explanations based on a combination of faithfulness and human preferences, as shown in Figure~\ref{fig:faith_comput}.}

To address these challenges, we propose \textbf{PASTA}—the \textit{Perceptual Assessment System for explanaTion of Artificial intelligence}—which aims to automate the human-aligned evaluation of XAI methods.
PASTA has two core components. First, a benchmark, \textbf{PASTA-dataset}, composed of four diverse image-based \remiaaai{classification datasets} with aligned concept annotations, enabling the comparison of \remiaaai{20} XAI techniques across multiple architectures. Second, the \textbf{PASTA-score}, a data-driven metric designed to predict human preferences, providing a scalable way to evaluate explanations from a perceptual standpoint. Unlike prior benchmarks focused solely on saliency or user studies \citep{colin2022cannot,dawoud2023human}, PASTA unifies both saliency-based and concept-based methods under a single evaluation framework.

\begin{figure}[t]
    \centering
    \includegraphics[width=0.9\columnwidth]{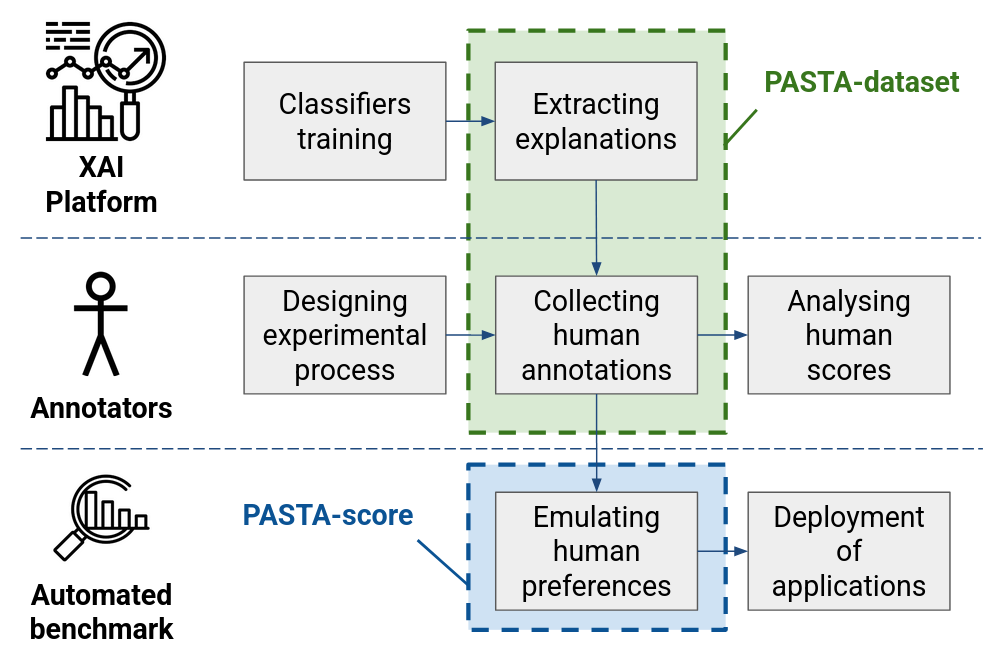}
    \caption{\remiicml{
    \textbf{Pipeline of the PASTA framework.} We first collect a dataset of human preferences called PASTA-dataset. This dataset is used to learn to emulate human preferences for the new samples on the test set using the PASTA-score. The trained PASTA-score can then be deployed to downstream applications as a consistent-over-time, quick and cost-effective replacement for human feedback.
    }}
    \label{fig:Pipeline_paper}
\end{figure}

Our contributions are:
 \textbf{(1) Comprehensive XAI Benchmark}: We introduce the PASTA-dataset, enabling the evaluation of both visual and concept-based explanations.
 \textbf{(2) Large-scale Method Evaluation}: We assess 20 XAI methods across multiple datasets and models, including both post-hoc and ante-hoc methods. Our first result suggests that human annotators tend to prefer saliency and perturbation-based techniques, like LIME and SHAP.
 \textbf{(3) Human-aligned Explanation Scoring}: We propose \textbf{PASTA-score}, an automated yet perceptually grounded assessment of explanations, trained on the PASTA-dataset to replicate human preferences.
 \textbf{(4) Practical Applications}: We present three use cases of PASTA-score, showing how it can guide the design of more interpretable models, and serve as a proxy for visual human assessment in practical deployments.
\remiicml{The pipeline of the global workflow is presented in Figure~\ref{fig:Pipeline_paper}.}

\remiaaai{The complete PASTA framework (code, annotation, and models) will be released upon acceptance.}

\section{Related Work}\label{sec:related_work}
\paragraph{\remi{Explainable AI.}}
\remirebutal{
To address the challenge of explaining DNNs, several specialized tools have been proposed, often categorized into \textit{post-hoc} and \textit{ante-hoc} methods \citep{arrieta2020explainable, rudin2022interpretable}.
\textit{Post-hoc} methods encompass any tool external to the model, allowing us to gain insights from any pre-trained DNN. Popular examples are GradCAM \citep{selvaraju2017grad}, LIME \citep{lime}, and SHAP \citep{lundberg2017unified}. While most post-hoc explainers agree in providing input regions most responsible for a certain prediction, they differ in many non-trivial details, and selecting and evaluating the most appropriate explainer for each task can be challenging \citep{leavitt2020towards, roy2022don}.
\textit{Ante-hoc} methods, instead, aim at modifying the underlying model architecture to provide explanations by design.
This can be done in the framework of Concept Bottleneck Models (CBMs) \citep{koh2020concept} by prompting the model to first predict a set of human-understandable high-level concepts, and then making the final prediction using a shallow and interpretable classifier that supports human inspection, or by decomposing the \textit{reasoning} of the model into smaller and more actionable steps \citep{ge2023chain}.}
\remi{
\paragraph{\remi{Evaluating explainability.}}
\remirebutal{
While several methods have been proposed to quantitatively measure explanation quality, such as faithfulness \citep{petsiuk2018rise, dasgupta2022framework, azzolin2024perks}, sparsity \citep{chalasani2020concise, benard2021interpretable}, robustness \citep{alvarezmelis2018robustnessinterpretabilitymethods, montavon2018methods}, sensitivity \citep{adebayo2018, hedstrom2024sanity} and alignment to an assumed ground truth \citep{colin2022cannot,mohseni2021quantitative,dawoud2023human}, they inherently overlook the perceptual aspect with respect to the human, who is the expected consumer of such explanations.
Evaluating explanations via user studies, e.g., where annotators are asked to rate and evaluate explanations \citep{chen2018neural,shu2019defend,yang2022hsi,kares2025makes}, are, however, very costly, prone to unreproducibility issues \citep{nauta2023anecdotal}, and often unfeasible for tasks that require trained users, like in the medical domain \citep{miro2022evaluating,muddamsetty2021expert}.
In this work, we take the first step towards standardizing the evaluation of human perception preferences of explanations \citep{nauta2023anecdotal}. We propose to overcome the issues of hard-to-reproduce large-scale user studies by automating the evaluation of XAI techniques through a multi-value scoring method that mimics human preferences while taking into account the users' diverse expectations, which naturally emerge in user-based studies.
}
\paragraph{Automated scoring.}
Automated scoring involves developing models that assign scores to inputs based on a reference dataset, often derived from human ratings. A particularly active area of research in this domain is automated essay scoring. Traditionally, this has been addressed through handcrafted feature extraction \citep{yannakoudakis2011new}, but modern methods tend to be closer to model as a judge \citep{lee2024prometheusvision,taghipour2016neural, chiang2024chatbot}. More recently, there has been a growing interest in using embeddings from large language models (LLMs) as features for scoring. The first successful attempt in this direction was made by \citet{yang2020enhancing}. Building on this trend, other approaches have incorporated LLM embeddings with models like LSTMs \citep{wang2022use}, integrated text generation into the training loop \citep{xiao2024automation}, or introduced multi-scale aspects to enhance performance \citep{li2023automatic}.
\section{\remi{Creating the PASTA-dataset}} \label{dataset}

\remiaaai{To assess the quality of XAI explanations for image classification decisions from a human-centric perspective, we constructed a comprehensive dataset comprising images, predictions, explanations, and evaluations of these explanations, as depicted in Figure \ref{fig:Pipeline_benchmark}. 
To account for the heterogeneity of different XAI methods, model backbones, and training scenarios, we constructed the PASTA-benchmark to include 1000 images sampled across 4 available datasets, 7 classification backbones, 20 XAI methods, 6 questions, and 5 annotations per image.
The challenge in annotating such a dataset resides in its multiplicative nature, where each question requires an annotation across multiple backbones, datasets, XAI methods, images, and human annotators.
Consequently, the PASTA-dataset contains an overall number of 633,000 samples, each corresponding to a unique Likert-like rating, which is the largest benchmark available of this kind.}

\remiaaai{To construct such a dataset, the initial phase involved developing a unified platform designed to integrate various models, XAI methods, and datasets in a streamlined manner. This platform encompasses a diverse array of models and XAI methods. Given the potential utility of this platform as a baseline for future research, we intend to release it publicly upon publication of this paper. Further details regarding the overall procedure, including details about the datasets employed, model training, and explanation extraction, are available in Section A.1 of the appendix.}
\begin{figure}[t]
    \centering
    \includegraphics[width=1\columnwidth]{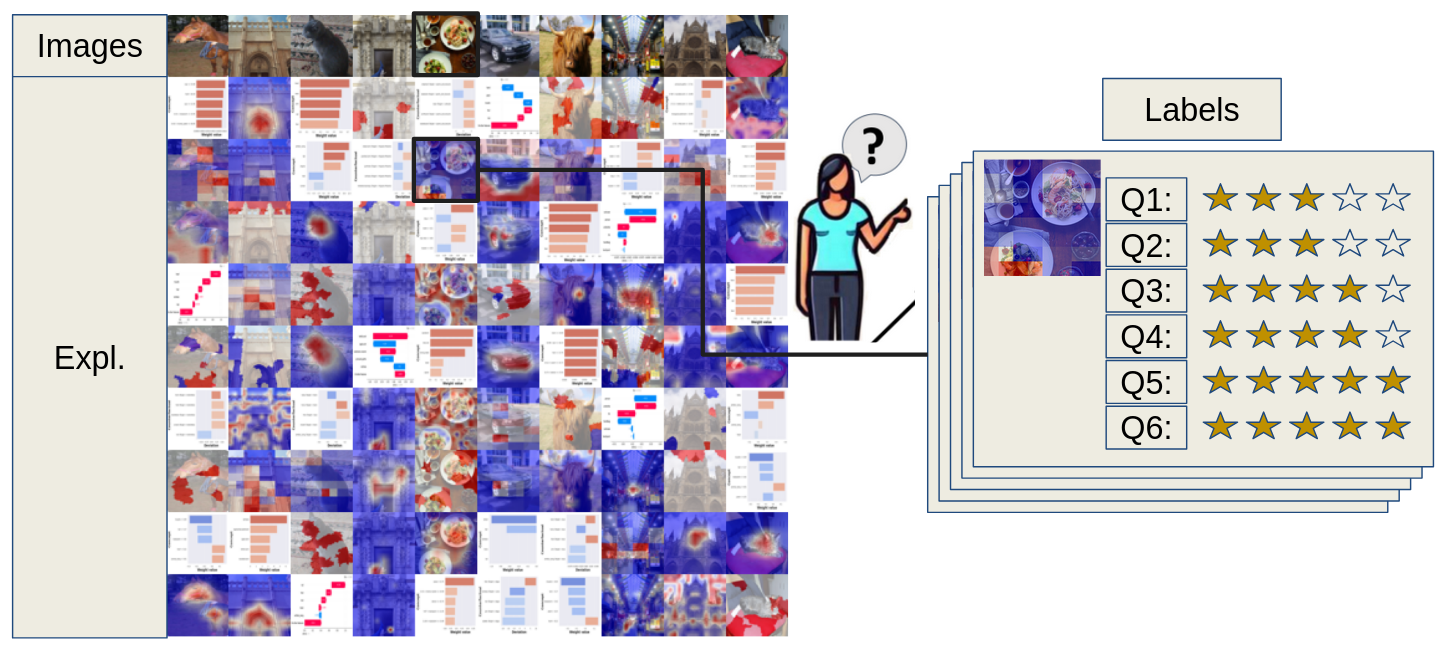}
    \caption{\textbf{Overview of the human annotation process in the PASTA-dataset.} %
    We compute a total number of 46 explanations for each image, out of which 21 are sampled and rated by humans according to six questions. Further details available in Appendix A.2.
    }
    \label{fig:Pipeline_benchmark}
\end{figure}
\remiaaai{The subsequent phase in dataset creation is dedicated to annotating the explanations, which involved 24 participants in an online process. A pivotal insight from existing user study literature \citep{xuan2025comprehension,liao2022connecting} is that human perception of explanations is not unidimensional; rather, it encompasses a range of potentially unaligned desiderata. For instance, a saliency-based explanation that highlights a dog to predict a cat can be entirely clear, thereby satisfying complexity desiderata, while simultaneously not fulfilling plausibility desiderata. Consequently, we pose multiple questions designed to encompass a spectrum of human assessment as broad as possible. Specifically, the following questions were asked:}
\remiaaai{\begin{itemize}
    \item \remi{Q1}: \remi{Is the provided explanation consistent with how I would explain the predicted class?}
    \item \remi{Q2}: \remi{Overall the explanation provided for the model prediction can be trusted?}
    \item \remi{Q3}: Is the explanation \remi{easy to understand}?
    \item \remi{Q4}: \remi{Can the explanation be understood by a large number of people, independently of their demographics (age, gender, country, etc.) and culture?}
    \item \remi{Q5}: With this perturbed image, \remi{to what extent has the explanation changed ?} (Examples with good predictions \remi{and light perturbations}) 
    \item \remi{Q6}: With this perturbed image, \remi{to what extent has the explanation changed?} (Examples with bad predictions \remi{and strong perturbations})
\end{itemize}}
\remiaaai{The selection of these questions reflects a broader discourse on user studies and desiderata. To ensure that annotators comprehensively understand the task and expectations, we implemented an evaluation protocol developed with the assistance of a psychologist. This protocol includes annotator training and continuous monitoring throughout the process. Discussions regarding the desiderata, a detailed evaluation protocol, and information about the annotators are available in Section A.2 of the Appendix.}

\section{Analysis of Human Preference} \label{sec:ResultsEval}
\begin{figure}[t!]
    \centering
    \includegraphics[width=\columnwidth, trim=0 35 0 0, clip]{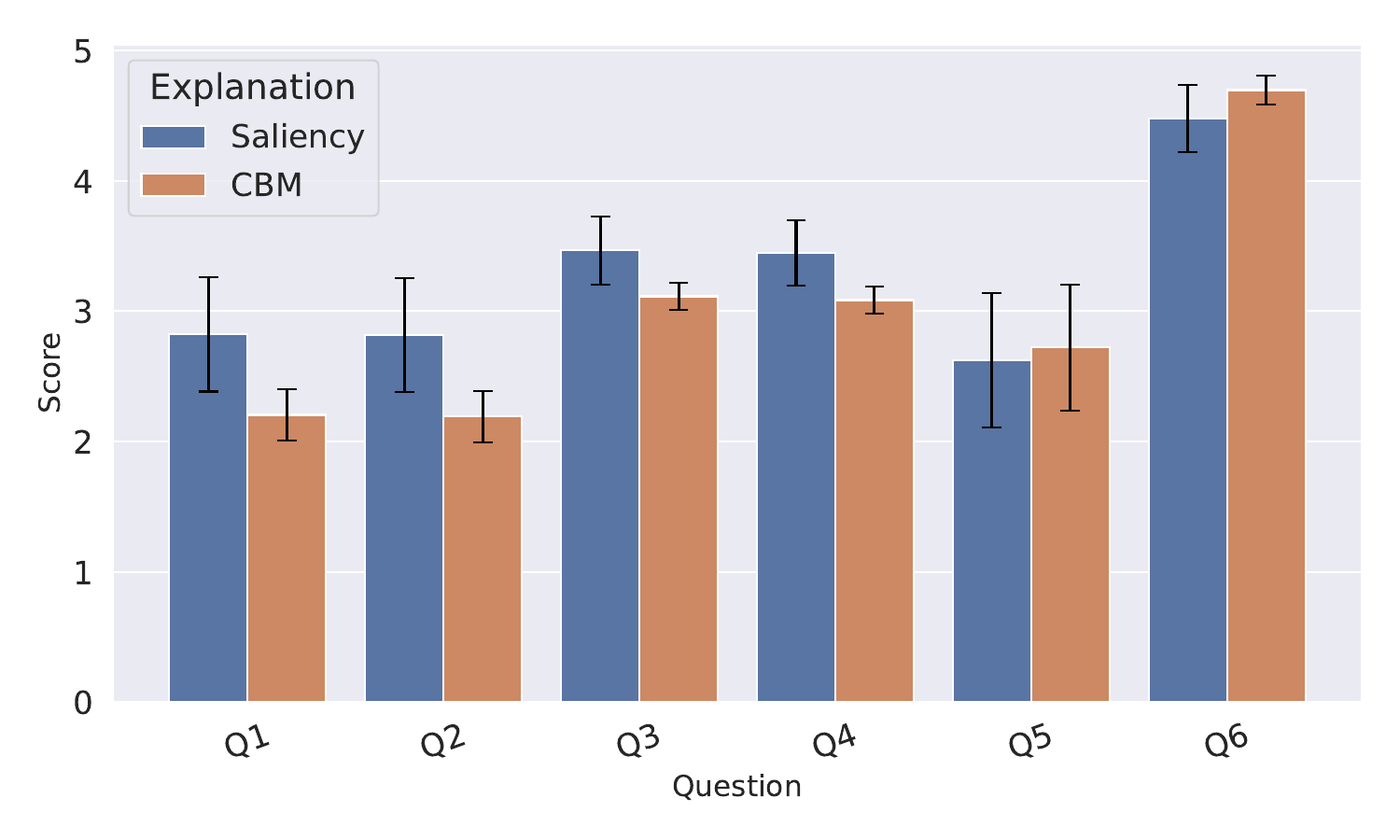}
    \caption{\remiaaai{\textbf{Scores for each question, for saliency-based and CBM-based explanation.} Overall, saliency-based explanations are preferred over CBM-based ones.}}
    \label{fig:saliency_vs_cbm}
\end{figure}
\begin{figure}[t!]
    \centering
    \includegraphics[width=\columnwidth, trim=0 35 0 0, clip]{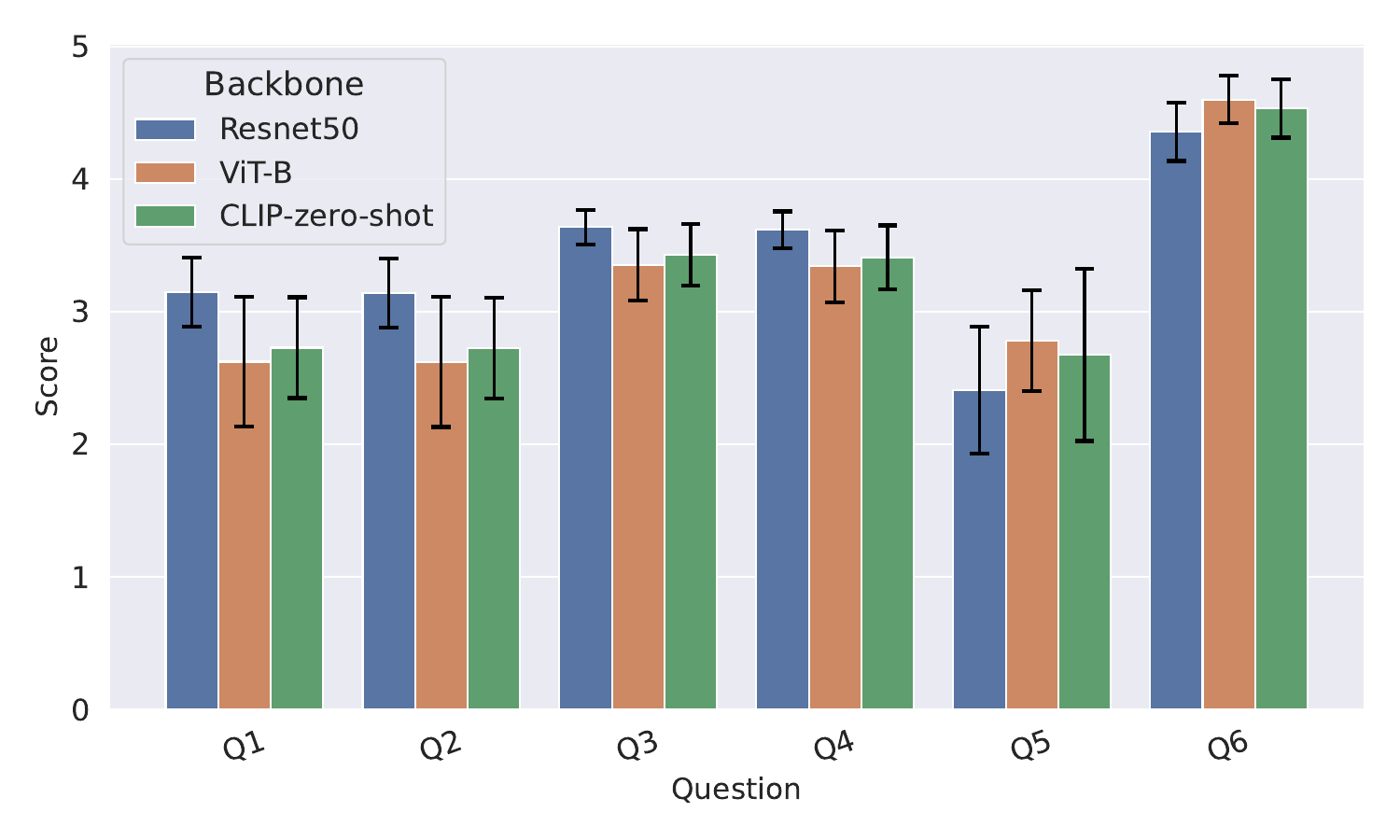}
    \caption{\remiaaai{\textbf{Scores for each question, for different backbones of saliency methods.} As backbones for saliency methods, ViT-B and CLIP obtain overall similar results, while ResNet50 has generally better scores.}}
    \label{fig:resnet_vs_vit_vs_clip}
\end{figure}
Having collected a large number of human preferences for different XAI models and backbones in the PASTA-dataset, we now proceed to analyze human scores in relation to each method. The full analysis is available in Appendix B.3.

\textbf{Human preferences for \remiaaai{output format}:}
As illustrated in Figure \ref{fig:saliency_vs_cbm}, results indicate that humans tend to prefer image-based explainers in relation to questions Q1-Q5, meaning that saliency maps are perceived as more interpretable than concept-based explanations. 
Although several factors may contribute to this behavior—such as the lower cognitive load of image-based explanations compared to concept-based ones—a comprehensive investigation of the underlying psychological causes is left for future work.
\remiaaai{The sole exception to this observation pertains to Q6 (note that for Q5, the ratings are inverted, with a low score indicating favorable behavior). This phenomenon can be explained by the fact that presenting explanations as a heatmap overlaid on the image reduces the perceptual impact of perturbations.}

\textbf{Human preferences for model architecture:}
\remiaaai{Figure \ref{fig:resnet_vs_vit_vs_clip} illustrates the average scores across XAI methods that use the same backbone, highlighting a general preference for ResNet50.}
CLIP and ViT achieve similar scores, likely due to the architectural similarities between the two models. ResNet50, which played a pivotal role in the development of many XAI methods, consistently scores higher. This could suggest a potential bias toward ResNet50 in the design and effectiveness of current XAI methods. \remiaaai{The results for the last two questions may be due to ViT being more sensitive to the perturbations used than Resnet50.}
\remiaaai{Among the methods we studied, those based on feature factorization—like EigenCAM~\citep{muhammad2020eigen} and Deep Feature Factorization~\citep{collins2018deep}—tend to give more consistent and preferred explanations. This may be because they remove complex components that can create confusing artifacts, making the explanations easier to understand.}
\section{Developing \remiicml{the PASTA-score}}\label{sec:PASTA-score}
\remi{To provide a tool for measuring human assessment of XAI techniques, we introduce the PASTA-score, which simulates a human evaluation.}
\remi{The global pipeline is illustrated in Figure \ref{fig:Pipeline_PASTA}. More precisely, the PASTA-score is composed of an embedding network, that processes both CBM outputs or saliency maps, and a scoring network, that computes scores from the embeddings.} \eloise{Using the data collected in \remiaaai{the PASTA-dataset}, the PASTA-score aims at predicting the human scores for questions \remirebutal{Q1 to Q6 }\remiaaai{for new explanations, playing the role of an automated benchmark}. 
\begin{figure}[t]
    \centering
    \includegraphics[width=1\columnwidth]{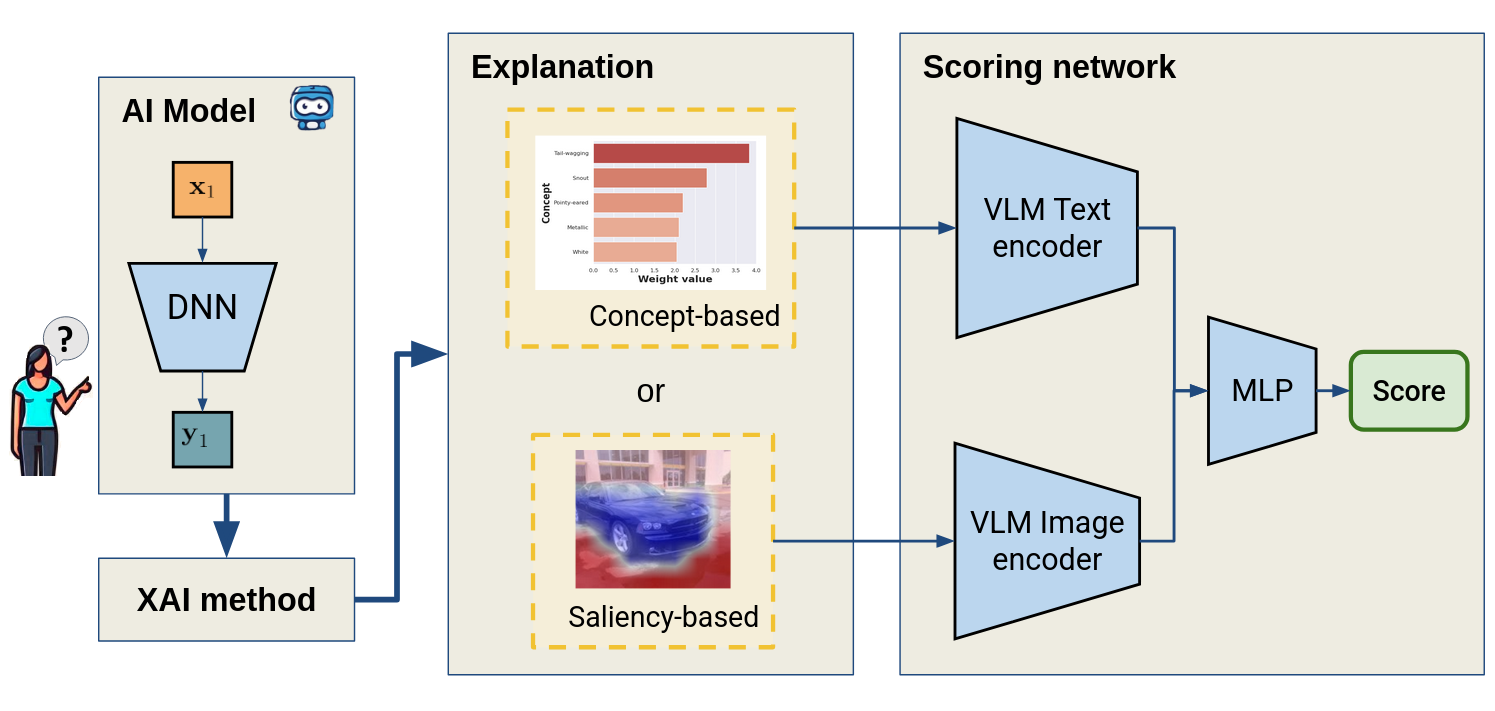}
    \caption{
    \textbf{Functioning of the PASTA-score.} First, we extract the embeddings for each explanation using a frozen image encoder of a VLM. Then, we employ a scoring network trained on the labels provided by the PASTA-dataset to generate a final score.
    }
    \label{fig:Pipeline_PASTA}
\end{figure}

\subsection{\remi{Computation of the embeddings}} \label{embeding_net}
\remi{Drawing inspiration from recent literature in automated \remiaaai{essay} scoring~\citep{yang2020enhancing,wang2022use}, \remiicml{which encounters similar challenges due to working with a dataset of \remiaaai{thousands of samples} \remiaaai{(21,110 samples per question, precisely)} and requiring a DNN to automatically learn the score, we opt for a \remiaaai{foundation model} %
that we will fine-tune using a multi-linear perceptron.} %
\remiaaai{However, unlike automated essay scoring, we deal with both image and textual inputs, making the use of a Vision Language Model (VLM) mandatory. We tested multiple candidates, such as CLIP \citep{yan2023learning}, BLIP \citep{li2022blip} and LLaVa \citep{liu2024visual}.}
This choice allows for a unified integration of both concept-based explanations, which can be transformed into text, and saliency map-based explanations, which can be projected into the same embedding space.} %
Let $\vx_i \in \mathbb{R}^{H \times W \times 3}$ be the $i$-th test image with height $H$ and width $W$, and let $\ve_i^{saliency} \in \mathbb{R}^{H \times W}$ be a saliency-based explanation for this image.
We denote the image encoder of a Vision-Language Model (VLM) as $\text{VLM}_{\text{image}}$. To embed a saliency explanation, we apply the encoder to the image overlaid with its heatmap:
\begin{equation}
    \phi_{\text{image}}(\ve_i^{saliency}) = \text{VLM}_{\text{image}}(Heatmap(\vx_i,\ve_i^{saliency})),
    \label{embed_saliency_bis}
\end{equation}
where $Heatmap$ generates the visual overlay of the explanation on the image.

For concept-based methods (CBMs), let $\ve_i^{CBM} \in \mathbb{R}^K$ be the explanation vector, where $K$ is the number of concepts. This vector is turned into a sentence using a text template, and then encoded with the VLM's text encoder $\text{VLM}_{\text{text}}$:
\begin{equation}
    \phi_{\text{text}}(\ve_i^{CBM}) = \text{VLM}_{\text{text}}(Sentence(\ve_i^{CBM})).
    \label{embed_cbm}
\end{equation}
Since the PASTA-score is compatible with different VLMs and does not rely on a specific one, we evaluate it using several VLMs: CLIP~\citep{liu2024visual}, SIGLIP~\citep{zhai2023sigmoid}, EVA~\citep{sun2023eva}, and BLIP~\citep{li2022blip}. This results in multiple variants of our metric: $\text{PASTA-score}^{\text{CLIP}}$, $\text{PASTA-score}^{\text{SIGLIP}}$, $\text{PASTA-score}^{\text{EVA}}$, and $\text{PASTA-score}^{\text{BLIP}}$.

To support our design choices, Appendix~D.2 presents extensive ablations on various factors: the impact of textual templates, the number of concepts $K$, the way saliency maps are visualized, and whether label information is included.
We also explore alternative versions of Equations~\ref{embed_saliency_bis} and~\ref{embed_cbm}, and how these choices affect the final score.

\subsection{\remi{Scoring network}} \label{scoring_net}
\remi{Once the embeddings are computed, \remiaaai{the label information is concatenated to the embedding, and} a scoring network composed of a \remiaaai{multi-layer perceptron} is used to predict scores. Inspired by  Automated Essay Scoring ~\citep{yang2020enhancing,wang2022use}, we use a loss~$L$ that combines a similarity loss~$L_s$, a mean squared error (MSE) loss $L_{mse}$, and a ranking loss~$L_r$. From a set of ground truth scores obtained from majority voting $\{m_k\}_{k \in [0, N_s]}$ and the predictions given by the scoring network $\{\hat{m}_k\}_{k \in [0, N_s]}$, the resulting loss is defined as:}
\remiaaai{
\begin{align}
    L(m_k,\hat{m}_k) = \alpha L_s(m_k,\hat{m}_k) + \beta L_{mse}(m_k,\hat{m}_k) \nonumber \\ + \gamma L_r(m_k,\hat{m}_k),
\end{align}}%
\remi{where $\alpha$, $\beta$, and $\gamma$ are hyperparameters controlling the relative importance of each component. \remirebutal{Formulas} about the different losses are given in Appendix D.1.  %
\remirebutal{Since the PASTA dataset provides 5 ground-truth votes per inference, we explored different aggregation strategies. To mitigate the phenomenon of high non-consensus, the mode was selected as the final choice.}}

\subsection{\remi{Classifier results}}

Note that in the PASTA-dataset each sample corresponds to a triplet (input image, explanation, human ratings). The same image thus appears multiple times for different XAI methods, and the same XAI method appears multiple times for different images. To guarantee that no leakage occurs between train-test splits, we design them to ensure that the same image, or the same XAI method, does not appear in different splits. Images and XAI methods included in the training splits are randomly chosen based on the run's random seed.
\remi{For Q1 to Q6, we calculate the Mean Square Error (MSE), Quadratic Weighted Kappa (QWK), and Spearman Correlation Coefficient (SCC) between the predicted and ground truth labels on the test set. The results are presented in Table~\ref{tab:results_pasta}, \remiicml{where we also ablate different choices of embedding methods. We also report the inter-annotator agreement values, which correspond to the expected deviation of the metrics between a randomly selected annotator's score and the mode.} 
\remiaaai{Our network best replicates answers to Q1 and Q2, with similar performance across $\text{PASTA-score}^{\text{CLIP}}$ and $\text{PASTA-score}^{\text{SIGLIP}}$. This is likely due to greater rating diversity and stronger agreement between annotators, which supports more stable training. In contrast, Q3 to Q5 shows lower agreement, and Q5–Q6 has less diverse ratings. While the MSE stays similar, it becomes harder to learn the ranking patterns, likely due to the more subjective nature of these questions and the added uncertainty from image perturbations in Q5 and Q6.}
\begin{table*}[h]
\centering
\caption{\remirebutal{\textbf{Mean Square Error (MSE), Quadratic Weighted Kappa (QWK), and Spearman Correlation Coefficient (SCC) for each question.} Each value is the average of 5 runs with standard deviation. \textit{Human} refers to inter-annotator agreement.}}
\label{tab:results_pasta}
\scalebox{0.85}{
\remiaaai{
\begin{tabular}{@{}lccccc|cc@{}}
\toprule
\textbf{Metric}            & \textbf{Model}        & \textbf{Q1}                  & \textbf{Q2}                  & \textbf{Q3}                  & \textbf{Q4}                  & \textbf{Q5}                  & \textbf{Q6}                  \\
\midrule
MSE $\downarrow$     & \( \text{PASTA-score}^{\text{CLIP}} \) & 0.990 $\pm$ 0.104 & \textbf{0.993 $\pm$ 0.096} & 2.111 $\pm$ 2.529 & \textbf{0.811 $\pm$ 0.095} & 1.476 $\pm$ 0.183 & 0.752 $\pm$ 0.127 \\
                           & \( \text{PASTA-score}^{\text{SIGLIP}} \) & \textbf{0.989 $\pm$ 0.113} & 1.009 $\pm$ 0.125 & \textbf{0.842 $\pm$ 0.094} & 0.840 $\pm$ 0.106 & \textbf{1.396 $\pm$ 0.177} & \textbf{0.739 $\pm$ 0.140} \\
                           & \( \text{PASTA-score}^{\text{BLIP}} \) & 3.297 $\pm$ 1.840 & 3.287 $\pm$ 1.835 & 5.938 $\pm$ 2.542 & 4.642 $\pm$ 3.135 & 3.005 $\pm$ 1.385 & 10.710 $\pm$ 4.943 \\
                           & \( \text{PASTA-score}^{\text{EVA}} \) & 1.666 $\pm$ 1.215 & 1.747 $\pm$ 1.174 & 3.355 $\pm$ 3.099 & 2.097 $\pm$ 2.608 & 1.767 $\pm$ 0.568 & 3.324 $\pm$ 5.091 \\
                           & \textit{Human}      & 0.415 $\pm$ 0.037  & 0.429 $\pm$ 0.049  & 0.562 $\pm$ 0.104  & 0.478 $\pm$ 0.080  & 0.509 $\pm$ 0.102  & 0.299 $\pm$ 0.051  \\
\midrule
QWK $\uparrow$      & \( \text{PASTA-score}^{\text{CLIP}} \) & 0.450 $\pm$ 0.066 & 0.452 $\pm$ 0.063 & 0.199 $\pm$ 0.040 & 0.216 $\pm$ 0.052 & 0.165 $\pm$ 0.060 & 0.159 $\pm$ 0.031 \\
                           & \( \text{PASTA-score}^{\text{SIGLIP}} \) & \textbf{0.471 $\pm$ 0.055} & \textbf{0.459 $\pm$ 0.056} & \textbf{0.237 $\pm$ 0.052} & 0.219 $\pm$ 0.035 & \textbf{0.177 $\pm$ 0.061} & 0.165 $\pm$ 0.018 \\
                           & \( \text{PASTA-score}^{\text{BLIP}} \) & 0.328 $\pm$ 0.023 & 0.340 $\pm$ 0.020 & 0.181 $\pm$ 0.003 & 0.173 $\pm$ 0.017 & 0.081 $\pm$ 0.081 & 0.159 $\pm$ 0.011 \\
                           & \( \text{PASTA-score}^{\text{EVA}} \) & 0.462 $\pm$ 0.050 & 0.457 $\pm$ 0.054 & 0.160 $\pm$ 0.099 & \textbf{0.230 $\pm$ 0.018} & 0.163 $\pm$ 0.029 & \textbf{0.185 $\pm$ 0.049} \\
                           & \textit{Human}      & 0.849 $\pm$ 0.013  & 0.845 $\pm$ 0.017  & 0.731 $\pm$ 0.050  & 0.748 $\pm$ 0.041  & 0.848 $\pm$ 0.029  & 0.796 $\pm$ 0.048  \\
\midrule
SCC $\uparrow$     & \( \text{PASTA-score}^{\text{CLIP}} \) & 0.484 $\pm$ 0.064 & 0.484 $\pm$ 0.062 & 0.213 $\pm$ 0.040 & 0.230 $\pm$ 0.050 & 0.197 $\pm$ 0.073 & 0.193 $\pm$ 0.030 \\
                           & \( \text{PASTA-score}^{\text{SIGLIP}} \) & \textbf{0.501 $\pm$ 0.052} & \textbf{0.490 $\pm$ 0.057} & \textbf{0.247 $\pm$ 0.048} & 0.223 $\pm$ 0.029 & 0.213 $\pm$ 0.071 & 0.196 $\pm$ 0.020 \\
                           & \( \text{PASTA-score}^{\text{BLIP}} \) & 0.397 $\pm$ 0.036 & 0.411 $\pm$ 0.033 & 0.207 $\pm$ 0.065 & 0.194 $\pm$ 0.014 & 0.088 $\pm$ 0.104 & 0.218 $\pm$ 0.012 \\
                           & \( \text{PASTA-score}^{\text{EVA}} \) & 0.484 $\pm$ 0.046 & 0.474 $\pm$ 0.048 & 0.150 $\pm$ 0.155 & \textbf{0.247 $\pm$ 0.015} & \textbf{0.216 $\pm$ 0.035} & \textbf{0.220 $\pm$ 0.057} \\
                           & \textit{Human}      & 0.844 $\pm$ 0.017  & 0.839 $\pm$ 0.019  & 0.722 $\pm$ 0.045  & 0.742 $\pm$ 0.038  & 0.850 $\pm$ 0.023  & 0.789 $\pm$ 0.039  \\
\bottomrule
\end{tabular}}
}
\end{table*}
\subsection{\remiaaai{Alignment with Established Benchmarks}}

\remiicml{In this subsection, we aim to evaluate whether the PASTA-score aligns with the findings of existing studies that compare human assessments of XAI methods. To accomplish this, we use the benchmark dataset established by \citet{yang2022hsi}, which offers comparative evaluations of GradCAM \citep{selvaraju2017grad}, RISE \citep{petsiuk2018rise} (at the image level), and Guided Backpropagation \citep{springenberg2014striving}. These quantitative evaluations are presented in the form of Mean Absolute Error (MAE) between generated saliency maps and those produced by a Human Saliency Imitator (HSI). Given that this benchmark aims to explore human expectations, we compare the results obtained by the authors (MAE) with the PASTA-score related to Q1.}

\begin{table}[h]
    \centering
    \footnotesize
    \caption{\remiaaai{\textbf{Comparison of Rankings Based on HSI and PASTA-Scores.} The results illustrate that the PASTA-score exhibits a correlation with the Human Saliency Imitator presented by \citep{yang2022hsi}.}}
    \setlength{\tabcolsep}{6pt}
        \remiaaai{\begin{tabular}{@{}lcc@{}}
            \toprule
            \textbf{Method} & \textbf{MAE Score} $\downarrow$ & \textbf{PASTA-score $\uparrow$} \\
            \midrule
            RISE & 0.442  & 3.113 \\
            GradCAM & 0.703  & 3.150 \\
            Guided Backpropagation     & 0.890   & 1.846 \\
            \bottomrule
        \end{tabular}}
    \label{tab:mae_pasta_comparison}
\end{table}
\remiaaai{The results are summarized in Table \ref{tab:mae_pasta_comparison}, and indicate that despite RISE and Guided Backpropagation not being included during the training of the PASTA metric, the obtained rankings remain consistent. This provides empirical evidence supporting the generalization of the PASTA-score to unseen XAI techniques.}

\section{\remiaaai{Applications}}

\remiaaai{
In this section, we explore three different applications using the PASTA-score as a replacement for human feedback, which would be difficult or too costly to run at scale without automation
We use PASTA-score to guide XAI methods toward better interpretability (in the first and third applications) and to analyze how model size affects interpretability (in the second application). All experiments use the PASTA-score model trained on Q1 for consistency.}

\subsection{\remiaaai{Mixture of XAI methods}}

Our first application is to dynamically select the explainer giving the explanation that best matches human judgments for each specific image, using a mixture of XAI methods. In our experiments, we fix the classifier to be a ResNet50, and we select the explanation with the highest PASTA-score for each image.  The distribution of selected XAI methods is shown in Table \ref{fig:dist-XAI-mixture}. 
\begin{figure}
    \centering
    \includegraphics[width=\linewidth,trim=0 40 0 0, clip]{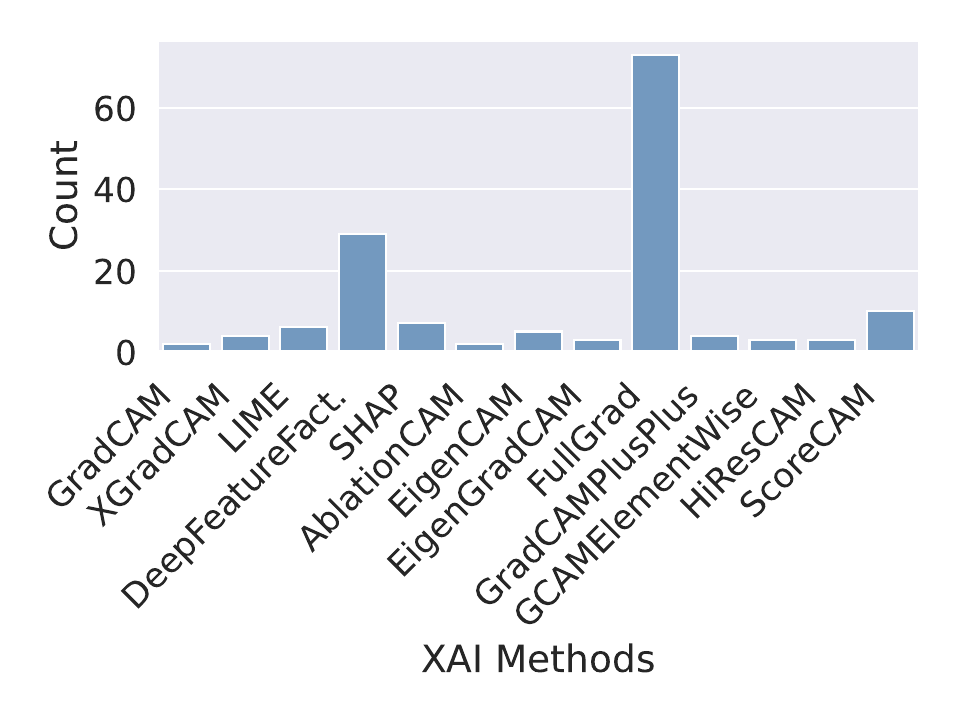}    \caption{\remiaaai{\textbf{Distribution of methods selected by our mixture of XAI techniques.} The x-axis denotes the number of images in the automated benchmark for which the respective XAI method attained the highest performance.}}
    \label{fig:dist-XAI-mixture}
\end{figure}
\remiaaai{The results indicate a substantial diversity in the methods employed, with FullGrad emerging as the most frequently used, selected nearly half of the time. This trend is reflective of user ratings within the PASTA-dataset, where FullGrad is identified as providing the most effective explanations according to annotators. In terms of faithfulness, the computation of the average faithfulness correlation across explanations selected by our PASTA-score yields a relatively stable value, with a slight improvement compared to the value obtained by averaging over every explainer (0.0627 for our selection versus 0.0579 for the average over every explainer). 
This confirms that it is possible to use PASTA to enhance the interpretability of explanations without compromising their faithfulness.}

\subsection{\remiaaai{Backbone size influence on the understanding of explanation}}

\remiaaai{
We use the PASTA-score to investigate whether model size influences the human perception of explanations, and how this relates to XAI methods.
To this end, we compute the average PASTA-score within an identical experimental framework, varying only the size of the backbone model. Specifically, we employ CLIP as the classifier and select backbones from among its ViT-B-16, ViT-L-14, ViT-H-14, and ViT-g-14 variants. The results of this analysis are presented in Figure~\ref{fig:exp_backbone_impact}.
}
\begin{figure*}[h]
    \begin{subfigure}[c]{0.45\linewidth}
        \centering        \includegraphics[width=0.99\linewidth]{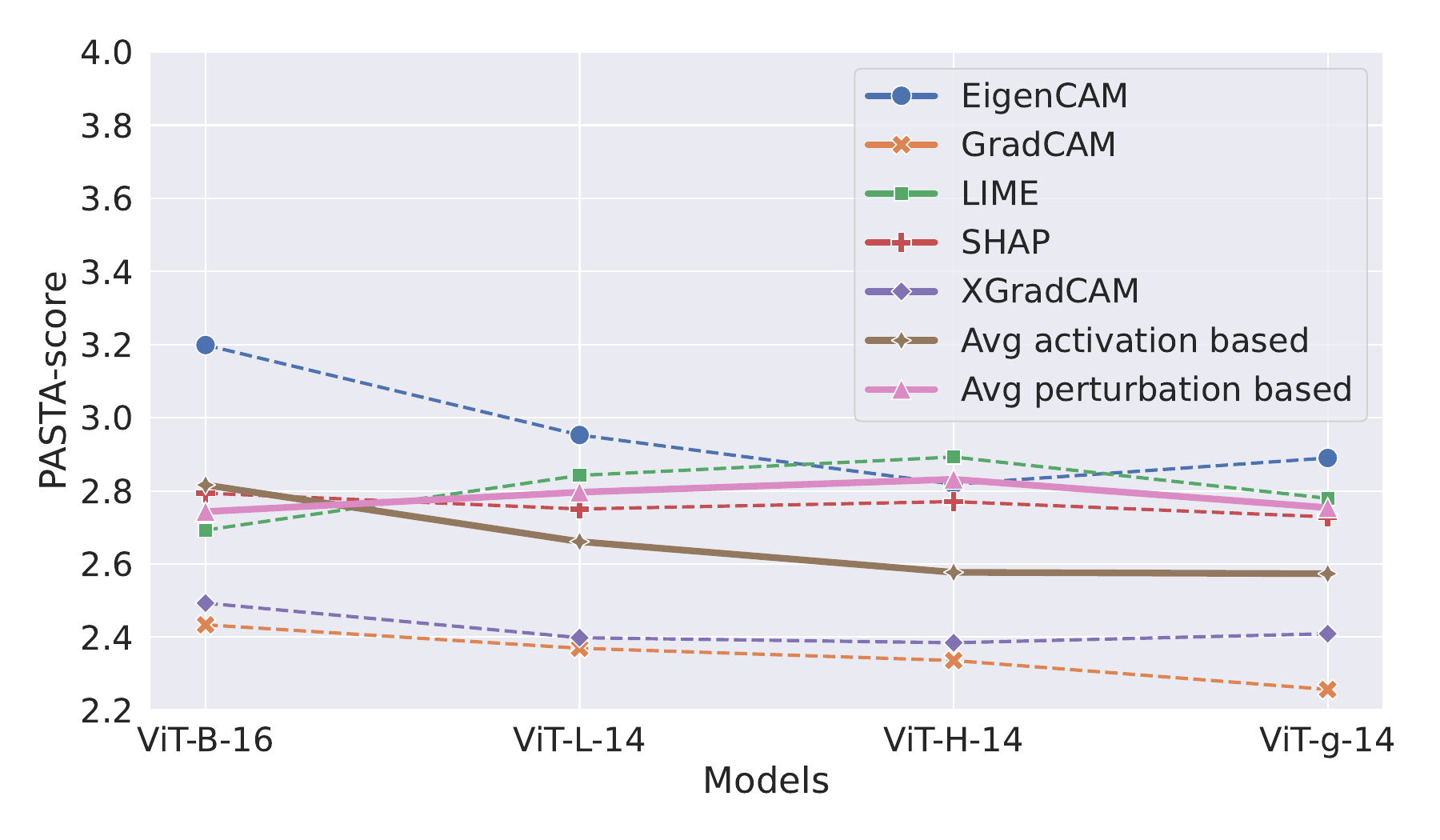}
        \caption{\remiaaai{Comparison of various XAI methods across different ViT models.}}
        \label{fig:graphmodelsize}
    \end{subfigure}
    \hfill
    \begin{subfigure}[c]{0.20\linewidth}
        \centering        \includegraphics[width=0.95\linewidth]{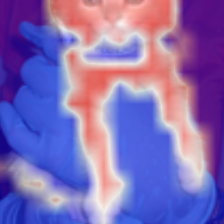}
        \caption{\remiaaai{ViT-B-16; PASTA-score of the explanation: 3.12}}
        \label{fig:sample_vitb}
    \end{subfigure}
    \hfill
    \begin{subfigure}[c]{0.20\linewidth}
        \centering        \includegraphics[width=0.95\linewidth]{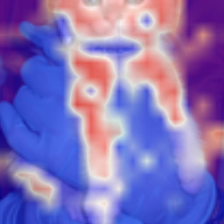}
        \caption{\remiaaai{ViT-g-14; PASTA-score of the explanation: 2.75}}
        \label{fig:sample_vitg}
    \end{subfigure}
\caption{\remiaaai{\textbf{Impact of classifier backbone size on the perceived interpretability of image explanations.} A notable decrease in the PASTA-score is observed as the model size increases (left). Examination of image samples suggests that artifacts present in the background are likely responsible for this decline.}}
    \label{fig:exp_backbone_impact}
\end{figure*}
\remiaaai{Our results show that for activation map-based XAI methods, performance metrics drop as the model size increases. This decline is particularly pronounced when transitioning from the ViT-B to the ViT-L architecture. Several hypotheses may account for this phenomenon. The most plausible explanation is the emergence of artifacts associated with high-norm tokens in the activation maps of larger models, which are used to store information \cite{darcet2023vision}. Interestingly, this decrease in score is not perceptible in image perturbation-based XAI techniques, which reinforces the hypothesis that activation artifacts contribute to the reduced interpretability of explanations.}

\subsection{\remiaaai{Steering XAI methods towards better alignment}}
\remiaaai{We propose to use the PASTA-score to enhance the interpretability of an off-the-shelf XAI method, namely RISE \citep{petsiuk2018rise}. Our approach is as follows: while RISE generates random masks and selects the one that has the best class scores $S_{proba}$, we introduce a regularization component based on the PASTA-score. Consequently, instead of rating masks using $S_{proba}$, we employ the following formula:}
\remiaaai{\begin{equation}
    w_{RISE+PASTA} = \lambda S_{PASTA} + (1-\lambda) S_{proba} \, ,
\end{equation}}
\remiaaai{where $\lambda \in [0, 1]$ is a hyperparameter. When $\lambda=0$, the generated explanation aligns with the original RISE method. Conversely, if $\lambda=1$, the explanation produced corresponds to a scenario that maximizes the PASTA-score. Note that setting $\lambda=1$ would result in an explainer optimizing only for human preferences while neglecting the true behavior of the model, which may not yield useful explanations.}

\begin{figure}[h]
    \begin{subfigure}[c]{0.30\linewidth}
        \centering        \includegraphics[width=0.99\linewidth]{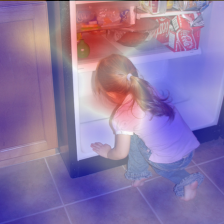}
        \captionsetup{justification=centering}
        \caption{ \remiaaai{ $\lambda = 0$\\ \textit{F} = $0.0983$ \\ \textit{P}= $4.13$}}
        \label{fig:optim11}
    \end{subfigure}
    \hfill
    \begin{subfigure}[c]{0.30\linewidth}
        \centering        \includegraphics[width=0.99\linewidth]{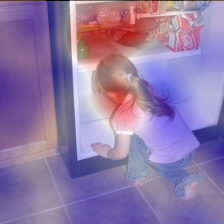}
        \captionsetup{justification=centering}
        \caption{ \remiaaai{ $\lambda = 0.7$\\ \textit{F} = $0.0874$ \\ \textit{P} = $4.19$}}
        \label{fig:optim12}
    \end{subfigure}
    \hfill
    \begin{subfigure}[c]{0.30\linewidth}
        \centering        \includegraphics[width=0.99\linewidth]{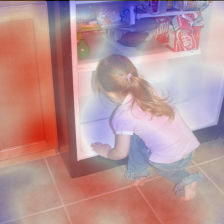}
        \captionsetup{justification=centering}
        \caption{ \remiaaai{ $\lambda = 1$\\ \textit{F} = $0.0326$ \\ \textit{P} = $4.32$}}
        \label{fig:optim13}
    \end{subfigure}
    \hfill
    \begin{subfigure}[c]{0.30\linewidth}
        \centering        \includegraphics[width=0.99\linewidth]{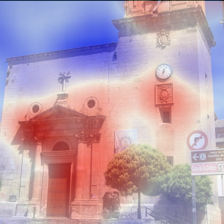}
        \captionsetup{justification=centering}
        \caption{ \remiaaai{ $\lambda = 0$\\ \textit{F} = $0.0692$ \\ \textit{P} = $2.82$}}
        \label{fig:optim21}
    \end{subfigure}
    \hfill
    \begin{subfigure}[c]{0.30\linewidth}
        \centering        \includegraphics[width=0.99\linewidth]{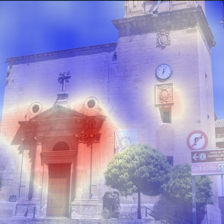}
        \captionsetup{justification=centering}
        \caption{ \remiaaai{ $\lambda = 0.7$ \\ \textit{F} = $0.0731$ \\ \textit{P} = $2.96$}}
        \label{fig:optim22}
    \end{subfigure}
    \hfill
    \begin{subfigure}[c]{0.30\linewidth}
        \centering        \includegraphics[width=0.99\linewidth]{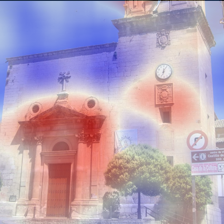}
        \captionsetup{justification=centering}
        \caption{ \remiaaai{ $\lambda = 1$ \\ F = $0.0674$ \\ \textit{P} = $3.15$}}
        \label{fig:optim23}
    \end{subfigure}
\caption{\remiaaai{\textbf{Optimized explanations derived through RISE adjusted with the PASTA-score.} Label of the top images: home\_or\_hotel. Label of the bottom images: Renaissance. \textit{F} denotes the faithfulness correlation score, \textit{P} denotes the PASTA-score.}}
    \label{fig:samples_optim}
\end{figure}

\remiaaai{Upon analyzing the samples generated through the optimization process, we initially observe a slight improvement in the localization of highlighted objects. For instance, the explanation depicted in Figure \ref{fig:optim22} exhibits fewer indecisive zones and demonstrates enhanced precision compared to the explanation shown in Figure \ref{fig:optim21}. Regarding the case where $\lambda=1$, we note that the explanation begins to hallucinate zones of interest while omitting others, like in Figure \ref{fig:optim13}. Additionally, one can observe that the PASTA-score favors large heatmaps. However, the optimized explanations do not systematically overlap with the entire zone of the prediction, suggesting that alignment with the segmentation map of the object to assess the quality of saliency-based explanations, as conducted in previous studies \cite{karmani2024kpca,li2022exploring}, may prove to be inadequate.}

\section{Conclusions}

We introduce PASTA, a novel perceptual scoring method designed to benchmark XAI techniques in a human-centric manner. We collect a large-scale benchmark dataset (\remiaaai{PASTA-dataset), and use it for an assessment of XAI explanations by human annotators. 
Based on this dataset, we develop an automated scoring method (PASTA-score) that spans previously unexplored modalities, effectively mimicking human preferences and allowing for circumventing resource-intensive user studies for applications that may benefit so.
Deploying PASTA allows new quantitative observations:
Our findings reveal a distinct preference for saliency-based explanations, identify a negative impact of backbone size, and demonstrate the potential to generate more human pleasant explanations without compromising faithfulness. These results not only align with human intuition but also corroborate visual examples, affirming the scalability and reliability of PASTA-score.}

\paragraph{Limitations:} 
First, the PASTA-score is trained on specific datasets and explanation modalities, which may limit its generalizability to other unseen domains, especially those with domain-specific semantics. Second, the human preferences captured by the PASTA-dataset may inherit the intrinsic biases of human annotators. Third, although PASTA reduces the need for costly user studies, it remains an approximation of subjective human judgment and may overlook nuanced or task-specific interpretability needs, which may justify the need for more resource-intensive ad-hoc human interactions in downstream use cases.

\paragraph{Broader impact:} Dynamic scoring approaches could be explored to capture the evolving nature of XAI techniques and their use in real-world applications. PASTA intends to take a step towards creating a transparent and trustworthy AI ecosystem. By aligning AI explanations with human preferences, we aim to foster the development of more interpretable AI systems that can be understood and trusted by users. \remirebutal{This work also introduces a perceptual metric, paving the way for future research to implement the PASTA-score as a perceptual loss aimed at enhancing the trustworthiness of networks, drawing for example, inspiration from the emerging use of LPIPS \citep{zhang2018unreasonable} in tasks such as image generation \citep{jo2020investigating}.}

\section{Acknoledgements}

This work was performed using HPC resources from GENCI-IDRIS (Grant 2024 - AD011014675R1).

\bibliography{aaai2026}

\appendix

\onecolumn

\section{A. PASTA-dataset: process}

\subsection{A.1 Generation of explanations}

\subsubsection{Classifier training}

The PASTA-dataset is designed to provide a benchmark for evaluating a wide range of XAI techniques across different explanation modalities. To ensure robustness and versatility, \remiaaai{our dataset is based on four}, publicly available datasets, each bringing distinct characteristics in terms of visual content and concept annotations. Choosing which task to focus on is a tough question. We have chosen to focus on image classification. This task can be performed in many different domains, but in order not to be too domain-specific, we decided to work on general datasets. These datasets enable the evaluation of both image-based and concept-based XAI methods. 

\remi{Then, the initial phase in constructing the PASTA-dataset involves training the various classifier models on which explanations will be generated. Specifically, we utilize ResNet50 \citep{he2016deep}, ViT-B \citep{dosovitskiy2020image}, ResNet50-BCos \citep{bohle2024b}, CLIP-Linear \citep{yan2023learning}, CLIP-QDA \citep{kazmierczak2024clip}, X-NeSyL \citep{diaz2022explainable}, and CBM \citep{koh2020concept}. These models are trained separately on each classifier's dataset referenced below.

\remiaaai{It is important to note that certain classifiers utilized in our platform necessitate concept-level annotations. Specifically, for each sample, information regarding the presence or absence of each concept in the image, along with their respective bounding boxes, is required. This detailed information is not natively available in each dataset. Consequently, we enhanced each dataset to meet this requirement. The datasets and the corresponding modifications are as follows:}

\begin{itemize}
    \item COCO: A widely-used dataset known for its complexity and variety, containing 117k training images, 4.5k validation images annotated with 80 object categories, which we consider to be concepts in the images. \remi{\remirebutal{The labels correspond for this specific dataset to indoor scene labeling}, to do so, we took the subset of images of indoor scenes (53,051 images). Then,} we labeled the images using a scene label DNN trained on the MIT SUN. 
    \item Pascal Part: This dataset focuses on detailed part-level annotations, providing fine-grained insights into object structure and component relationships. It is composed of \remi{13,192 training images, 39 concepts, and 16 classes.}
    \item Cats Dogs Cars: A curated dataset featuring images of cats, dogs, and cars. The goal of this dataset is to explore if color biases are present in the model or not. It is composed of \remi{3,858 training images, 39 concepts, and 3 classes. Since this network does not include annotated concepts, we used Grounding DINO \citep{liu2023grounding} as an annotator. \remirebutal{Since the number of images that constitute Cate Dogs Cars is sufficiently small, we manually checked the bounding boxes generated and found no significant errors.}}
    \item Monumai: A specialized dataset containing images of monuments, with annotations that include both the overall structures and specific architectural features. It is composed of \remi{908 images, 15 concepts, and 4 classes.}
\end{itemize}

Each dataset in the classifier's training datasets is annotated at two levels:
\begin{itemize}
    \item Image-level annotations: These are traditional class labels \remirebutal{(Table \ref{table:table_classes})} or object categories that describe the primary content of the image.
    \item Concept-level annotations: These describe specific, human-understandable features within the image, enabling the application of Concept Bottleneck Models (CBMs) and other concept-based XAI methods. The list of concepts for each dataset is detailed in Table~\ref{table:table_concepts}.
\end{itemize}
\begin{table}[h]
\centering
\remi{
\footnotesize
\setlength{\tabcolsep}{3pt}
\caption{\remi{\textbf{List of concepts used in all our CBMs.} For each \textit{Dataset} used, we choose a different set to fit the annotations.}}
\begin{tabular}{@{}lp{10cm}@{}}
\toprule
\textit{Dataset} & \textit{Concepts} \\ \midrule
\textbf{catsdogscars, pascalpart} & engine, artifact\_wing, animal\_wing, stern, tail, locomotive, arm, hair, wheel, chain\_wheel, handlebar, hand, headlight, saddle, body, bodywork, beak, head, eye, foot, leg, neck, torso, cap, license\_plate, door, mirror, window, ear, muzzle, horn, nose, hoof, mouth, eyebrow, plant, pot, coach, screen \\ 
\textbf{monumai} & horseshoe-arch, lobed-arch, pointed-arch, ogee-arch, trefoil-arch, serliana, solomonic-column, pinnacle-gothic, porthole, broken-pediment, rounded-arch, flat-arch, segmental-pediment, triangular-pediment, lintelled-doorway \\ 
\textbf{coco} & person, backpack, umbrella, handbag, tie, suitcase, bicycle, car, motorcycle, airplane, bus, train, truck, boat, traffic light, fire hydrant, stop sign, parking meter, bench, bird, cat, dog, horse, sheep, cow, elephant, bear, zebra, giraffe, frisbee, skis, snowboard, sports ball, kite, baseball bat, baseball glove, skateboard, surfboard, tennis racket, bottle, wine glass, cup, fork, knife, spoon, bowl, banana, apple, sandwich, orange, broccoli, carrot, hot dog, pizza, donut, cake, chair, couch, potted plant, bed, dining table, toilet, tv, laptop, mouse, remote, keyboard, cell phone, microwave, oven, toaster, sink, refrigerator, book, clock, vase, scissors, teddy bear, hair drier, toothbrush \\ \bottomrule
\label{table:table_concepts}
\end{tabular}}
\end{table}

\begin{table}[t]
\centering
\remi{
\footnotesize
\setlength{\tabcolsep}{3pt}
\caption{\remirebutal{\textbf{List of classes used in all the datasets used to train our inference models.}}}
\remirebutal{
\begin{tabular}{@{}lp{10cm}@{}}
\toprule
\textit{Dataset} & \textit{Labels} \\ \midrule
\textbf{catsdogscars} & cat, dog, car \\
\textbf{pascalpart} & aeroplane, bicycle, bird, bottle, bus, car, cat, cow, dog, horse, motorbike, person, pottedplant, sheep, train, tvmonitor \\
\textbf{monumai} & Baroque, Gothic, Hispanic-Muslim, Renaissance \\ 
\textbf{coco} & shopping\_and\_dining, workplace, home\_or\_hotel, transportation, sports\_and\_leisure, cultural \\ \bottomrule
\label{table:table_classes}
\end{tabular}}}
\end{table}

In Figure \ref{fig:Classdistribution}, we observe the class distribution across the different datasets. While the distributions are not perfectly uniform, they generally reflect the original composition of the datasets, ensuring that the diversity of the data is preserved in the evaluation process.

\begin{figure}[t]
\begin{center}\footnotesize
    \begin{tabular}{c c}
\includegraphics[width=0.45\linewidth]{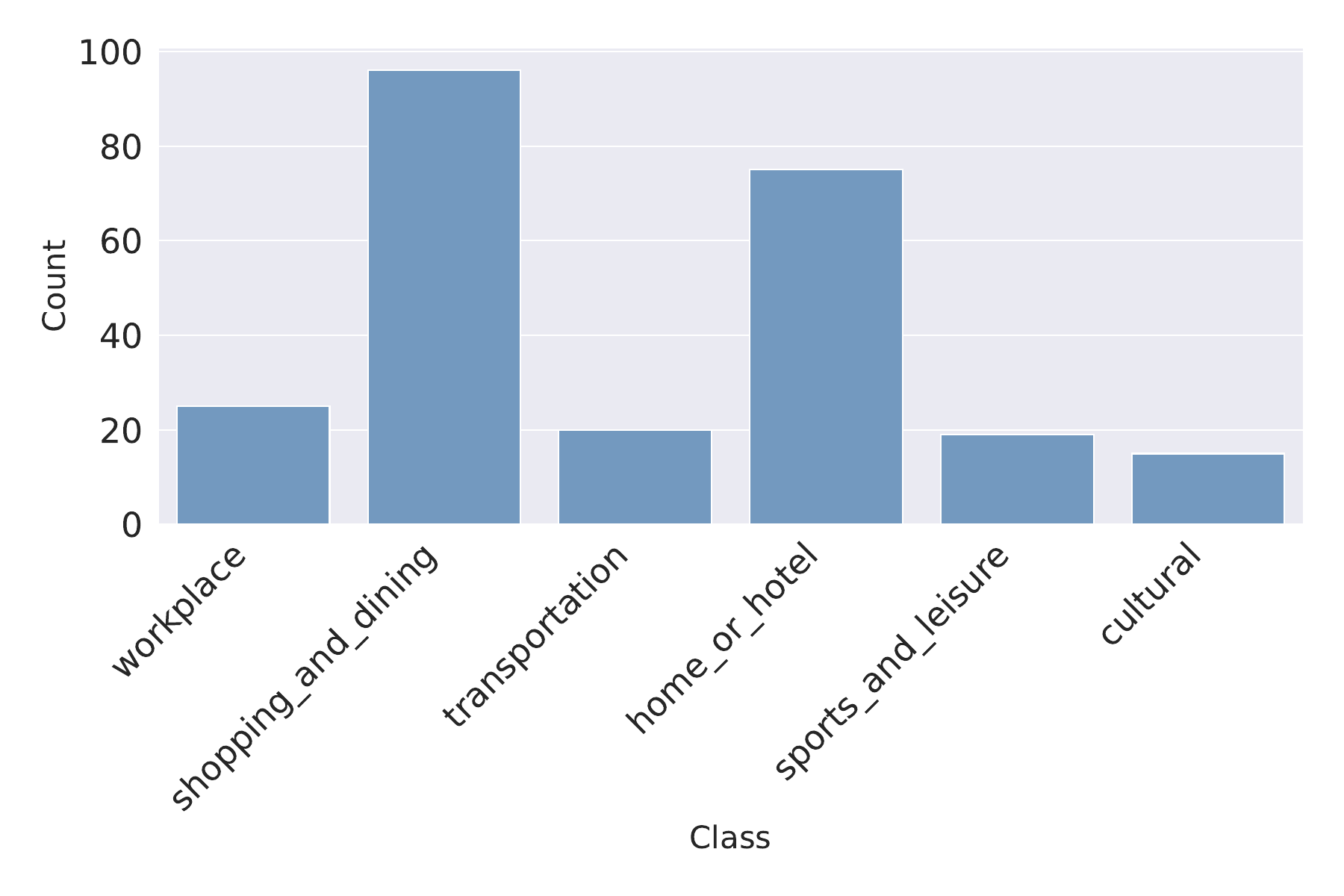} &\includegraphics[width=0.45\linewidth]{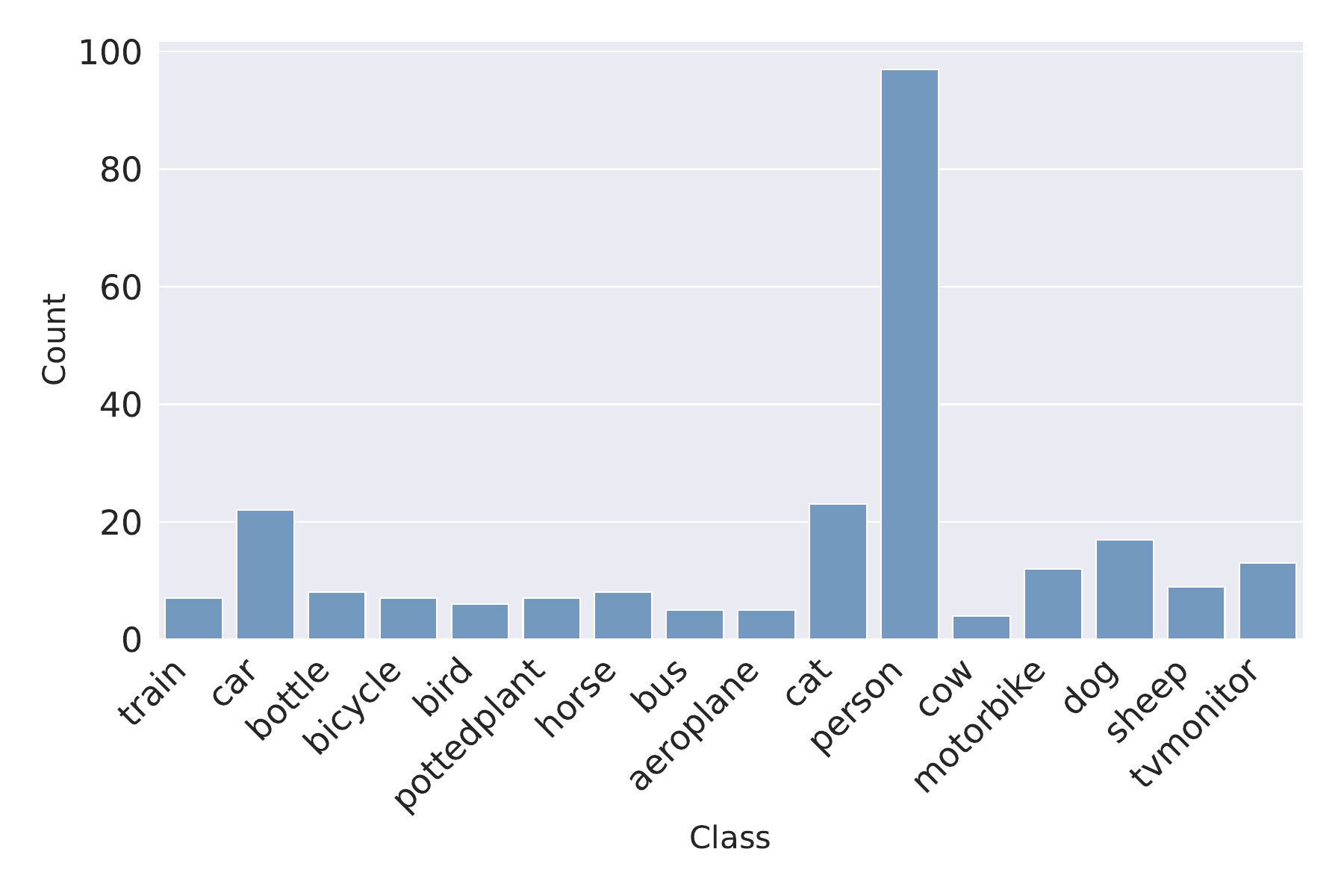} \\
      \textbf{COCO} & \textbf{Pascal Part}\\
      \\
\includegraphics[width=0.45\linewidth]{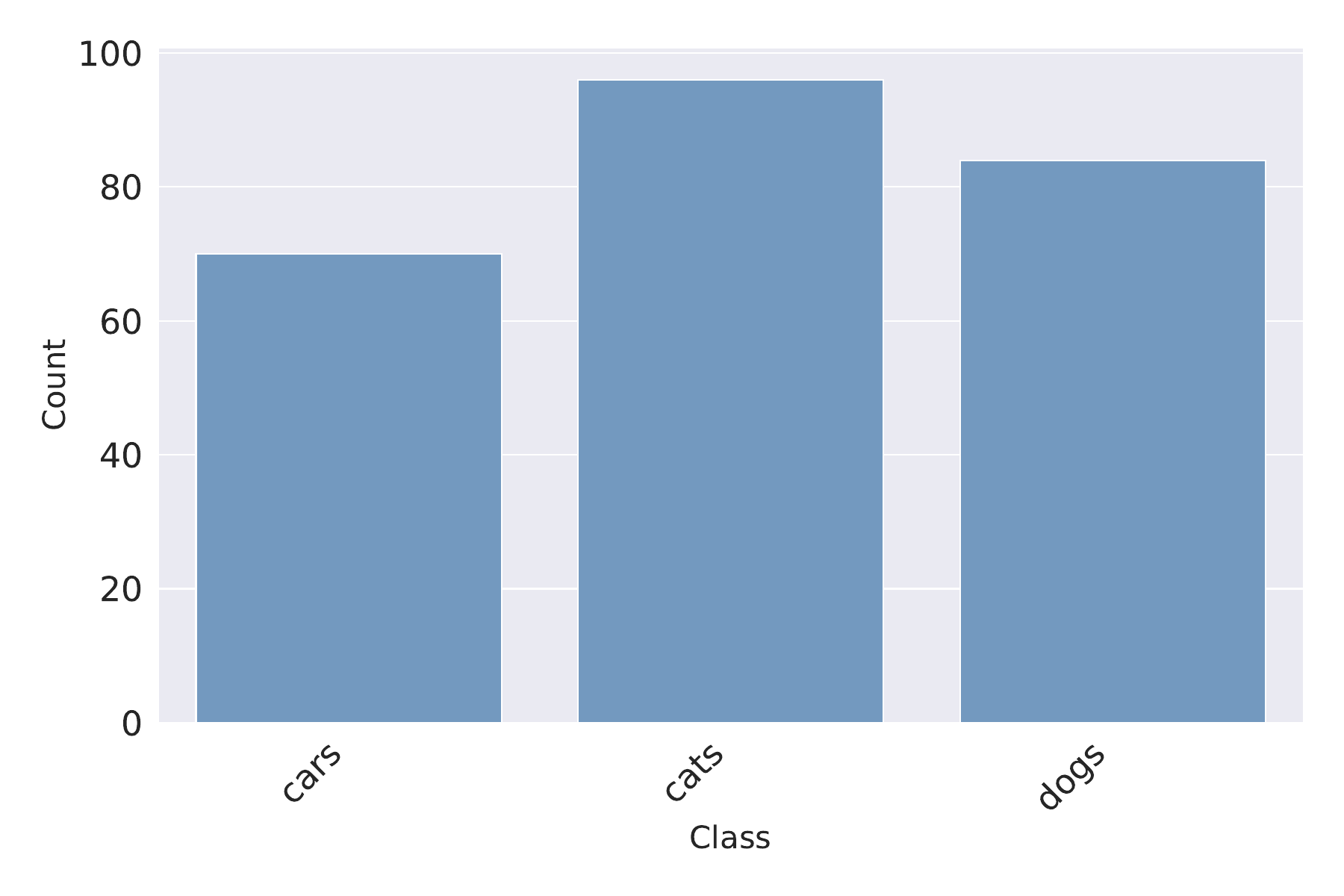} &\includegraphics[width=0.45\linewidth]{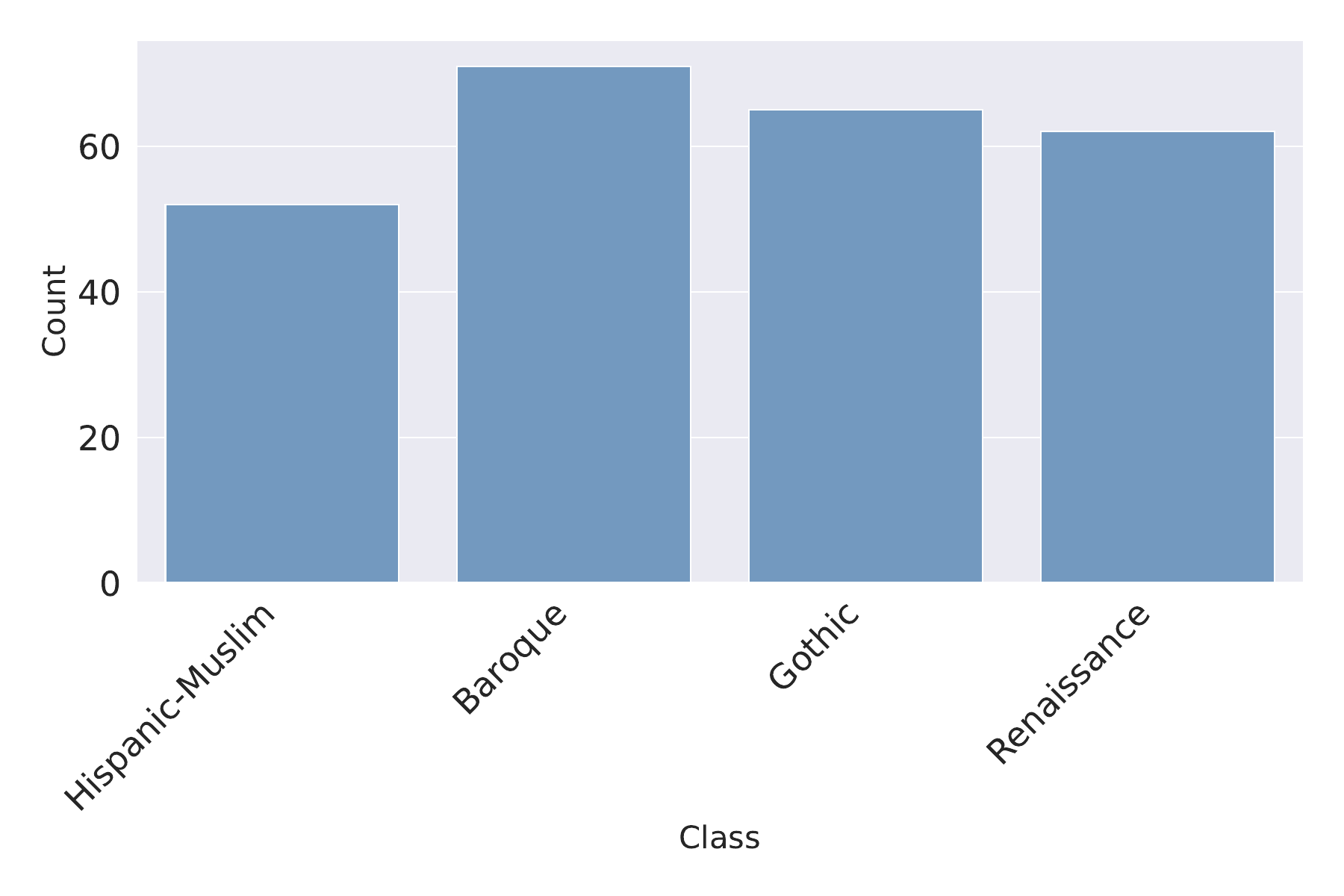} \\
      \textbf{Cats Dogs Cars} & \textbf{Monumai}
    \end{tabular}
\end{center}
\caption{ \textbf{Class distribution across the different test sets.} }
\label{fig:Classdistribution}
\end{figure}

\remiaaai{As previously indicated, additional care must be taken when training CBMs. }To explain the various training procedures for our \remi{CBMs}, we decompose \remi{them} into two components: the concept extractor and the classifier. The concept extractor generates an embedding from an input image, with each element representing a concept, while the classifier predicts the label from this embedding. We categorize the CBMs we use based on the training methods for these two components.
For CLIP-based CBMs (LaBo, CLIP-linear, and CLIP-QDA), the concept extraction is performed in a zero-shot manner \textit{i.e.}, we only use the training images and labels to train the classifier. For CBMs that require training the concept extractor \remiaaai{(X-NeSyL and ConceptBottleneck)}, we use the concept annotations provided by each dataset.

\remi{For explanations that involve the application of post-hoc techniques on black-box models, we selected the following DNNs: ResNet 50, ViT, and CLIP (zero-shot). For ResNet 50 and ViT, a separate network was trained for each dataset. For CLIP (zero-shot), we followed the standard procedure proposed by~\citet{radford2021learning}, which classifies by selecting the highest similarity score between the image embedding and all the text embeddings.} For post-hoc explanations, we directly extract the explanation after training. 

As illustrated in Figure \ref{fig:datasetaccu}, the models used in this study achieve an accuracy of at least  59\%. Notably, one of the models, the zero-shot CLIP, exhibits difficulty specifically with the Monumai dataset, which explains some of the performance variability. Despite this, the overall accuracy of the models remains relatively consistent across datasets. For CBMs, achieving high accuracy across all models required certain compromises, particularly with respect to the concepts used. Although for uniformity we used the same  concept sets across different models, it was not always guaranteed that the trained model is the best model. 

\begin{figure}[t]
\begin{center}
\includegraphics[width=0.5\linewidth]{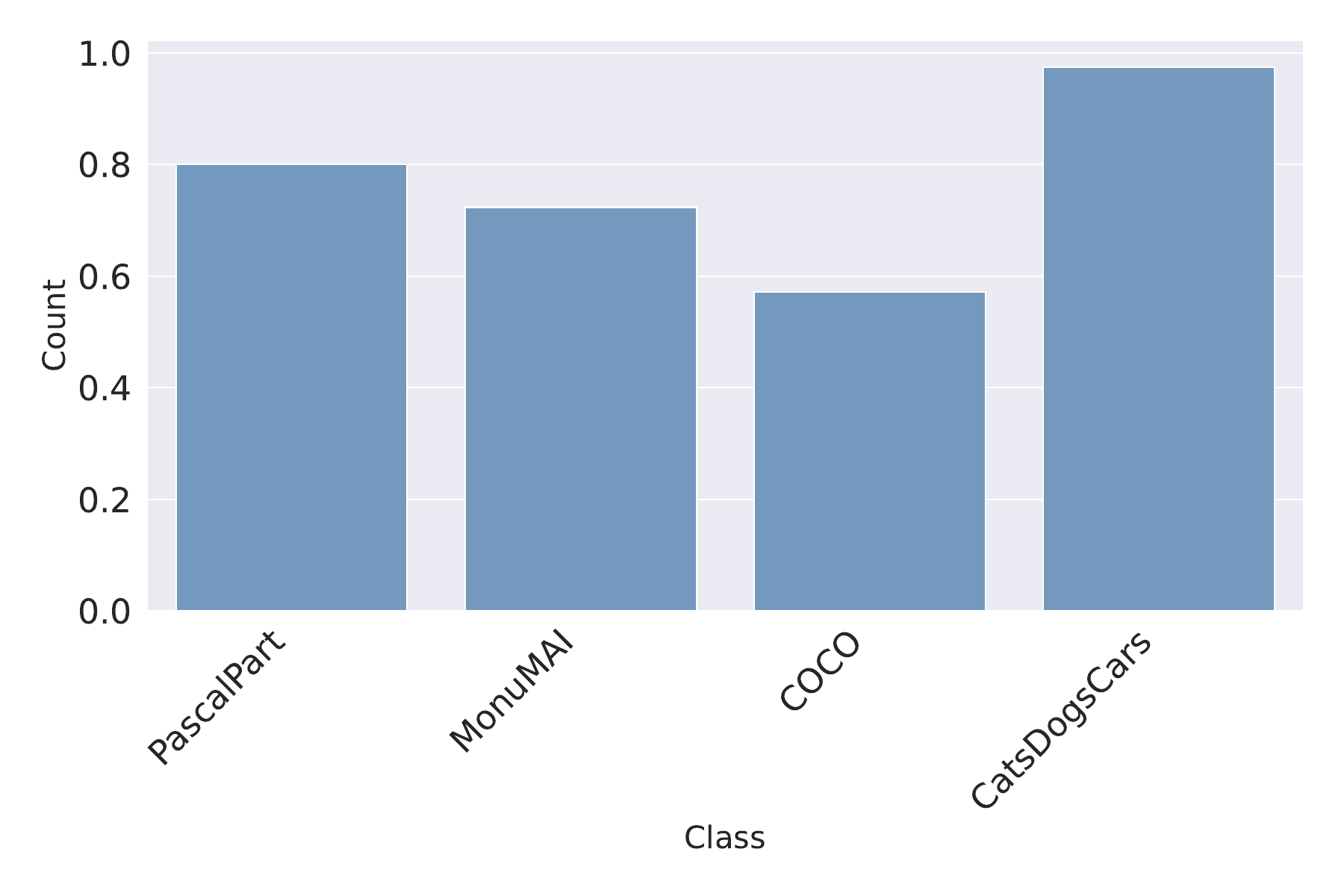} 
\end{center}
\caption{\textbf{ Accuracy across the different test sets of the different models.} }
\label{fig:datasetaccu}
\end{figure}

\subsubsection{Computation of explanations}

\remiaaai{Upon completion of the training process for all classifiers, the subsequent phase involves the generation of explanations for their respective inferences. A comprehensive enumeration of the XAI methods, along with the corresponding classifiers to which each method is applied, is presented in Table \ref{table:tested_methods}.}
\remi{For this, we use for the computation of explanations a subset of} \remiaaai{250} images \remi{of the classifier's dataset test split} per dataset, %
that serve as the basis of the PASTA-dataset. 
This diverse selection ensures a broader generalization of the XAI techniques across datasets being assessed. Note that, unlike traditional datasets, our benchmark dataset comprises a triplet of images, explanations, and labels. This triplet enables us to quantitatively assess the quality of XAI techniques.

\remiaaai{Precisely, we tested 14 saliency-based XAI techniques and 6 concept-based XAI techniques, with a particular emphasis on a variety of functioning of methods. \remirebutal{A beneficial aspect of our approach is that many of the methods are tested on multiple backbone architectures. \remirebutal{This aspect is particularly important, as many of the post-hoc methods evaluated in our study are designed to be DNN-agnostic.} 
In such cases, the XAI method is applied to independently trained models. For instance, GradCAM is evaluated on ResNet50, ViT-B, and CLIP-zero-shot models, while SHAP-CBM is applied to both CLIP-QDA and the original Concept Bottleneck model as proposed by \citet{koh2020concept}.}}

\begin{table}[htbp]

\remi{
\caption{\remi{\textbf{XAI methods included in our dataset.} \textit{Name} denotes the identifier of the utilized XAI method. \textit{Functioning} specifies the mechanism of the explanation computation, including methods that rely on gradient weighting (Gradient), probing reactions to localized perturbations (Perturbation), abstracting activations through factorization (Factorization), leveraging directly interpretable latent spaces (Interpretable latent space), or searching for counterfactuals (Counterfactual). \textit{Attribution} indicates the data type on which the attribution weights are applied: either on input images (Image) or on a computed representation of the image as concepts (Concepts). \textit{Stage} indicates whether the explanation is produced by a ante-hoc or a post-hoc process.
\protect\label{table:tested_methods}}}}
\centering
\scalebox{0.8}{
\setlength{\tabcolsep}{3pt}
\begin{tabular}{@{}lllll@{}}
\toprule
\textit{Name} & \textit{Functioning} & \textit{Attribution on} & \textit{Stage} & \textit{Applied on}\\
\midrule
BCos \citep{bohle2024b} & Interpretable latent space & Image & Ante-hoc & ResNet50-BCos \\
GradCAM \citep{selvaraju2017grad} & Gradient & Image & Post-hoc & ViT, ResNet50, CLIP (zero-shot) \\
HiResCAM \citep{draelos2020use} & Gradient & Image & Post-hoc & ViT, ResNet50, CLIP (zero-shot) \\
GradCAMElementWise \citep{pillai2021explainable} & Gradient & Image & Post-hoc & ViT, ResNet50, CLIP (zero-shot) \\
GradCAM++ \citep{chattopadhay2018grad} & Gradient & Image & Post-hoc & ViT, ResNet50, CLIP (zero-shot) \\
XGradCAM \citep{fu2020axiom} & Gradient & Image & Post-hoc & ViT, ResNet50, CLIP (zero-shot) \\
AblationCAM \citep{ramaswamy2020ablation} & Perturbation & Image & Post-hoc & ViT, ResNet50, CLIP (zero-shot) \\
ScoreCAM \citep{wang2020score} & Perturbation & Image & Post-hoc & ViT, ResNet50 \\
EigenCAM \citep{muhammad2020eigen} & Factorization & Image & Post-hoc & ViT, ResNet50, CLIP (zero-shot) \\
EigenGradCAM \citep{muhammad2020eigen} & Gradient+Factorization & Image & Post-hoc & ViT, ResNet50, CLIP (zero-shot) \\
FullGrad \citep{srinivas2019full} & Gradient & Image & Post-hoc & ViT, ResNet50\\
Deep Feature Factorizations \citep{collins2018deep} & Factorization & Image & Post-hoc & ViT, ResNet50, CLIP (zero-shot) \\
SHAP \citep{lundberg2017unified} & Perturbation & Image & Post-hoc & ViT, ResNet50, CLIP (zero-shot) \\
LIME \citep{lime} & Perturbation & Image & Post-hoc & ViT, ResNet50, CLIP (zero-shot) \\
X-NeSyL \citep{diaz2022explainable} & Interpretable latent space & Concepts & Ante-hoc & X-NeSyL \\
CLIP-linear-sample \citep{yan2023learning} & Interpretable latent space & Concepts & Ante-hoc & CLIP-linear \\
CLIP-QDA-sample \citep{kazmierczak2024clip} & Counterfactual & Concepts & Ante-hoc & CLIP-QDA \\
LIME-CBM \citep{kazmierczak2024clip}& Perturbation & Concepts & Post-hoc & CLIP-QDA, ConceptBottleneck \\
SHAP-CBM \citep{kazmierczak2024clip}& Perturbation & Concepts & Post-hoc & CLIP-QDA, ConceptBottleneck \\
RISE\remiicml{-CBM} \citep{petsiuk2018rise}& Perturbation & Concepts & Post-hoc & ConceptBottleneck \\
\bottomrule
\end{tabular}
}
\end{table}

A brief description of each method is provided below to summarize their key features and mechanisms.

\textbf{LIME (Local Interpretable Model-agnostic Explanations)}: LIME explains individual predictions of any classifier by approximating it locally with an interpretable model. It perturbs the input and observes how the predictions change, identifying the most influential parts of the input for the prediction.

\textbf{SHAP (SHapley Additive exPlanations)}: SHAP is a unified approach to interpreting model predictions based on Shapley values from cooperative game theory. It assigns each feature an importance value for a particular prediction, offering a sound measure of feature importance.

\textbf{GradCAM (Gradient-weighted Class Activation Mapping)}: GradCAM visualizes the regions in an image that contribute to the classification. It uses the gradients of the target concept (e.g., a specific class) flowing into the final convolutional layer to produce a coarse localization map highlighting important regions.

\textbf{AblationCAM}: AblationCAM improves GradCAM by iteratively removing parts of the input and observing the output effect to identify important regions.

\textbf{EigenCAM}: EigenCAM applies PCA to the activations of the last convolutional layer to produce a saliency map. It highlights the directions in which activations show the most variance, identifying critical features.

\textbf{FullGrad}: FullGrad computes gradients of the output with respect to both the input and intermediate layer outputs, aggregating these gradients to generate a comprehensive saliency map.

\textbf{GradCAMPlusPlus}: GradCAMPlusPlus improves GradCAM with a refined weighting scheme for the gradients, allowing better handling of multiple occurrences of the target concept.

\textbf{GradCAMElementWise}: GradCAMElementWise extends GradCAM by considering element-wise multiplications of gradients and activations, producing more precise visual explanations. 

\textbf{HiResCAM}: HiResCAM improves on class activation mapping by using higher-resolution feature maps for more detailed visual explanations.

\textbf{ScoreCAM}: ScoreCAM improves CAM methods by using output scores to weight the activation maps' importance, providing a more faithful saliency map without relying on gradients.

\textbf{XGradCAM}: XGradCAM integrates cross-layer information to combine saliency maps from different layers.

\textbf{DeepFeatureFactorization}: This method decomposes feature representations learned by a deep model into interpretable factors. It provides insights into how features contribute to the model's decisions.

\textbf{CLIP-QDA-sample}: This model uses the CLIP framework and applies Quadratic Discriminant Analysis (QDA) for classification. \remiaaai{This methodology employs counterfactuals on the conceptual representations of images to generate explanations}.

\textbf{CLIP-Linear-sample}: \remiaaai{This model also uses the CLIP framework but employs logistic regression for classification, thereby offering interpretable explanations grounded in the transparency of the regression analysis.}

\textbf{X-NeSyL}: X-NeSyL identifies concepts using object detection and applies a small DNN to these concepts, using the weights assigned to each concept for explanation.

\textbf{LIME CBM}: This model generates a list of concepts and applies logistic regression. \remiaaai{The methodology employs LIME to identify and highlight the most significant concepts at the conceptual level for classification purposes.}

\textbf{SHAP CBM}: This model generates a list of concepts and applies logistic regression, using SHAP \remiaaai{on the concept level} to emphasize the most crucial concepts in classification.

\textbf{Labo}: \remiaaai{Similar to CLIP-Linear-sample, Labo extracts human-interpretable concepts and maps them to the model's internal representations to facilitate more comprehensible decision-making explanations. This process leverages the transparency of its classification mechanism, albeit utilizing a custom transparent network.}

\textbf{RISE}: RISE (Randomized Input Sampling for Explanation) generates heatmaps by perturbing input regions and measuring their impact on model outputs. This technique identifies the most influential regions in the model’s decision-making process.

\textbf{BCos}: BCos introduces specific layers to encourage alignment between weights and activation maps, which can then be used for explainability.

\remiaaai{The PASTA dataset comprises 21,100 instances, each containing images, predictions, and explanations. These instances were subsequently evaluated by human annotators, resulting in 633,000 unique Likert ratings. This extensive evaluation was achieved by asking six questions to five annotators for each instance. We aggregate these evaluations using majority voting to favor consensus opinions. This dataset size is fairly standard in the Automated Essay Scoring literature \citep{lee2024prometheusvision}}, where the objective is to train a model to predict human-assigned scores.} Figure \ref{fig:XAIdistribution} shows the distribution of XAI techniques applied across the datasets. To enhance the generalizability of our results, we increased the diversity of XAI techniques used. This was achieved by not applying every technique to every image uniformly, allowing for a more diverse set of explanations to be generated. This variability ensures that our analysis captures a broad spectrum of interpretability techniques, providing deeper insights into the performance of XAI techniques across different datasets and models.

\begin{figure}[t]
\begin{center}\footnotesize
    \begin{tabular}{c c}
\includegraphics[width=0.49\linewidth]{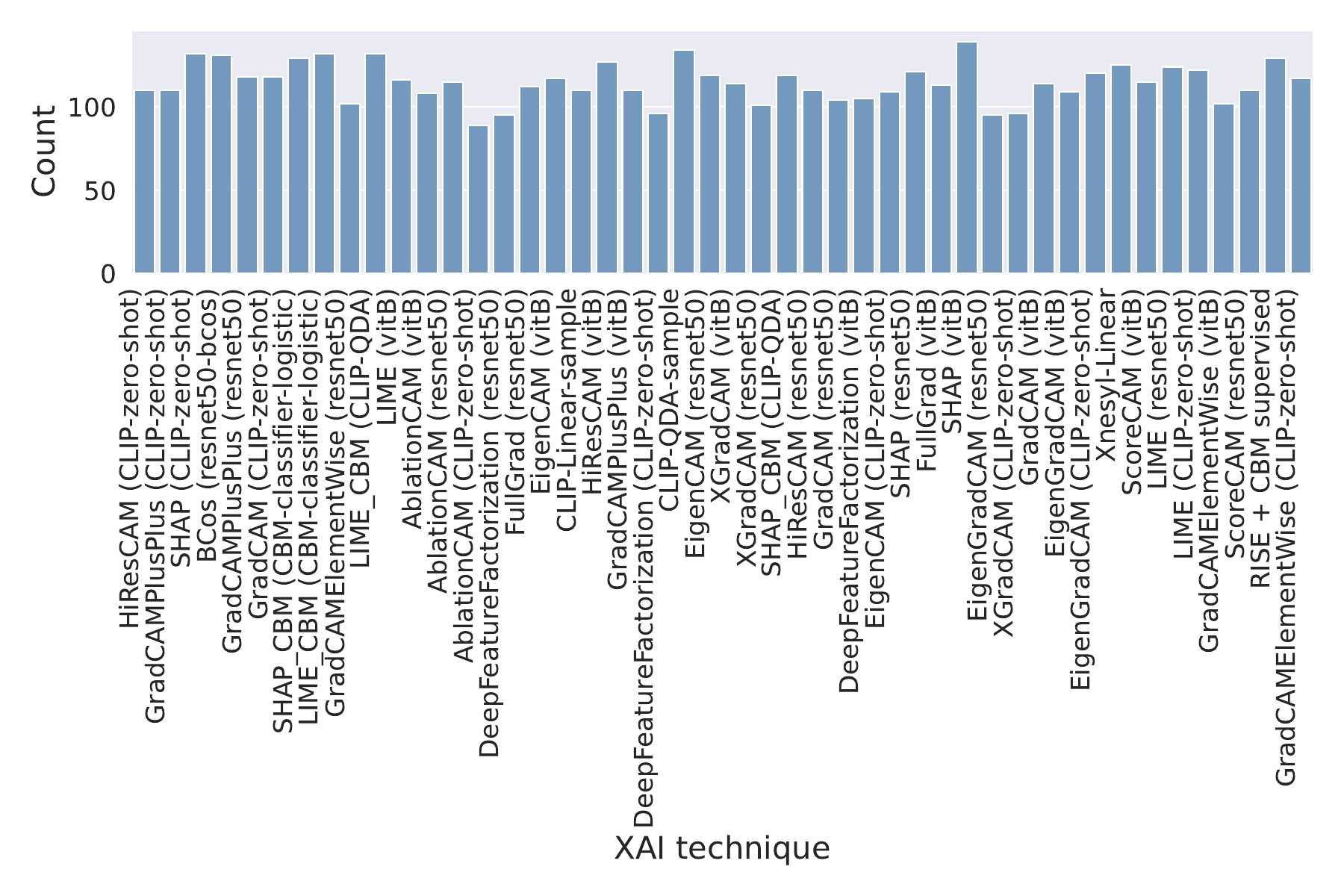} &\includegraphics[width=0.49\linewidth]{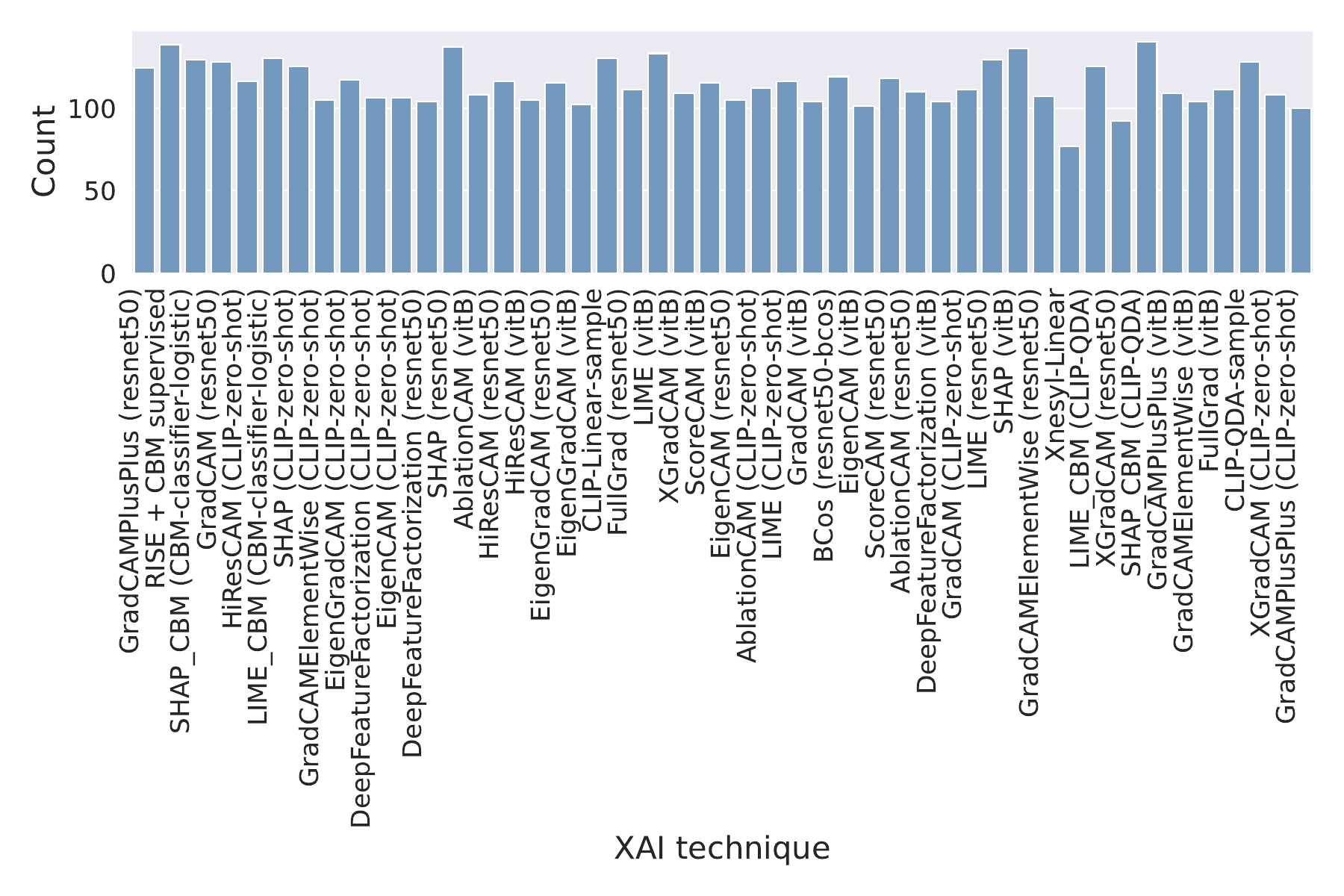} \\
      \textbf{COCO} & \textbf{Pascal Part}\\
      \\
\includegraphics[width=0.49\linewidth]{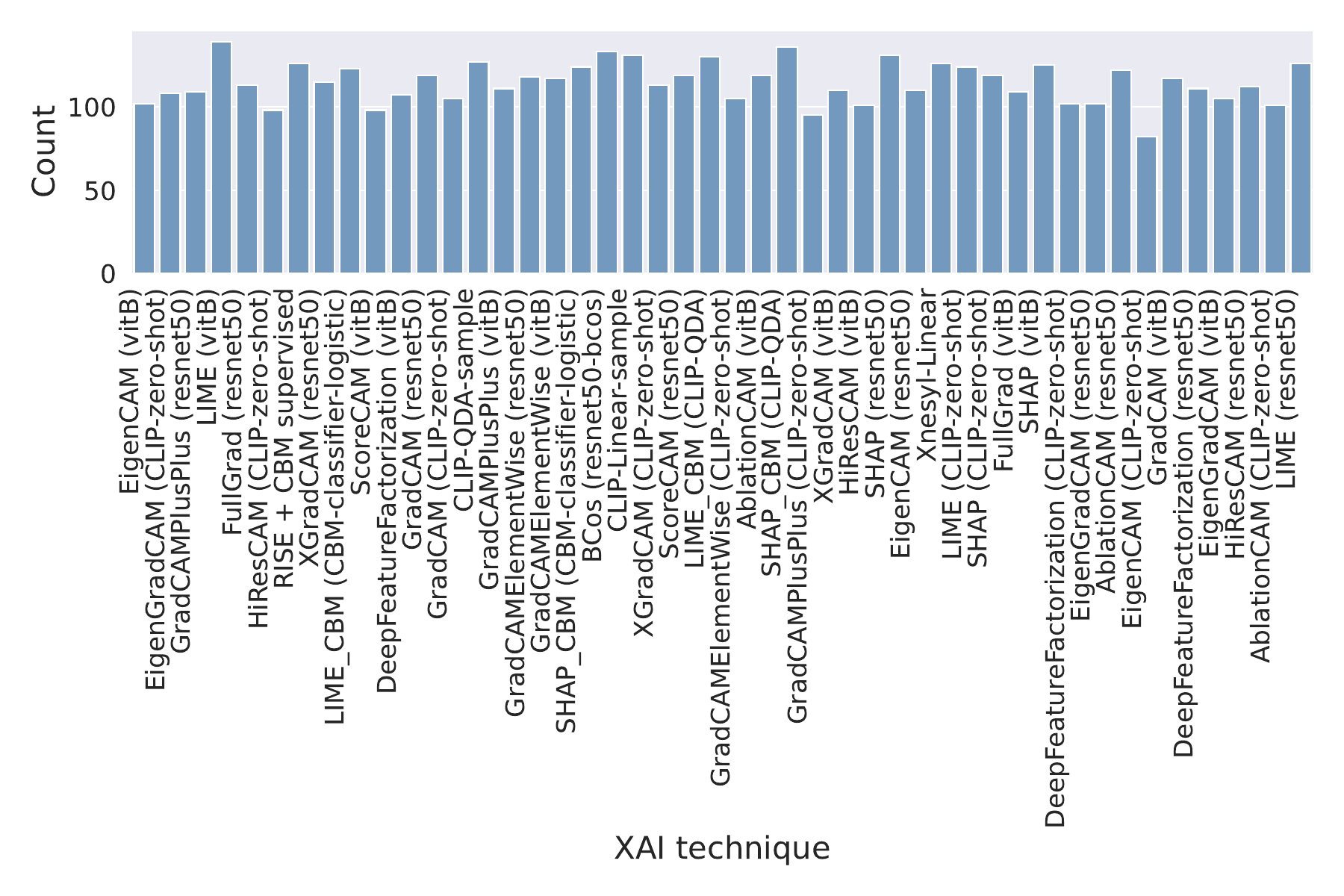} &\includegraphics[width=0.49\linewidth]{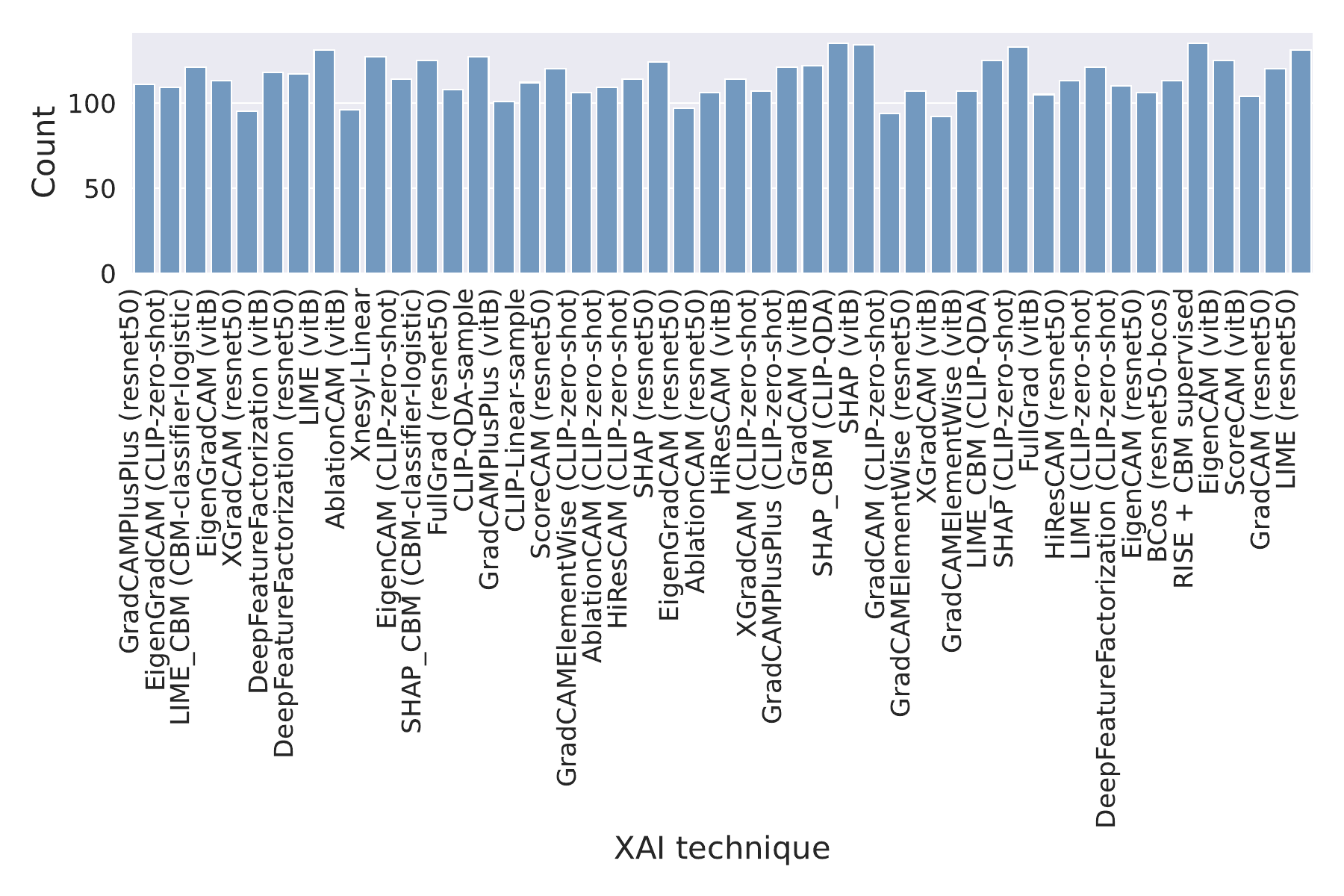} \\
      \textbf{Cats Dogs Cars} & \textbf{Monumai}
    \end{tabular}
\end{center}
\caption{\textbf{Distribution across the different XAI techniques across the different datasets.} }
\label{fig:XAIdistribution}
\end{figure}

\subsection{A.2 Human evaluation protocol }

\remiaaai{Once the explanations are performed,} we can quantify the interpretability and usefulness of XAI techniques accurately, using a human evaluation of the quality of \remiicml{explanations}. 
Our human-centric approach complements existing approaches that focus primarily on assessing the model's internal behavior. For example, traditional evaluations of \textit{faithfulness} measure how closely an explanation corresponds to the model's true functioning, \remi{while we assess in our dataset how the explanation fits human expectations.}

\subsubsection{Desiratas}

First, we establish a comprehensive set of assessment criteria that are evaluated on a graded scale. 
We consolidate different criteria from the literature into the following set of \emph{desiderata} for XAI explanations that we wish to evaluate:
\begin{itemize}[leftmargin=1.25em] 
\itemsep0em 
\item \textit{\remiaaai{Trustworthiness}}\citep{arrieta2020explainable} \remiaaai{measures the extent to which an explanation accurately reflects whether a model will act as intended when facing a given problem.}
\item \textit{Robustness} \citep{doshi2017towards, agarwal2022rethinking, yeh2019fidelity} assesses the stability and relevance of the explanation across a broad range of models and inputs. 
\item \textit{Complexity} \citep{nauta2023anecdotal, nguyen2020quantitative, bhatt2020faithcorr} checks whether the explanation is both simple and informative, balancing clarity and detail. 
\item \textit{Objectivity} \citep{bennetot2022greybox} evaluates whether the explanation is interpreted consistently by the majority %
within a given audience. 
\end{itemize}

\subsubsection{Evaluation Protocol}

Then, we apply an evaluation protocol, developed with the help of a psychologist, to ensure that annotators fully understand the task and the expectations. This includes annotator training and close monitoring throughout the %
process.
Interfaces, as well as the formulation of questions,  play a key role in the quality of the annotations~\citep{pommeranz2012designing}, and their design must be considered cautiously to avoid confounding cognitive biases. 
The formulation of the questions has been carefully chosen to ensure that they are fully understood by each annotator.
To maintain consistency and reliability, all annotators undergo a training session before starting the actual annotation task. This training familiarizes them with the XAI techniques, evaluation criteria, rating scale and datasets, ensuring a uniform understanding of the task and the expectations.

The annotation process took place via an online web application, created and deployed by a contracting company. \remiaaai{24} participants were recruited to take part in the annotation process. These participants ranged in age from 19 to 37 (mean age \remiaaai{24.58}, standard deviation  \remiaaai{3.19}). Figure \ref{fig:agedistribution} shows the age distribution. Among the participants, \remiaaai{7} identified themselves as male, \remiaaai{17} as female, 0 as non-binary, 0 did not wish to say. All participants were based in India. Each participant's task was to annotate 147 %
explanations. \remiaaai{To address the subjectivity inherent in the task, each instance, comprising a triplet of image, explanation, and label, is evaluated by five different annotators.}

\begin{figure}[t]
\begin{center}
\includegraphics[width=0.5\linewidth]{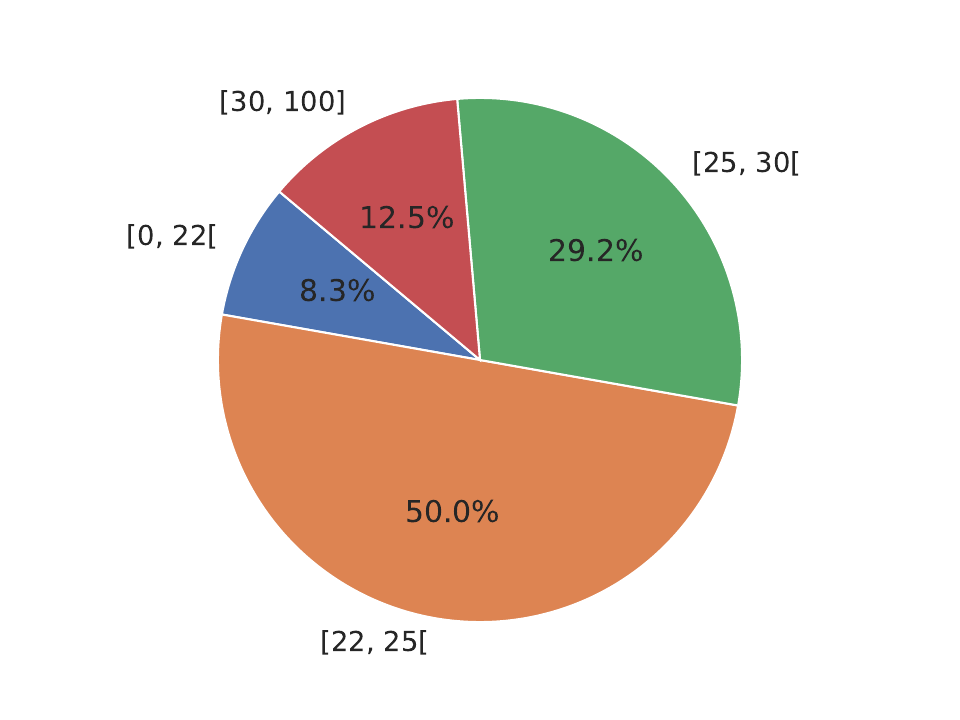} 
\end{center}
\caption{ \textbf{Age distribution of the annotators.}}
\label{fig:agedistribution}
\end{figure}

\remiicml{Concretely, annotators are shown a succession of samples that consist in an image, a prediction, and an explanation.  They are then asked to respond to a set of questions corresponding to the desiderata outlined above (indicated in italics):}
\begin{itemize}[leftmargin=1.25em]
    \itemsep0em 
    \item \remi{Q1}: \remi{Is the provided explanation consistent with how I would explain the predicted class?} \textit{\remiaaai{Trustworthiness}}
    \item \remi{Q2}: \remi{Overall the explanation provided for the model prediction can be trusted?} \textit{\remiaaai{Trustworthiness}}
    \item \remi{Q3}: Is the explanation \remi{easy to understand}? \textit{Complexity}
    \item \remi{Q4}: \remi{Can the explanation be understood by a large number of people, independently of their demographics (age, gender, country, etc.) and culture?} \textit{Objectivity}
    \item \remi{Q5}: With this perturbed image, \remi{to what extent has the explanation changed ?} (Examples with good predictions \remi{and light perturbations}) \textit{Robustness} %
    \item \remi{Q6}: With this perturbed image, \remi{to what extent has the explanation changed?} (Examples with bad predictions \remi{and strong perturbations}) \textit{Robustness}
\end{itemize}
\remiaaai{A screenshot of the interface used by the annotators to answer is available in Figure~\ref{fig:screenshot-interface}}. The first four questions \remi{(Q1 to Q4)} concerned Section 1, which showed the original image on the left and the explanation on the right. These first four questions enabled participants to assess the levels of reliability, complexity and objectivity of the explanation. The fifth question \remi{(Q5)} concerned Section 2, showing the slightly disturbed original image and the corresponding explanation. The sixth question \remi{(Q6)} concerned Section 3, showing a more disturbed image and the corresponding explanation. These last two questions were intended to assess the robustness of the XAI technique. For each question, participants had to answer with a 5-point Likert scale. In question \remi{5}, we apply a weak data augmentation to the input image, which is designed to preserve the classifier's prediction. This allows us to evaluate whether the explanation changes when the prediction remains constant. Measuring changes in explanations is challenging, as studied by \citep{fel2022good}. To address this, we leverage human evaluators who can more effectively discern subtle changes in explanations. Q6 follow a similar approach but involve strong data augmentation. The goal here is to determine if the explanations remain consistent when the predictions change due to data augmentation or variations in the models. If the predictions change, the explanations should reflect these changes. For more details, refer to \remi{Figure \ref{fig:Q5Q6}}.

\begin{figure}[t]
    \centering
    \begin{subfigure}[b]{0.60\textwidth}
        \centering
        \includegraphics[width=\textwidth]{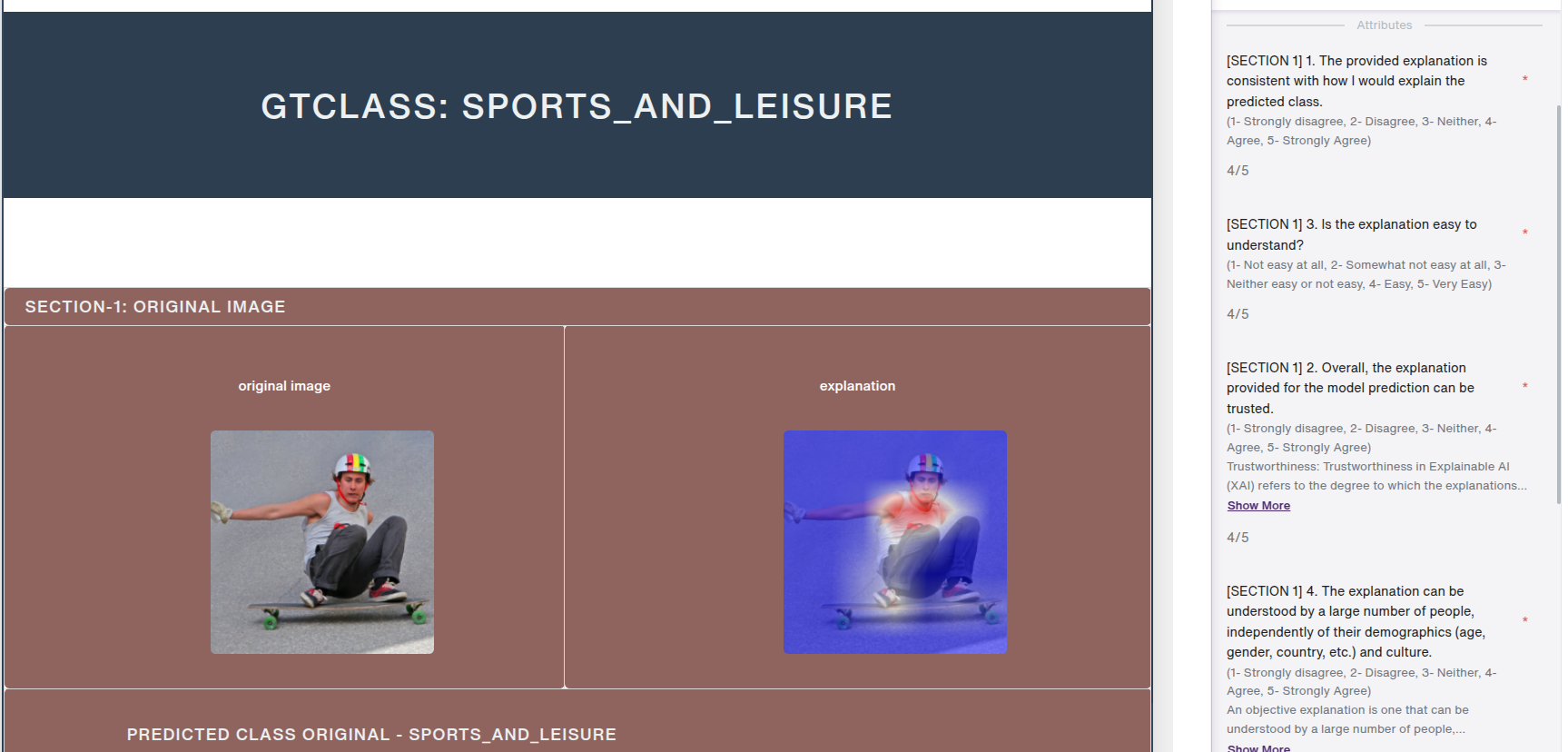}
        \label{fig:q5_1}
    \end{subfigure}
    \begin{subfigure}[b]{0.60\textwidth}
        \centering
        \includegraphics[width=\textwidth,trim={1.5cm 0 0 0},clip]{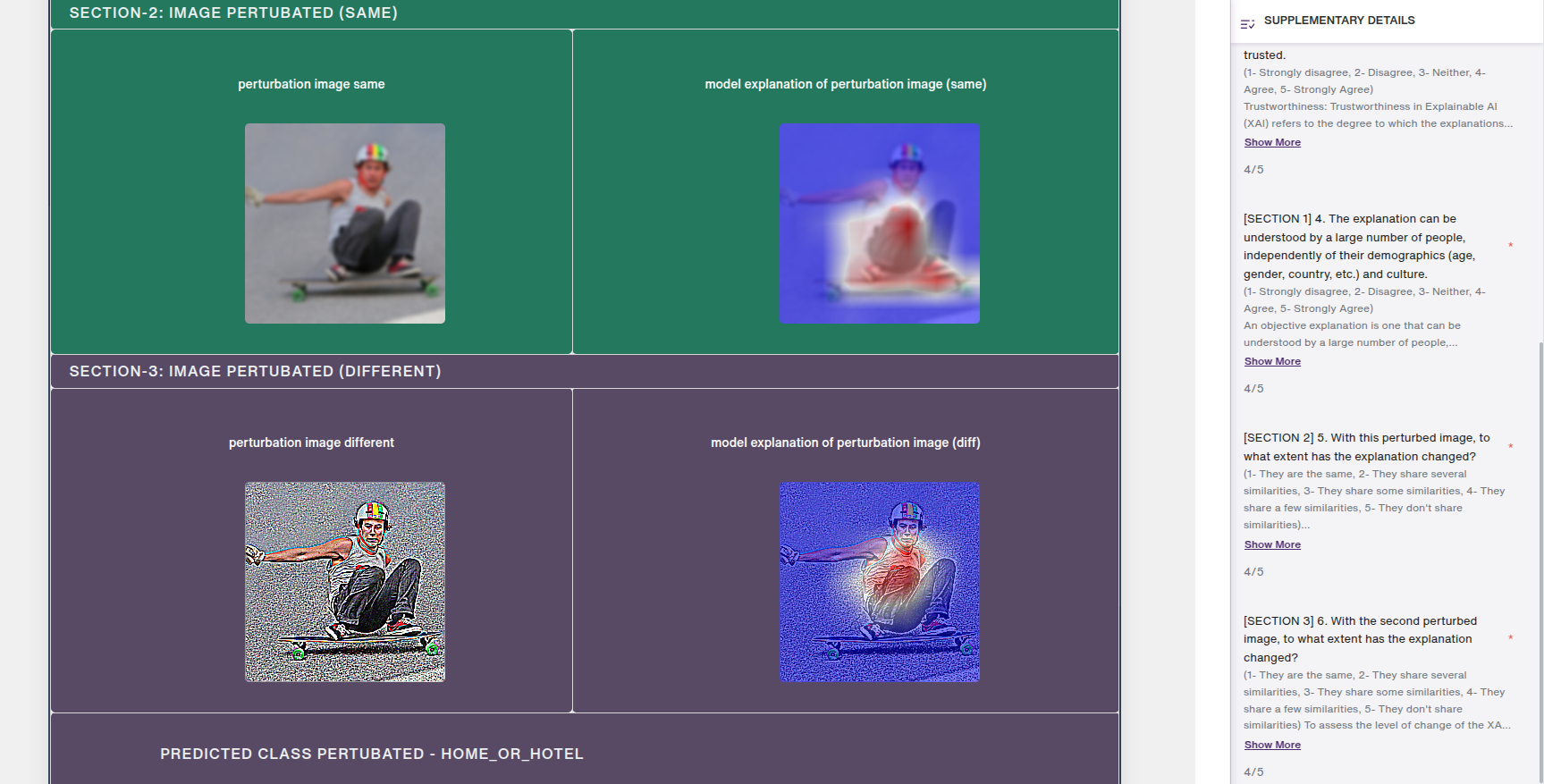}
        \label{fig:q5_2}
    \end{subfigure}
    \caption{\textbf{Screenshot of the annotation interface.} Questions are on the right-part of the interface. Middle panel shows Section 2: slightly disturbed original image with the explanation.}
    \label{fig:screenshot-interface}
\end{figure}

\remirebutal{We outline the perturbation process below. To capture a diverse range of model responses, we applied 12 distinct types of perturbations, each with a tunable magnitude parameter to adjust perturbation intensity. The transformations include both standard torchvision operations (\url{https://pytorch.org/vision/0.9/transforms.html}) and custom-designed modifications:}

\remirebutal{\begin{itemize}
    \item \textbf{Color Jitter (Brightness)}: Adjusts the brightness of the image. The magnitude lower the brightness.
    \item \textbf{Color Jitter (Contrast)}: Modifies image contrast. The magnitude lower the contrast. 
    \item \textbf{Random Resized Crop}: Performs a random crop and resize. The magnitude augments the scale of the crop.
    \item \textbf{Gaussian Blur}: Blurs the image using a Gaussian filter. The magnitude augments the values of the standard deviation.
    \item \textbf{Random Perspective}: Applies a perspective transformation. The magnitude augments the distortion scale.
    \item \textbf{Brightness Transform}: Independently changes brightness levels. The magnitude lower the brightness.
    \item \textbf{Color Transform}: Adjusts color balance. The magnitude augments the saturation.
    \item \textbf{Contrast Transform}: Further modifies contrast. The magnitude augments the contrast.
    \item \textbf{Sharpness Transform}: Changes image sharpness. The magnitude augments the sharpness factor.
    \item \textbf{Posterize Transform}: Reduces color depth. The magnitude augments the number of bits to keep for each channel.
    \item \textbf{Solarize Transform}: Inverts colors above a certain threshold. The magnitude augments the threshold.
    \item \textbf{Random Masking}: Masks out random sections of the image by applying patches. The magnitude augments the number of patches.
\end{itemize}}

Our aim was to avoid influencing responses and to eliminate any ambiguities that could lead to inaccurate answers. Human decision-making, especially when it involves assessing the quality of explanations, is complex. To address this, we provided training on deep learning and various XAI techniques, ensuring the content was clearly understandable by the annotators. After the initial training, the annotators answered the questions, and we held weekly meetings to clarify any confusion they encountered.

\begin{figure}[h]
    \centering
    \begin{subfigure}[b]{0.48\textwidth}
        \centering
        \includegraphics[width=\textwidth]{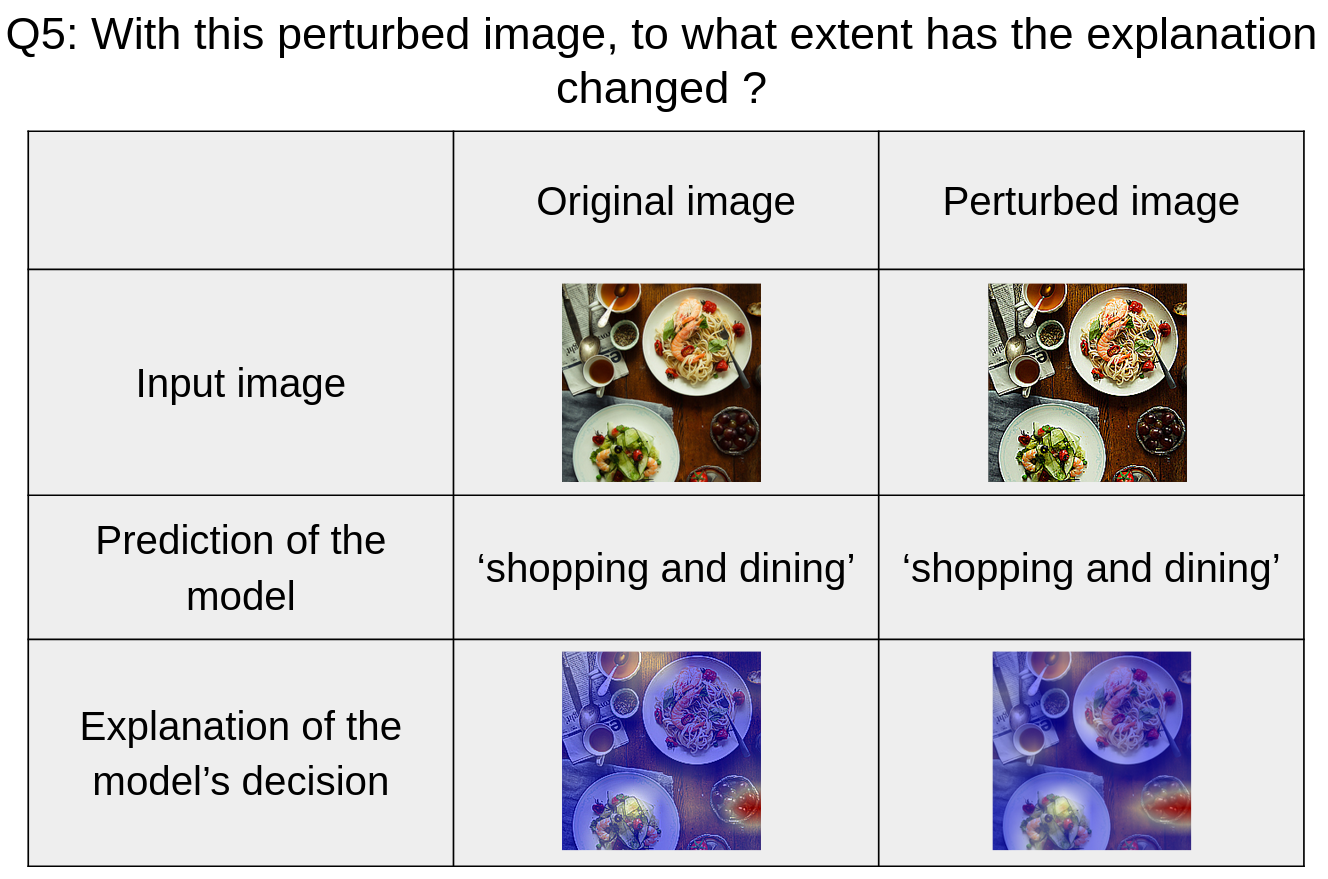}
        \caption{A light perturbation is applied, resulting in no label change.}
        \label{fig:q5}
    \end{subfigure}
    \hfill
    \begin{subfigure}[b]{0.48\textwidth}
        \centering
        \includegraphics[width=\textwidth]{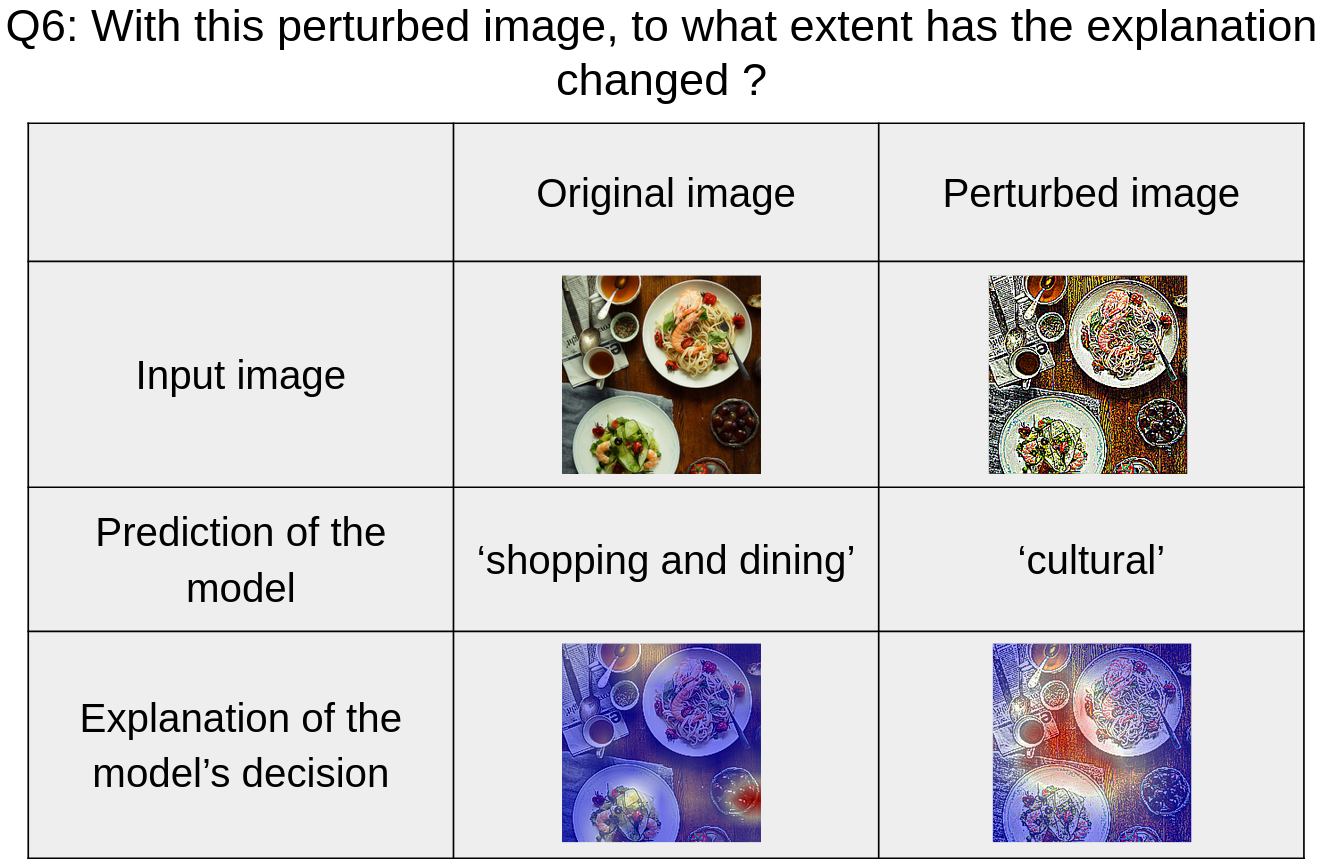}
        \caption{A strong perturbation is applied, leading to a label change.}
        \label{fig:q6}
    \end{subfigure}
    \caption{Samples corresponding to questions 5 and 6.}
    \label{fig:Q5Q6}
\end{figure}

Each explanation is rated on a \remiicml{five points Likert scale}, where one star indicates \remiicml{a total disagreement} with the annotator's reasoning, and five stars represent \remiicml{total agreement}.

\section{\remiaaai{B. PASTA-dataset: Additional analysis}}

\subsection{B.1 Comparison with existing benchmark} \label{comp_exist_bench}

\remirebutal{We compiled a set of related works that perform human assessment in the context of XAI. Specifically, we noted key details such as the dataset size, the number of participants involved, the diversity of questions posed, and the overall scope of the study, where this information was available. A summary of these details is presented in Table \ref{tab:xai_evaluation_frameworks}.}

\remirebutal{The PASTA-dataset distinguishes itself in several key aspects. First, it evaluates a significantly higher number of XAI methods compared to existing datasets. This design emphasizes the diversity of techniques over the number of samples tested, offering a complementary approach to datasets that prioritize varied input data but evaluate fewer methods. Second, PASTA involves the lowest number of participants among the datasets listed, allowing for reduced variability due to potential outliers and better control over annotator behavior. However, this could introduce inherent biases tied to the limited participant pool. Finally, the PASTA-dataset provides a substantially larger number of samples and uniquely combines image-based explanations with concept-based explanations, making it the first dataset to address both modalities simultaneously.}

\begin{table}[htbp]
\centering
\caption{\remirebutal{\textbf{Overview of datasets and human evaluation frameworks for XAI methods.} \textbf{Name} refers to the reference of the dataset used. \textit{Annotations} refers to the type of labels used: Likert refers to Likert scale, Saliency refers to pseudo saliency maps, 2AFC refers to two alternative forced choices, Clictionary refers to the clicktionary game defined in \citep{dawoud2023human}, MCQ refers to multiple-choice question, and Binary refers to binary choises. \textbf{$N_{Samples}$} refers to the total number of samples constituting the dataset. \textbf{$N_{Part}$} refers to the number of participants involved during the labeling. \textit{Modality} refers to the different modalities the dataset deals with: I refers to image, C to concepts, and T to text. \textbf{$N_{Q}$} refers to the number of different questions asked to the annotators. \textbf{$N_{XAI}$} refers to the number of XAI methods tested during the experiments, \textit{No} indicates that the dataset only asked to label data with what they consider as ground truth explanation, without further comparison with XAI methods. \textbf{$N_{Data}$} refers to the number of different samples (for example images) shown to annotators.}}
\label{tab:xai_evaluation_frameworks}
\remirebutal{\begin{tabular}{llcclccc}
\hline
\textbf{Name} & \textit{Annotations} & \textbf{$N_{Samples}$} & \textbf{$N_{Part}$} & \textit{Modality} & \textbf{$N_{Q}$} & \textbf{$N_{XAI}$} & \textbf{$N_{Data}$} \\ 
\hline
PASTA-dataset & Likert & \remiaaai{633,000} & \remiaaai{24} & I + C & 6 & \remiaaai{20} & \remiaaai{1000} \\ 
\citet{yang2022psychological} & Saliency, 2AFC & 356 & 46 & I & 2 & 1 & 89 \\ 
\citet{colin2022cannot} & Classification & 1,960 & 241 & I & 1 & 6 & NA \\ 
\citet{dawoud2023human} & Clicktionary & 3,836 & 76 & I & 1 & 3 & 102 \\ 
\citet{mohseni2021quantitative}& Saliency & 1,500 & 200 & I + T & 1 & \textit{No} & 1,500 \\ 
\citet{herm2021don} & Likert & NA & 165 & C & 1 & 6 & NA \\ 
\citet{morrison2023eye} & Clicktionary/QCM & 450 & 50 & I & 1 & 3 & 39 \\ 
\citet{spreitzer2022evaluating} & Likert/Binary & 4,050 & 135 & C & 9 & 2 & NA \\ 
\citet{xuan2023can} &  Likert/Binary & 3,600 & 200 & C & 4 & 2 & 1,326 \\ 
\hline
\end{tabular}}
\end{table}

\subsection{B.2 Comparison with existing metrics}\label{appendix:XAIMetrics}

\paragraph{\textit{Faithfulness}:} 
\textit{How much does the explanation describe the true behavior of the model?} 
A number of different ways to compute \textit{faithfulness} exist, but they all broadly fit the same framework of measuring how much model predictions change in response to input perturbations \citep{bhatt2020faithcorr, alvarez2018towards, yeh2019fidelity, rieger2020irof, arya2019one, nguyen2020quantitative, bach2015pixel, samek2016evaluating, montavon2018methods, ancona2017towards, dasgupta2022framework}.
Intuitively, an explanation is faithful if perturbing regions deemed irrelevant by the explanation bring little to no change in model output, whereas perturbing regions deemed relevant bring a considerable change. 
In this analysis, we resort to the evaluation protocol outlined in \citet{azzolin2024perks}, which generalized a number of common \textit{faithfulness} metrics into a common mold\footnote{They focus on \textit{faithfulness} for graph explanations, but the evaluation protocol is aligned with that of images.}.
Specifically, \textit{faithfulness} is estimated as the harmonic mean of \textbf{sufficiency} \remirebutal{($\mathsf{Suf}$)} and \textbf{necessity} \remirebutal{($\mathsf{Nec}$)}, which account for the degree of prediction changes after perturbing irrelevant or relevant portions of the input, respectively. \remirebutal{Formally, given an input image $\vx$ with associated explanation $\ve$, and a model to be explained $p_\theta(Y \mid \vx)$, sufficiency and necessity are defined as:}
\remirebutal{
\begin{align}
\label{eq:suf-nec}
    & \mathsf{Suf}_{d, p_R}(\vx, \ve) =
        \mathbb{E}_{\vx' \sim p_R} [
            d( p_\theta( \cdot \mid \vx) \ \| \ p_\theta( \cdot \mid \vx'))
        ]
    \\ \nonumber
    & \mathsf{Nec}_{d, p_C}(\vx, \ve) =
        \mathbb{E}_{\vx' \sim p_C} [
            d( p_\theta( \cdot \mid \vx) \ \| \ p_\theta( \cdot \mid \vx'))
        ],
\end{align}
}
\remirebutal{where $d$ is a divergence between distributions of choice, and $p_C$ and $p_R$ are interventional distributions specifying the set of allowed perturbations to the explanation and its complement, respectively.}
\remirebutal{Eq. \ref{eq:suf-nec} are then normalised to $[0, 1]$, the higher the better, via a non-linear transformation, i.e., taking $\exp( - \mathsf{Suf}_{d,p_R}(\vx, \ve))$ and $1 - \exp(-\mathsf{Nec}_{d,p_C}(\vx, \ve))$.}
\remirebutal{Operationally, for a given instance ($\vx$, $\ve$) sampling from $p_C(\vx, \ve)$ equals to generating a new image where the complement of the explanation is left intact, and where perturbations are applied to the explanation.}
\remirebutal{The set of allowed perturbations $p_C$ and $p_R$ can be arbitrarily defined}, and different techniques are oftentimes reported to give different interpretations \citep{Hase2021oodproblem, rong2022consistent}. To avoid this confounding effect, we report the results for three different baseline perturbations, namely uniform and Gaussian noise, and black patches, along with a more advanced information-theoretic strategy named ROAD \citep{rong2022consistent}. 
Since explanations are oftentimes in the form of soft relevance scores over the entire input, a threshold is needed to tell apart relevant from irrelevant image regions. To avoid relying upon this hard-to-define hyperparameter, we aggregate the scores across multiple thresholds keeping only the best value.
Therefore, for each explanation threshold value, pixels are sorted based on their relevance\footnote{For sufficiency, pixels are sorted in ascending order. For necessity, in descending order.} and progressively perturbed until reaching the fixed threshold value, while leaving the others unchanged. 
\remirebutal{For each of those samples, we evaluate the normalized Eq. \ref{eq:suf-nec} where $d$ is the absolute difference in class-predicted confidence between clean and perturbed images, i.e., $| p_\theta(\hat{y} \mid \vx) - p_\theta(\hat{y} \mid \vx') |$, and average across the number of perturbed pixel for each threshold value. This procedure is detailed in Algorithm \ref{alg:faith}.
\remiaaai{Considering the novelty of our human driven study and the existence of such metrics, an interesting experiment is to measure if the the rating given by humans correlate with ROAD. Results, shown in 
Table~\ref{tab:faith_correlation} indicates a rather weak correlation.
We conclude that our human scores indeed cover an aspect of explanation quality unrelated to that of perceptual quality, as predicted by~\citet{biessmann2021quality}. The results cover only image-level attribution methods (see Table \ref{table:tested_methods}), as CBMs do not support such kinds of input-level manipulations.}
\remirebutal{Our findings reinforce the idea that human evaluations and computational metrics measure complementary aspects of XAI methods. Human evaluations excel at assessing the usefulness of explanations, aligning with their primary purpose of serving a human audience. In contrast, computational metrics, such as faithfulness, focus on evaluating the alignment between the explanation and the model's actual internal functioning. This aspect lies beyond the reach of human judgment, as humans cannot directly access or fully comprehend the internal mechanisms of the model.}
Evaluating the quality of an explanation typically involves estimating different and potentially orthogonal aspects of it. In addition to the perceptual quality addressed in this work, others can be numerically simulated by having access to model weights. In this additional analysis, we consider some of those aspects and measure how much they correlate with human scores. 
\begin{table}
    \centering
    \footnotesize
    \caption{\remiaaai{\textbf{Pearson Correlation Coefficient (PCC) and Spearman rank Correlation Coefficient (SCC) between \textit{faithfulness} computed with different methods and human scores.} Values are presented with p-values in parentheses.}}
    {
    \setlength{\tabcolsep}{4pt}
    \remiaaai{
        \begin{tabular}{@{}lrrrrrrrr@{}}
            \toprule
              & \multicolumn{2}{c}{ROAD} & \multicolumn{2}{c}{Black patches} & \multicolumn{2}{c}{Uniform noise} & \multicolumn{2}{c}{Gaussian noise}\\
             \cmidrule(lr){2-3} \cmidrule(lr){4-5} \cmidrule(lr){6-7} \cmidrule(lr){8-9}
             & \multicolumn{1}{c}{PCC} & \multicolumn{1}{c}{SCC} & \multicolumn{1}{c}{PCC} & \multicolumn{1}{c}{SCC} & \multicolumn{1}{c}{PCC} & \multicolumn{1}{c}{SCC} & \multicolumn{1}{c}{PCC} & \multicolumn{1}{c}{SCC} \\
             \midrule
\textbf{Q1} &  0.029 (1e-4)  & -0.047 (8e-10)  & 0.111 (1e-47)  & 0.077 (1e-23)  & 0.037 (1e-6)  & -0.036 (2e-6)  & 0.036 (1e-6)  & -0.032 (3e-5)  \\
\textbf{Q2} &  0.030 (1e-4)  & -0.044 (1e-9)  & 0.108 (3e-45)  & 0.075 (2e-22)  & 0.034 (9e-6)  & -0.037 (1e-6)  & 0.034 (1e-5)  & -0.032 (3e-5)  \\
\textbf{Q3} &  0.035 (6e-6)  & 0.003 (7e-1)  & 0.079 (3e-25)  & 0.065 (2e-17)  & 0.029 (2e-4)  & 0.005 (5e-1)  & 0.027 (1e-4)  & 0.004 (6e-1)  \\
\textbf{Q4} &  0.033 (1e-5)  & -0.008 (3e-1)  & 0.081 (3e-26)  & 0.062 (1e-15)  & 0.030 (9e-5)  & -0.001 (9e-1)  & 0.029 (2e-4)  & -0.001 (9e-1)  \\
\textbf{Q5} &  -0.060 (8e-15) & 0.003 (7e-1)   & -0.067 (3e-18) & -0.028 (1e-4)  & -0.066 (6e-18) & -0.025 (1e-3) & -0.065 (3e-17) & -0.022 (4e-3)  \\
\textbf{Q6} &  -0.019 (1e-2) & 0.059 (1e-14)   & -0.053 (4e-12) & -0.031 (6e-5) & -0.023 (1e-3) & 0.029 (1e-4) & -0.022 (5e-3) & 0.030 (1e-4)  \\
            \bottomrule
        \end{tabular}
    }}
\label{tab:faith_correlation}
\end{table}

\begin{algorithm}
\caption{\remirebutal{Pseudo code for computing sufficiency/necessity}}
\label{alg:faith}
    \begin{algorithmic}[1]
        \begingroup
            \REQUIRE Image $\vx$, explanation $\ve$, and set of explanation-size thresholds $\mathcal{T}$.
        
            \STATE $\text{values}  = [ ]$
            \FOR{each threshold $t \in \mathcal{T}$}
        
                \IF{computing sufficiency}
                    \STATE Sort pixels of $\vx$ in \textbf{ascending} order of relevance scores from $\ve$.
                \ELSE
                    \STATE Sort pixels of $\vx$ in \textbf{descending} order of relevance scores from $\ve$.
                \ENDIF
                
                \STATE $\text{arr} \leftarrow [ ]$
                \FOR{$i$ in range(start=1, end=$t$, step=2)}
                    \STATE $x' \leftarrow $ Apply the specified perturbations to the first $i \%$ sorted pixels.
                    \STATE Append $d = \left| p_\theta(\hat{y} \mid \mathbf{x}) - p_\theta(\hat{y} \mid \mathbf{x}') \right|$ to $\text{arr}$
                \ENDFOR

                \IF{computing sufficiency}
                    \STATE Append $\exp( - \text{mean}(\text{arr}))$ to values
                \ELSE
                    \STATE Append $1 - \exp(- \text{mean}(\text{arr}))$ to values
                \ENDIF
            \ENDFOR
        
            \STATE \textbf{Output:} $\max(\text{arr})$
        \endgroup
    \end{algorithmic}
\end{algorithm}

\paragraph{\textit{Robustness}:} 
\textit{Robustness} roughly refers to how stable the explanation is to small input perturbations. Different ways to estimate it exist \citep{alvarezmelis2018robustnessinterpretabilitymethods, montavon2018methods, yeh2019fidelity, dasgupta2022framework, agarwal2022rethinking}. In our analysis, we focused on MaxSensitivity \citep{yeh2019fidelity}, which applies random input perturbations to the entire image and measures the pixel-wise difference between the original explanation, and the one obtained on the perturbed sample. \remirebutal{Formally:}
\remirebutal{
\begin{equation}
    \text{MaxSensitivity} = \max \lVert \ve - \ve' \lVert
\end{equation}
}
where $\ve$ and $\ve'$ are the explanations for the original and the perturbed image, respectively.
Again, different perturbation techniques can be applied, and we resort to the two simple baselines, namely Uniform and Gaussian noise.
No normalization is applied, therefore the values are the higher the worse.
More advanced techniques like ROAD \citep{rong2022consistent} cannot be applied in this context, since the perturbation is applied uniformly over the entire image.
In Table~\ref{tab:maxsensitivity_correlation}, we report the correlation between MaxSensitivity and human scores, outlining a non-significant correlation with the metric and some questions. Surprisingly, the most correlated questions are \textbf{Q1-4}, which are not requesting humans to assess the stability of the explanation, something instead partially addressed by \textbf{Q5} and \textbf{Q6}. However, the correlation is very weak anyway, questioning any further claims.

\begin{table}[t]
    \centering
    \footnotesize
    \caption{\textbf{Pearson Correlation Coefficient (PCC) and Spearman rank Correlation Coefficient (SCC) between MaxSensitivity computed with different perturbation strategies and human scores.} In parentheses, the respective p-values.}
        \remiaaai{\begin{tabular}{lrrrr}
            \toprule
             & \multicolumn{2}{c}{Uniform noise} & \multicolumn{2}{c}{Gaussian noise}\\
             \cmidrule(lr){2-3} \cmidrule(lr){4-5}
             & \multicolumn{1}{c}{PCC} & \multicolumn{1}{c}{SCC} & \multicolumn{1}{c}{PCC} & \multicolumn{1}{c}{SCC} \\
             \midrule
\textbf{Q1} & -0.262 (9e-178)  & -0.326 (4e-280)  & -0.250 (4e-165)  & -0.301 (7e-243) \\
\textbf{Q2} & -0.258 (4e-172)  & -0.323 (4e-273)  & -0.247 (2e-161)  & -0.297 (7e-236) \\
\textbf{Q3} & -0.164 (7e-69)  & -0.193 (3e-95)  & -0.175 (1e-80)  & -0.188 (3e-93) \\
\textbf{Q4} & -0.176 (6e-80)  & -0.201 (2e-103)  & -0.182 (8e-88)  & -0.190 (3e-95) \\
\textbf{Q5} & 0.136 (9e-48)   & 0.169 (1e-73)   & 0.104 (2e-29)   & 0.103 (6e-29) \\
\textbf{Q6} & 0.101 (5e-27)   & 0.125 (5e-41)   & 0.100 (3e-27)   & 0.132 (1e-46) \\
            \bottomrule
        \end{tabular}}
    \label{tab:maxsensitivity_correlation}
\end{table}

\paragraph{\textit{Complexity}:} 
As humans have an implicit tendency to favor simple alternatives when facing a comparison between different hypotheses, providing simple and compact explanations is vital for human-machine synergy \citep{cowan2001magical}. 
Alternative methods for estimating the \textit{complexity} of an explanation are available, from simple above-threshold counting to more advanced information-theoretic techniques \citep{chalasani2020concise, bhatt2020faithcorr, nguyen2020quantitative}.
To test whether those metrics are correlated to human scores, we report in Table \ref{tab_sparseness_correlation} the correlation between human votes and Sparseness \citep{chalasani2020concise}, which estimates explanation \textit{complexity} as the Gini Index \citep{hurley2009comparing} of the absolute values of the image attribution.
The result is a metric value in the range $[0,1]$, where higher values indicate more sparseness.
\remirebutal{The computation of the Gini index is detailed in Algorithm \ref{alg:gini}.}

\begin{algorithm}
    \caption{\remirebutal{Pseudo code for Gini coefficient calculation from \citet{hedstrom2023quantus}}}
    \label{alg:gini}
    \begin{algorithmic}[1]
        \begingroup
            \REQUIRE Explanation $\ve$
            \STATE $array \leftarrow flatten(\ve)$
            \STATE $array \leftarrow |array|$ \COMMENT{Take absolute value}
            \STATE $array \leftarrow sort(array, ascending=True)$
            \STATE $index \leftarrow arange(1, array.shape[0] + 1)$
            \STATE $n \leftarrow array.shape[0]$ 
            \STATE return $\frac{\sum (2 \cdot index - n - 1) \cdot array}{n \cdot \sum array}$
        \endgroup
    \end{algorithmic}
\end{algorithm}

\begin{table}[t]
    \centering
    \caption{\textbf{Pearson Correlation Coefficient (PCC) and Spearman rank Correlation Coefficient (SCC) between Sparseness and human scores.} In parentheses the respective p-values.}
    \footnotesize
        \remiaaai{\begin{tabular}{lcc}
            \toprule
             & \multicolumn{2}{c}{\textit{Complexity}} \\
             \cmidrule(lr){2-3}
             & PCC & SCC\\
             \midrule
\textbf{Q1} & -0.137 (4e-72)  & -0.134 (3e-69)\\
\textbf{Q2} & -0.134 (6e-69)  & -0.131 (5e-66)\\
\textbf{Q3} & -0.080 (3e-25)  & -0.075 (9e-23)\\
\textbf{Q4} & -0.079 (6e-25)  & -0.073 (2e-21)\\
\textbf{Q5} & -0.085 (1e-28)  & -0.087 (6e-30)\\
\textbf{Q6} & 0.052 (9e-12)   & 0.073 (9e-22)\\  
            \bottomrule
        \end{tabular}}
    \label{tab_sparseness_correlation}
\end{table}

We used the Quantus library \citep{hedstrom2023quantus} for implementing the previous metrics, and we present the raw metric values in Table \ref{table:raw-metric-values}, aggregated by explainer and model.
Overall, none of the above metrics exhibits a significant correlation with user scores.

\begin{table}[t]
\centering
\caption{\textbf{Raw metric values averaged for each explainer and model.} Each value is the average result on 5 runs with the standard deviation. }
\resizebox{\textwidth}{!}{%
\begin{tabular}{@{}llccc@{}}
\toprule
\textbf{Saliency Method} & \textbf{Model} & \textbf{Faithfulness (ROAD)} & \textbf{Robustness (Gaussian noise)} & \textbf{Sparseness} \\ 
\midrule
GradCAM  & resnet50 & 								0.12 $\pm$ 0.20 & 0.92 $\pm$ 0.35 & 0.65 $\pm$ 0.11 \\
GradCAM  & vitB & 								0.02 $\pm$ 0.03 & 1.65 $\pm$ 1.77 & 0.50 $\pm$ 0.26 \\
AblationCAM  & vitB & 								0.02 $\pm$ 0.03 & 1.51 $\pm$ 1.19 & 0.74 $\pm$ 0.21 \\
AblationCAM  & CLIP-zero-shot & 								0.02 $\pm$ 0.02 & 1.39 $\pm$ 0.70 & 0.79 $\pm$ 0.16 \\
EigenCAM  & resnet50 & 								0.12 $\pm$ 0.21 & 0.93 $\pm$ 0.47 & 0.79 $\pm$ 0.07 \\
EigenCAM  & vitB & 								0.02 $\pm$ 0.03 & 1.03 $\pm$ 0.33 & 0.58 $\pm$ 0.06 \\
EigenCAM  & CLIP-zero-shot & 								0.02 $\pm$ 0.02 & 0.78 $\pm$ 0.38 & 0.56 $\pm$ 0.09 \\
EigenGradCAM  & resnet50 & 								0.12 $\pm$ 0.20 & 0.96 $\pm$ 0.57 & 0.78 $\pm$ 0.07 \\
EigenGradCAM  & vitB & 								0.02 $\pm$ 0.03 & 2.40 $\pm$ 1.74 & 0.83 $\pm$ 0.19 \\
EigenGradCAM  & CLIP-zero-shot & 								0.02 $\pm$ 0.03 & 1.26 $\pm$ 0.61 & 0.72 $\pm$ 0.14 \\
FullGrad  & resnet50 & 								0.11 $\pm$ 0.19 & 0.48 $\pm$ 0.15 & 0.40 $\pm$ 0.06 \\
FullGrad  & vitB & 								0.02 $\pm$ 0.02 & 1.08 $\pm$ 0.48 & 0.38 $\pm$ 0.06 \\
GradCAM  & CLIP-zero-shot & 								0.02 $\pm$ 0.02 & 1.21 $\pm$ 0.66 & 0.57 $\pm$ 0.17 \\
GradCAMPlusPlus  & resnet50 & 								0.12 $\pm$ 0.20 & 0.71 $\pm$ 0.26 & 0.60 $\pm$ 0.10 \\
GradCAMPlusPlus  & vitB & 								0.02 $\pm$ 0.02 & 2.33 $\pm$ 2.17 & 0.59 $\pm$ 0.28 \\
GradCAMPlusPlus  & CLIP-zero-shot & 								0.02 $\pm$ 0.02 & 1.46 $\pm$ 0.93 & 0.63 $\pm$ 0.24 \\
GradCAMElementWise  & resnet50 & 								0.11 $\pm$ 0.20 & 0.72 $\pm$ 0.25 & 0.58 $\pm$ 0.10 \\
GradCAMElementWise  & vitB & 								0.02 $\pm$ 0.02 & 1.20 $\pm$ 0.40 & 0.59 $\pm$ 0.09 \\
GradCAMElementWise  & CLIP-zero-shot & 								0.02 $\pm$ 0.02 & 0.71 $\pm$ 0.28 & 0.38 $\pm$ 0.08 \\
HiResCAM  & resnet50 & 								0.12 $\pm$ 0.20 & 0.91 $\pm$ 0.35 & 0.65 $\pm$ 0.11 \\
HiResCAM  & vitB & 								0.02 $\pm$ 0.03 & 1.92 $\pm$ 1.13 & 0.71 $\pm$ 0.24 \\
HiResCAM  & CLIP-zero-shot & 								0.02 $\pm$ 0.02 & 1.43 $\pm$ 0.49 & 0.69 $\pm$ 0.13 \\
LIME  & resnet50 & 								0.15 $\pm$ 0.22 & 0.49 $\pm$ 0.06 & 0.12 $\pm$ 0.04 \\
ScoreCAM  & resnet50 & 								0.12 $\pm$ 0.21 & 0.78 $\pm$ 0.31 & 0.59 $\pm$ 0.10 \\
ScoreCAM  & vitB & 								0.02 $\pm$ 0.03 & 1.22 $\pm$ 0.65 & 0.53 $\pm$ 0.18 \\
XGradCAM  & resnet50 & 								0.12 $\pm$ 0.20 & 0.88 $\pm$ 0.34 & 0.65 $\pm$ 0.11 \\
XGradCAM  & vitB & 								0.02 $\pm$ 0.03 & 1.37 $\pm$ 0.30 & 0.62 $\pm$ 0.07 \\
XGradCAM  & CLIP-zero-shot & 								0.02 $\pm$ 0.02 & 1.30 $\pm$ 0.20 & 0.59 $\pm$ 0.06 \\
DeepFeatureFactorization  & resnet50 & 								0.11 $\pm$ 0.18 & 0.95 $\pm$ 0.52 & 0.34 $\pm$ 0.12 \\
DeepFeatureFactorization  & vitB & 								0.02 $\pm$ 0.03 & 0.63 $\pm$ 0.23 & 0.28 $\pm$ 0.05 \\
DeepFeatureFactorization  & CLIP-zero-shot & 								0.02 $\pm$ 0.03 & 0.65 $\pm$ 0.31 & 0.33 $\pm$ 0.08 \\
LIME  & vitB & 								0.02 $\pm$ 0.03 & 0.50 $\pm$ 0.05 & 0.15 $\pm$ 0.03 \\
BCos  & resnet50-bcos & 								0.15 $\pm$ 0.24 & 0.88 $\pm$ 0.31 & 0.50 $\pm$ 0.09 \\
LIME  & CLIP-zero-shot & 								0.02 $\pm$ 0.03 & 0.58 $\pm$ 0.09 & 0.16 $\pm$ 0.05 \\
SHAP  & resnet50 & 								0.13 $\pm$ 0.21 & 1.37 $\pm$ 0.46 & 0.50 $\pm$ 0.10 \\
SHAP  & vitB & 								0.02 $\pm$ 0.03 & 1.05 $\pm$ 0.45 & 0.40 $\pm$ 0.09 \\
SHAP  & CLIP-zero-shot & 								0.02 $\pm$ 0.03 & 1.19 $\pm$ 0.38 & 0.46 $\pm$ 0.10 \\
AblationCAM  & resnet50 & 								0.12 $\pm$ 0.19 & 1.44 $\pm$ 0.48 & 0.71 $\pm$ 0.11 \\
\bottomrule
\end{tabular}%
}
\label{table:raw-metric-values}
\end{table}

\subsection{B.3 Dataset analysis}\label{appendix:Datasetanalysis}

\begin{figure}[t]
\begin{center}
\includegraphics[width=0.85\linewidth]{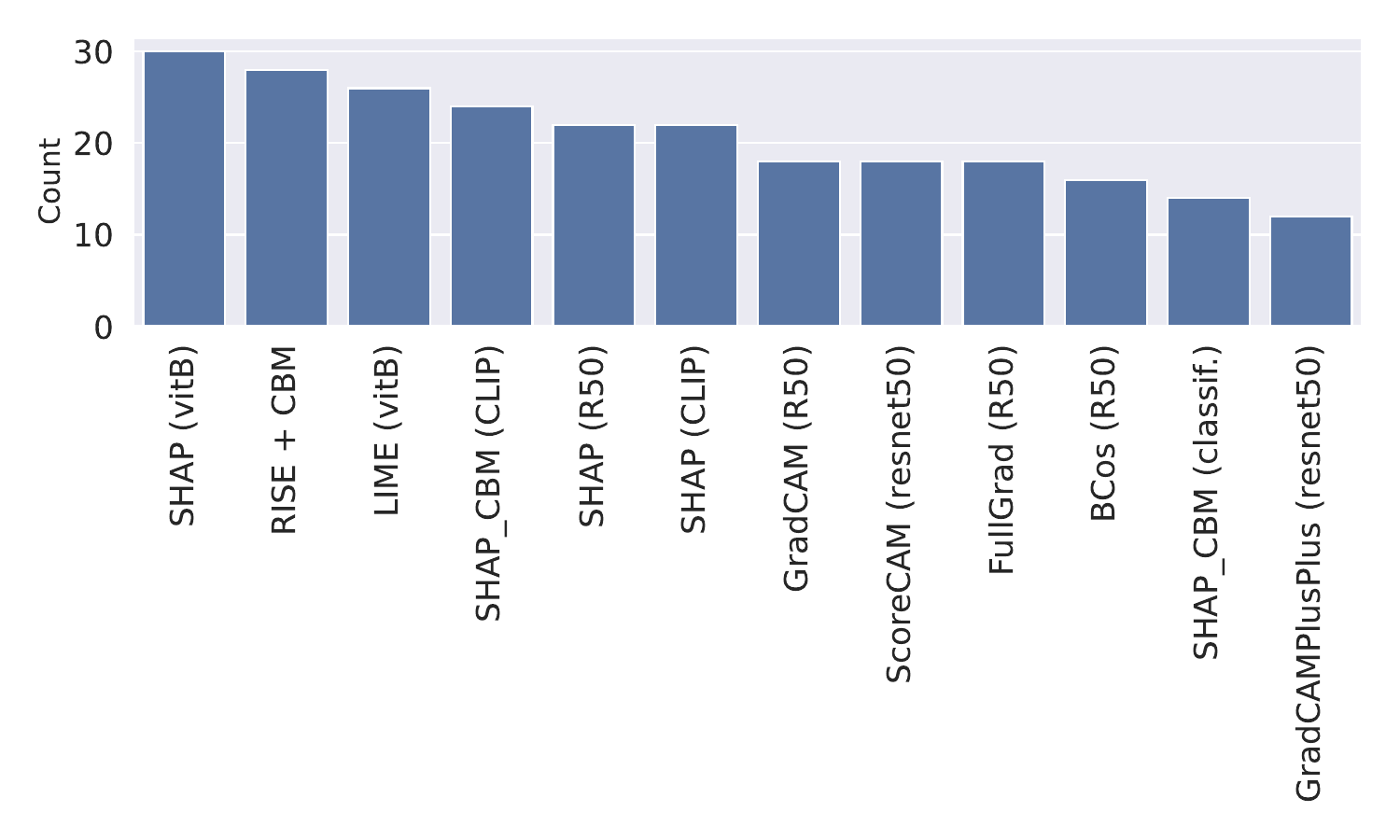} 
\end{center}
\caption{ \textbf{Histogram showing the top-12 XAI techniques preferred by each annotator.}}
\label{fig:top_xai}
\end{figure}
To further explore annotator preferences, we identified the top-12 XAI techniques selected by each annotator and visualized the results in the histogram shown in Figure \ref{fig:top_xai}. Unlike Table \ref{table:resultxai_annot}, this analysis focuses on the top-12 techniques per annotator, removing the influence of votes among the top-12 techniques to reduce noise and better capture annotator preferences.  From this figure, we observe that classical methods such as LIME and SHAP stand out as \remiaaai{one of} the most frequently preferred. This suggests a strong preference for well-established saliency-based methods. \remi{Additionally, there is a notable inclination towards methods that probe the model's reaction to input perturbations. A distinct aspect is the inclusion of B-cos, which generates explanations through the incorporation of a dedicated layer, offering a unique mechanism compared to other perturbation-based techniques.} Of the 11 most popular techniques among annotators, only \remiaaai{two} are based on CBMs, indicating a general preference for saliency maps over concept-based explanations. 

Next, we aggregated the scores using majority voting and calculated \remi{QWC scores} to measure the agreement between individual annotators and the aggregated score. We further analyzed the \remi{QWC scores} by gender and age groups to assess any systematic differences in interpretation. As shown in Figure \ref{fig:kappa_sex_age}, the kappa scores indicate that there is generally consistent agreement across different age and sex groups, although older annotators show slightly less consistency. This highlights that while demographic factors may introduce some variation, they do not substantially impact the overall interpretability evaluation. \remiaaai{Notable exceptions are observed for Q3 and Q4, which address the notions of objectivity and clarity. The decrease in agreement can be attributed to the increased subjectivity required by these questions compared to others.}

\begin{figure}[ht]
\begin{center}
    \begin{tabular}{c c}
\includegraphics[width=0.45\linewidth]{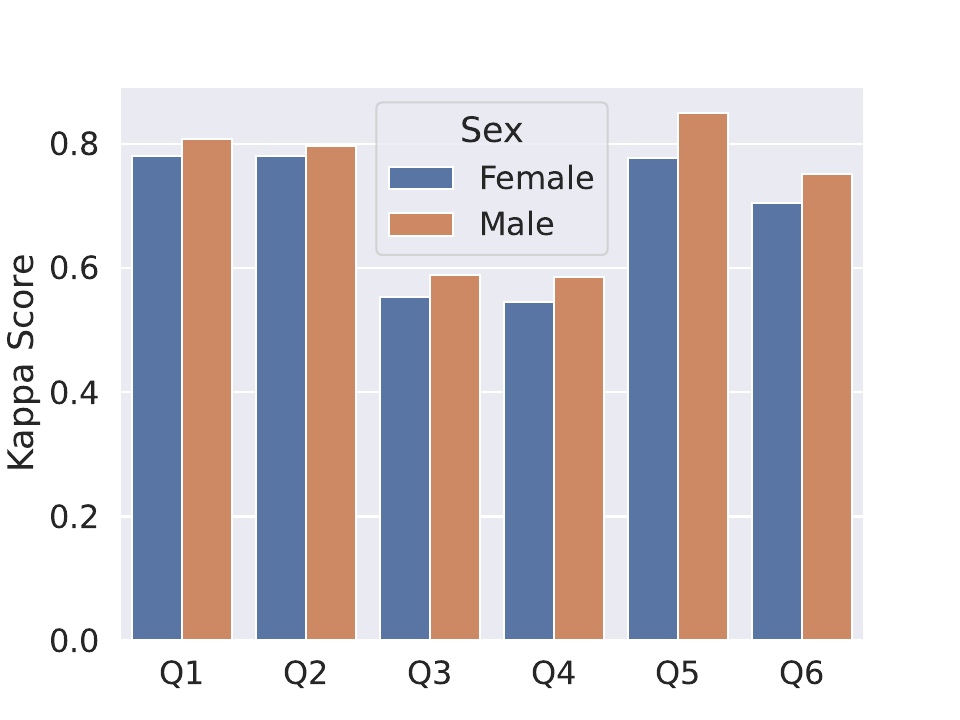} &\includegraphics[width=0.45\linewidth]{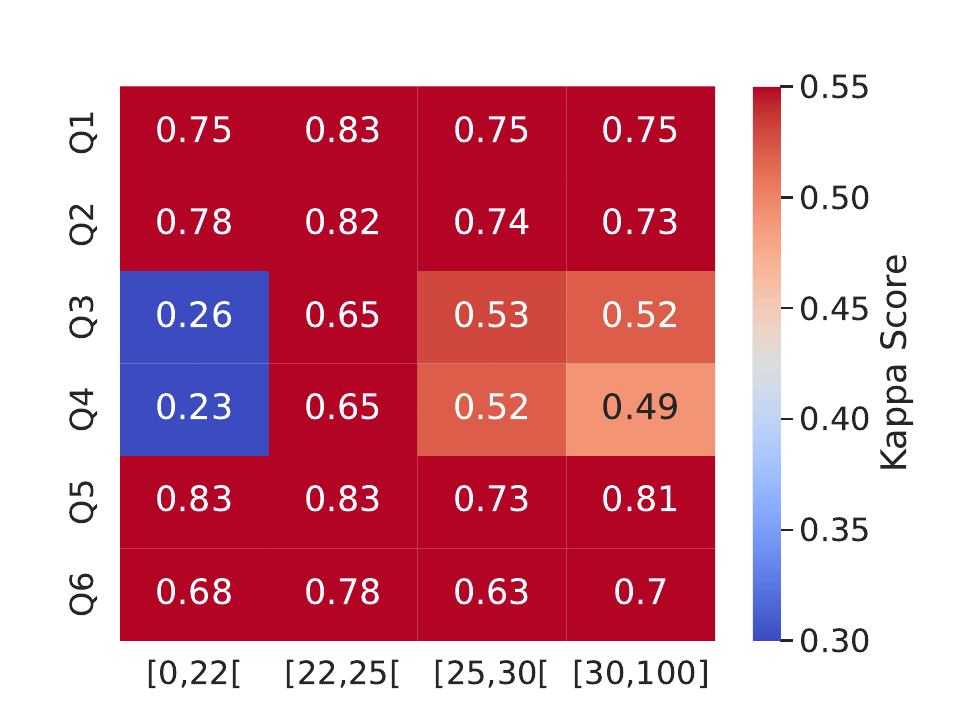} \\
    \end{tabular}
\end{center}
\caption{ \textbf{Cohen's kappa statistics showing agreement between annotators, aggregated by sex and age groups.} }
\label{fig:kappa_sex_age}
\end{figure}

\remiaaai{In Table \ref{table:resultxai_annot}, we present the average mode of votes for each XAI technique. The primary observation is the challenge in discerning clear differences among XAI methods, largely attributable to their high sensitivity to the specific backbones to which they are applied. Regarding the differences among questions, we note significant similarities between Q1 and Q2, as well as between Q3 and Q4. It is particularly noteworthy that raters globally assessed explanations as clearer and more objective, with mean scores of 3.46 for Q3 and 3.44 for Q4, compared to their faithfulness to the model, which had mean scores of 2.83 for Q1 and 2.82 for Q2. Beyond confirming the multifaceted nature of human assessment, this underscores the difficulties encountered by current XAI methods in conveying the behavior of the model, rather than merely creating clear methods for displaying explanations.}

\begin{table}[htbp]
\centering
\caption{\textbf{XAI techniques with aggregated scores across different evaluation metrics.}}
\label{table:resultxai_annot}
{
\remiaaai{\begin{tabular}{lrrrrrr}
\toprule
\textbf{XAI Technique} & \textbf{Q1} & \textbf{Q2} & \textbf{Q3} & \textbf{Q4} & \textbf{Q5} & \textbf{Q6} \\
\midrule
GradCAM (ResNet50) & 3.22 & 3.22 & 3.67 & 3.66 & 2.22 & 4.65 \\
GradCAM (ViT-B) & 2.63 & 2.64 & 3.41 & 3.41 & 2.53 & 4.62 \\
GradCAM (CLIP-zero-shot) & 2.40 & 2.40 & 3.30 & 3.29 & 2.63 & 4.79 \\
LIME (ResNet50) & 2.51 & 2.49 & 3.21 & 3.20 & 3.78 & 4.51 \\
LIME (ViT-B) & 3.00 & 2.98 & 3.55 & 3.53 & 3.23 & 4.53 \\
LIME (CLIP-zero-shot) & 2.75 & 2.74 & 3.35 & 3.33 & 3.61 & 4.53 \\
SHAP (ResNet50) & 3.00 & 2.99 & 3.50 & 3.49 & 2.64 & 4.70 \\
SHAP (ViT-B) & 3.10 & 3.09 & 3.58 & 3.55 & 2.57 & 4.70 \\
SHAP (CLIP-zero-shot) & 3.14 & 3.13 & 3.55 & 3.52 & 2.64 & 4.68 \\
AblationCAM (ResNet50) & 2.71 & 2.70 & 3.46 & 3.46 & 3.07 & 4.17 \\
AblationCAM (ViT-B) & 2.29 & 2.28 & 3.21 & 3.21 & 2.48 & 4.66 \\
AblationCAM (CLIP-zero-shot) & 2.03 & 2.03 & 2.99 & 3.00 & 2.34 & 4.64 \\
EigenCAM (ResNet50) & 3.12 & 3.11 & 3.65 & 3.63 & 2.26 & 4.25 \\
EigenCAM (ViT-B) & 3.13 & 3.13 & 3.65 & 3.63 & 2.27 & 4.29 \\
EigenCAM (CLIP-zero-shot) & 3.27 & 3.28 & 3.81 & 3.76 & 1.98 & 4.11 \\
EigenGradCAM (ResNet50) & 3.09 & 3.09 & 3.61 & 3.60 & 2.13 & 4.32 \\
EigenGradCAM (ViT-B) & 2.00 & 2.00 & 2.90 & 2.90 & 2.98 & 4.69 \\
EigenGradCAM (CLIP-zero-shot) & 2.73 & 2.73 & 3.45 & 3.43 & 2.62 & 4.58 \\
FullGrad (ResNet50) & 3.50 & 3.49 & 3.86 & 3.80 & 1.98 & 4.03 \\
FullGrad (ViT-B) & 2.28 & 2.27 & 3.16 & 3.16 & 3.13 & 4.64 \\
GradCAMPlusPlus (ResNet50) & 3.26 & 3.26 & 3.73 & 3.69 & 2.20 & 4.38 \\
GradCAMPlusPlus (ViT-B) & 2.58 & 2.58 & 3.33 & 3.31 & 3.10 & 4.69 \\
GradCAMPlusPlus (CLIP-zero-shot) & 2.31 & 2.31 & 3.27 & 3.26 & 2.78 & 4.79 \\
GradCAMElementWise (ResNet50) & 3.23 & 3.22 & 3.69 & 3.67 & 2.19 & 4.32 \\
GradCAMElementWise (ViT-B) & 2.12 & 2.12 & 2.95 & 2.97 & 2.69 & 4.65 \\
GradCAMElementWise (CLIP-zero-shot) & 3.10 & 3.10 & 3.64 & 3.61 & 2.25 & 4.37 \\
HiResCAM (ResNet50) & 3.28 & 3.27 & 3.72 & 3.69 & 2.21 & 4.60 \\
HiResCAM (ViT-B) & 2.03 & 2.03 & 2.98 & 2.98 & 2.93 & 4.80 \\
HiResCAM (CLIP-zero-shot) & 2.38 & 2.38 & 3.21 & 3.21 & 2.73 & 4.80 \\
ScoreCAM (ResNet50) & 3.32 & 3.32 & 3.73 & 3.70 & 2.32 & 4.50 \\
ScoreCAM (ViT-B) & 3.19 & 3.18 & 3.73 & 3.68 & 2.44 & 4.66 \\
XGradCAM (ResNet50) & 3.26 & 3.27 & 3.71 & 3.69 & 2.10 & 4.63 \\
XGradCAM (ViT-B) & 2.37 & 2.37 & 3.29 & 3.28 & 3.78 & 4.65 \\
XGradCAM (CLIP-zero-shot) & 2.59 & 2.59 & 3.46 & 3.45 & 3.88 & 4.70 \\
DeepFeatureFactorization (ResNet50) & 3.41 & 3.41 & 3.86 & 3.82 & 2.22 & 4.18 \\
DeepFeatureFactorization (ViT-B) & 3.03 & 3.02 & 3.72 & 3.68 & 2.18 & 4.26 \\
DeepFeatureFactorization (CLIP-zero-shot) & 3.22 & 3.22 & 3.78 & 3.72 & 2.02 & 4.07 \\
CLIP-QDA-sample & 2.20 & 2.19 & 3.11 & 3.09 & 2.72 & 4.70 \\
CLIP-Linear-sample & 3.10 & 3.08 & 3.48 & 3.45 & 1.68 & 4.46 \\
LIME\_CBM (CLIP-QDA) & 2.69 & 2.69 & 3.38 & 3.36 & 3.64 & 4.59 \\
SHAP\_CBM (CLIP-QDA) & 3.09 & 3.08 & 3.53 & 3.52 & 2.82 & 4.54 \\
LIME\_CBM (CBM-classifier-logistic) & 2.90 & 2.88 & 3.38 & 3.35 & 2.77 & 4.49 \\
SHAP\_CBM (CBM-classifier-logistic) & 3.08 & 3.06 & 3.46 & 3.43 & 2.43 & 4.30 \\
Xnesyl-Linear & 2.36 & 2.34 & 3.22 & 3.19 & 2.45 & 3.83 \\
BCos (ResNet50-BCos) & 2.86 & 2.85 & 3.49 & 3.46 & 2.30 & 3.84 \\
RISE-CBM (CBM-classifier-logistic) & 3.28 & 3.27 & 3.66 & 3.62 & 2.22 & 4.35 \\
\bottomrule
\end{tabular}}
}
\end{table}

\section{C. PASTA-dataset: Extra questions}

\remiaaai{To enhance the study, we solicited responses from a subset of annotators (15 annotators) to address additional questions. These questions did not directly involve generated explanations. Instead, the questions focused on their overall perceptions of XAI and how they personally approach explanations. Consequently, this section aims to take a broader perspective and reflect more comprehensively from a user study viewpoint.}

\subsection{C.1 Image labeling}

For each image, participants had to explain what led them to classify the displayed image as the model's prediction. Participants responded openly using a text form.

A first set of questions aims at having annotators establish a baseline, i.e., by interpreting and explaining what makes an image recognizable as a specific object or class. 
\remiaaai{The process is constituted of theses two questions:}
\begin{itemize}[leftmargin=1.25em]
     \itemsep0em 
     \item \remi{Q0.1}: What part makes you classify this image as \remi{***}? (write an explanation extracting concepts) %
     \item \remi{Q0.2}: What part of the input helps the prediction? \eloise{(draw bounding boxes on the image)} %
\end{itemize}

Similarly, they were asked to describe the elements of the image that helped them make the decision to classify the image as the model did. These two questions \remi{(Q0.1 and Q0.2)} are used to establish a baseline for interpreting what makes an image recognizable as a specific object or class, and what are the salient features of the images that would explain this choice. 

We investigated potential biases in the annotations themselves by examining the differences in how annotators approached CBM-based and saliency-based explanations. For CBM explanations, we focused on the text written by annotators in response to question Q0.1, assessing whether annotators preferred explanations that closely resembled their own textual responses. To quantify this, we transformed CBM explanations into text by concatenating the top concepts used in the explanation and calculated the BLEU \citep{papineni2002bleu} and ROUGE scores \citep{lin2004rouge} between these explanations and the annotators’ text responses. As shown in Figure \ref{fig:correlationxai}, the ROUGE score reveals a slight correlation between the explanations and the annotators’ expectations for questions Q1, Q2, Q3, and Q4. This suggests that annotators are inclined to favor explanations that align with their preconceived notions, potentially introducing a bias toward consistency with their initial answers.

Moreover, we observed that questions Q1, Q2, Q3, and Q4 exhibit high intercorrelation, as do questions Q5 and Q6. This clustering indicates that annotators tend to evaluate explanations similarly across these sets of questions, which may reflect underlying patterns in how different types of explanations are perceived.

For saliency-based explanations, we analyzed the bounding boxes provided by annotators in response to question Q0.2. We evaluated the correlation between various metrics and the annotators' answers, including: 
\begin{enumerate}
    \item the total sum of pixel intensities in the saliency map (``SUM\_all''),
    \item the sum of pixel intensities within the area of the image identified by the bounding box (``SUM\_pos''), 
    \item the sum of pixel intensities outside the bounding box (``SUM\_neg''),
    \item the entropy of the saliency map (``Entropy'').
\end{enumerate}

Figure \ref{fig:correlationxai} shows that questions Q1, Q2, Q3, and Q4 are highly correlated with each other, as are questions Q5 and Q6. 
Additionally, all metrics except for ``SUM\_pos'' show some correlation with questions Q1–Q4. This suggests that annotators may focus heavily on background features and salient objects when answering these questions, potentially overlooking finer details in the bounding box area.

Overall, these analyses highlight several potential biases in the dataset. Annotators exhibit a preference for certain types of explanations, particularly saliency maps, and tend to favor explanations that align with their expectations, as evidenced by the correlation between their text responses and the explanations. Additionally, while demographic factors such as age and gender do not significantly impact the overall evaluation, the slight decrease in consistency among older annotators warrants further investigation.  The study involved 15 annotators, all from the same cultural background, which may introduce some shared perspectives or biases. To mitigate this, future studies could benefit from a more diverse group of annotators.%

\begin{figure}[ht]
\begin{center}\footnotesize
    \begin{tabular}{c c}
\includegraphics[width=0.45\linewidth]{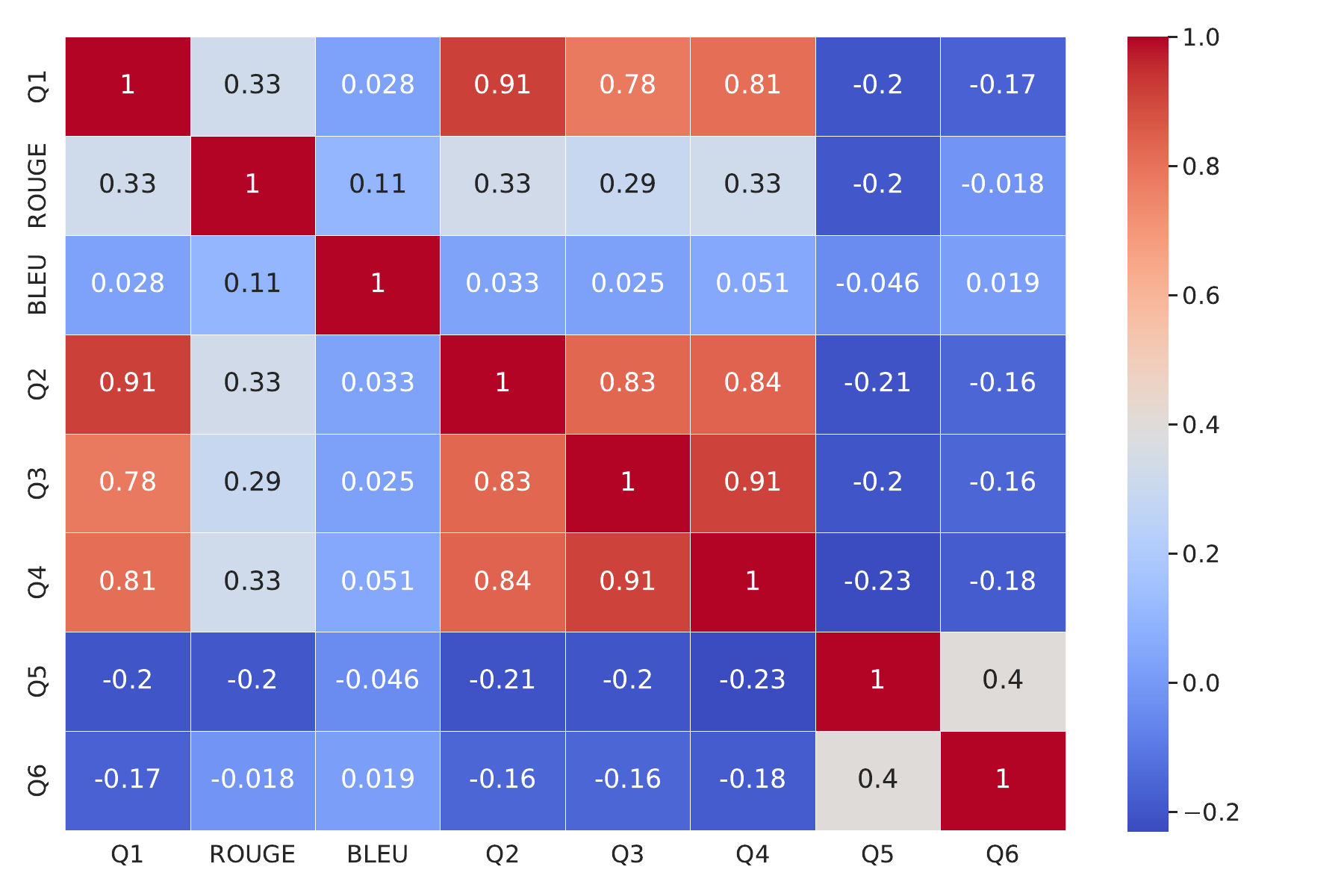} &\includegraphics[width=0.45\linewidth]{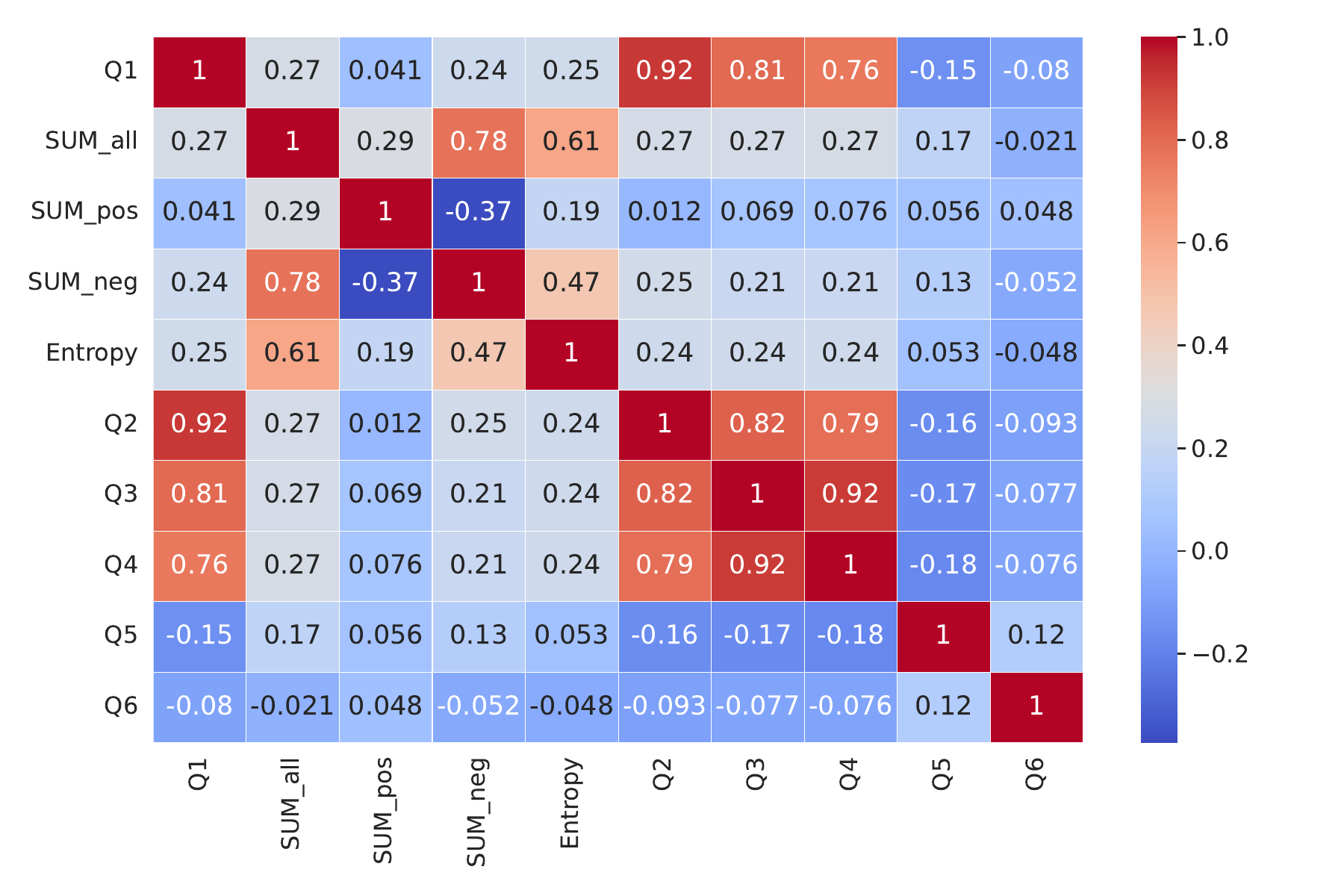} \\
      \textbf{CBM} & \textbf{Saliency map}\\
    \end{tabular}
\end{center}
\caption{ \textbf{Correlation between the various questions and key metrics for saliency and CBM explanations.} For CBM, the criteria used are the 
BLEU \citep{papineni2002bleu} and ROUGE scores \citep{lin2004rouge}
scores between the explanation and the text from question Q0.1. For saliency maps, the metrics include the total pixel sum (SUM\_all), the sum of pixels within the annotator-provided bounding box (SUM\_pos) from question Q0.2, the sum of pixels outside the bounding box (SUM\_neg), and the entropy of the saliency map (Entropy). }
\label{fig:correlationxai}
\end{figure}

\subsection{C.2 Desiratas ranking}

\remiicml{After the annotation process is completed, evaluators are presented with a set of questions assessing their stance on the questions asked.}

\remirebutal{In addition to annotating the samples that comprise the PASTA-dataset, each participant in the study was asked to respond to the following questions:}

\remirebutal{
\begin{itemize}
    \item Q7: Could you rank the different qualities of the explanation in order of importance? \textit{Expectedness}, \textit{Trustworthiness}, \textit{Complexity}, \textit{Robustness}, \textit{Objectivity}.
    \item Q8: Can you please order the different questions from Q1, Q2, Q3, Q4, Q5, and Q6 from the less important to the more important questions to assess the quality of the explanation?
    \item Q9: Can you please order the different questions from Q1, Q2, Q3, Q4, Q5 and Q6 from the most difficult to the easiest?
\end{itemize}}

\remirebutal{The results for Q7 are presented in Table \ref{table:resultsQ7}. The data indicate that evaluators place a higher value on Trustworthiness and Complexity, while Objectivity is ranked significantly lower. This finding aligns with the work of \citet{liao2022connecting}, who asked a similar question but focused on a pool of experts.}

\remirebutal{Table \ref{table:resultsQ8Q9} summarizes the results for Q8 and Q9. According to user feedback, Q1 is deemed the most important question. Notably, there is a strong correlation between the responses to Q8 and Q9: the questions perceived as easiest to answer are also regarded as the most important. Interestingly, there is a low correlation between the importance assigned to each question and the axioms they represent, highlighting a distinction between the perception of an axiom and the execution of the associated task. Furthermore, both Q8 and Q9 reveal a separation in ranking among Q1 to Q4 and Q5 and Q6, which resonates with the dataset analysis.}

\begin{table}[h!]
\centering
\remirebutal{
\caption{\remirebutal{\textbf{Average positions of axioms for question 7.} Lower rankings indicate that the axiom is considered more important by evaluators, while higher rankings suggest the axiom is considered less important. Results sorted by ascending order.}}
\label{table:resultsQ7}
\begin{tabular}{lc}
\toprule
\textit{Axiom} & \textit{Average Position (Q7)} $\downarrow$ \\ \midrule
\textit{Trustworthiness}  & \textbf{2.47} $\pm$ 1.48 \\
\textit{Complexity}   & 2.53 $\pm$ 1.58 \\
\textit{Robustness}   & 2.97 $\pm$ 1.06 \\
\textit{Expectedness} & 3.20 $\pm$ 1.05 \\
\textit{Objectivity}  & 3.77 $\pm$ 1.36 \\ \bottomrule\end{tabular}}
\end{table}

\begin{table}[h!]
\centering
\remirebutal{
\caption{\remirebutal{\textbf{Average positions of questions 1 to 6 for Q8 and Q9.} For Q8, higher rankings indicate that the question is considered more important by evaluators, while lower rankings suggest the question is considered less important. For Q9, lower rankings indicate that the question is considered more difficult to evaluate, while lower rankings suggest the question is considered less difficult.}}
\label{table:resultsQ8Q9}
\begin{tabular}{lcc}
\toprule
\textit{Question} & \textit{Average Position (Q8) $\uparrow$} & \textit{Average Position (Q9) $\downarrow$} \\ \midrule
Q1 & \textbf{4.10} $\pm$ 1.32 & 4.00 $\pm$ 1.21 \\
Q2 & 3.90 $\pm$ 1.29 & 4.23 $\pm$ 1.35 \\
Q3 & 4.03 $\pm$ 1.77 & 4.53 $\pm$ 1.70 \\
Q4 & 3.13 $\pm$ 1.41 & 3.67 $\pm$ 1.40 \\
Q5 & 2.63 $\pm$ 1.66 & \textbf{1.80} $\pm$ 0.57 \\
Q6 & 3.00 $\pm$ 1.95 & 2.54 $\pm$ 1.72 \\ \bottomrule
\end{tabular}}
\end{table}

\section{D. PASTA-score: Implementation details}

\subsection{D.1 Technical information}

\paragraph{Hyperparameters}

\remiaaai{In all experiments, we employed the Adam optimizer \citep{kingma2017adam} with a batch size of 256, training for 600 epochs at a learning rate of $2\times10^{-6}$. Additionally, we configured the parameters as follows: $\alpha$ was set to 1, $\beta$ to $0.001$, and $\gamma$ to $0.01$. We also applied a weight decay of $1\times10^{-6}$. The model utilized a hidden layer size of [512, 64]. For concept-based explanations, we used the top 15 concepts, with no template. For Q3 to Q6, we excluded a part of the annotations related to COCO, due to a presence of outliers.}

\remiaaai{Regarding the division between the training, validation, and test sets, we imposed restrictions on the selected samples to ensure no overlap in XAI methods or images on which the explanations are based, thereby preventing overfitting. To achieve this, we proceeded as follows. First, we assigned an ID to each of the images ($\text{IMG\_ID}$) and each of the XAI methods ($\text{XAI\_ID}$). As mentioned in the dataset specifications, $\text{IMG\_ID} \in \llbracket 1~;~1000 \rrbracket$ and $\text{XAI\_ID} \in \llbracket 1~;~46\rrbracket$. Secondly, given these sets and a seed, we randomly selected 70\% of the $\text{XAI\_ID}$ and 70\% of the $\text{IMG\_ID}$. Thirdly, we included in the training set only those samples that presented both a valid $\text{XAI\_ID}$ and $\text{IMG\_ID}$ ($\neg \left( (\text{XAI\_ID\_train} = \text{XAI\_ID\_test}) \lor (\text{IMG\_ID\_train} = \text{IMG\_ID\_test}) \right)$). We followed the same procedure with the remaining set to construct the validation and test sets. All remaining samples were discarded for the run.}

\subsubsection{\remi{Evaluation metrics}} \label{appendix:eval_metrics}

\remi{The quadratic weighted Cohen's Kappa (QWK) measures inter-rater agreement, adjusting for chance and penalizing disagreement based on its magnitude. The formula is:}
\remi{\begin{equation}
    \text{QWK} = \frac{\sum_{i,j} w_{ij} O_{ij} - \sum_{i,j} w_{ij} E_{ij}}{1 - \sum_{i,j} w_{ij} E_{ij}} ,
\end{equation}}%
\remi{where:
\begin{itemize}
    \item \(O_{ij}\) and \(E_{ij}\) are the observed and expected frequencies, respectively.
    \item \(w_{ij} = 1 - \frac{(i - j)^2}{(k - 1)^2}\) is the quadratic weight for categories \(i\) and \(j\).
\end{itemize}}

\remi{The Mean Squared Error (MSE) measures the average squared difference between predicted and actual values. It is given by:}
\remi{\begin{equation}
    \text{MSE} = \frac{1}{N_s} \sum_{k=1}^{N_s} (m_k - \hat{m}_k)^2  \, .  
\end{equation}}

\remi{The Spearman Correlation Coefficient (SCC) measures the rank correlation between two variables. It is calculated using the ranks of the data points and is given by:}
\remi{
\begin{equation}
    \text{SCC} = 1 - \frac{6 \sum_{i} d_i^2}{N_s(N_s^2 - 1)}  , 
\end{equation}}
\remi{where \(d_i\) is the difference between the ranks of each pair of values.
}

\paragraph{Loss Functions} \label{appendix:loss_functions}

Here, we define the three losses used to train our PASTA-model.

The Cosine Similarity Loss measures the cosine similarity between the predicted explanations \remiaaai{scores} $\hat{m}_k$ and the ground truth explanations \remiaaai{scores} $m_k$, ensuring alignment in their direction:

\begin{equation}
    L_s = 1 - \frac{\sum_{k=1}^{N_s} \hat{m}_k m_k}{\sqrt{\sum_{k=1}^{N_s} \hat{m}_k^2} \sqrt{\sum_{k=1}^{N_s} m_k^2}}
\end{equation}

The Mean Squared Error (MSE) loss measures the squared difference between predicted and true explanations, penalizing larger errors more heavily:

\begin{equation}
    L_{mse} = \frac{1}{N_s} \sum_{k=1}^{N_s} (\hat{m}_k - m_k)^2
\end{equation}

This Ranking Loss ensures the correct ranking of explanations by penalizing cases where the predicted ranking contradicts the true ranking:

\begin{equation}
    L_r = \frac{1}{\hat{N_s}} \sum_{k_1,k_2} \max(0, -(\hat{m}_{k_1} - \hat{m}_{k_2})(m_{k_1} - m_{k_2}))
\end{equation}

Here, $\hat{N_s}$ represents the total number of pairs considered for the ranking loss.

\subsubsection{\remiicml{Computational resources}} \label{appendix:computtime}

\remi{The experiments were conducted using a \remiaaai{V100-32GB GPU}. The training and inference times are summarized in Table \ref{table:time_comput}. It is important to note that, for both inference and testing, the majority of the computational time is dedicated to precomputing VLM embeddings, as the scoring network itself is relatively lightweight and requires minimal computational resources.}
\begin{table}[t!]
\caption{\remi{\textbf{Training and Inference Times on GPU and CPU.} \remiaaai{Results for the $\text{PASTA-score}^{\text{SIGLIP}}$ variant.}}}
\label{table:time_comput}
\centering
\remiaaai{\begin{tabular}{lrr}
\toprule
\textit{Device} & \textit{Training Time (s)} & \textit{Inference Time (s)} \\ \midrule
GPU  & 1048.74 & 36.12  \\
CPU  & 10035.98 & 796.41 \\ \bottomrule
\end{tabular}}
\end{table}

\subsection{D.2 Ablation studies}

\paragraph{Aggregation of the votes} \label{appendix:Aggregation}

\subsubsection{\remiaaai{Formulas}} \label{formula_aggreg}

\remi{In the dataset, we have access to 5 votes per question. Then, if we denote the set of votes as $\{m_i^j[a]\}_{a=1}^{N_a}$, where $N_a=5$ is the number of annotations: }
\remi{
\begin{equation}
    \text{Mode}(m_i^j) = \underset{a}{\arg\max} \; \text{Count}(m_i^j[a]) \, ,
\end{equation}
where \( \text{Count}(m_i^j[a]) \) represents the frequency of each vote \(a\) in the set. 
} %

\remi{Given the subjective nature of the annotations and the presence of multiple responses to the same question (five answers per question), we explored different methods for determining the ground truth. In Table \ref{ablation_gt_comput}, we tested how PASTA-score training is affected when using the mean, mode, or median as the ground truth.
Since this parameter significantly impacts the dispersion of the samples, it is not surprising that the results vary, particularly when using the mean. However, in all our experiments, we opted to use the mode, as phenomena of high non-consensus were observed.}

\begin{table}[t!]
\centering
\caption{\textbf{Comparison of the influence of ground truth generation methods.} Each value is the average result on 5 runs with the standard deviation.}
\label{ablation_gt_comput}
\remiaaai{
\begin{tabular}{lccc}
\toprule
\textit{Label Type} & \textit{MSE} & \textit{QWK} & \textit{SCC} \\ \midrule
Mode   & 0.989 $\pm$ 0.113 & 0.471 $\pm$ 0.055 & 0.501 $\pm$ 0.052 \\
Mean   & \textbf{0.774 $\pm$ 0.067} & \textbf{0.506 $\pm$ 0.049} & \textbf{0.532 $\pm$ 0.043} \\
Median & 0.924 $\pm$ 0.093 & 0.479 $\pm$ 0.052 & 0.509 $\pm$ 0.046 \\
\bottomrule
\end{tabular}}
\end{table}

\paragraph{\remirebutal{Add of label information in the PASTA-score embedding.}}

\remirebutal{In the current iteration of the PASTA-score, information regarding the predicted class is integrated into the embeddings supplied to the scoring network. Specifically, each predicted label is encoded as a one-hot vector and concatenated with the embedding. In this study, we aim to measure the impact of this design choice by comparing it with an alternative approach that utilizes only embeddings as input to the scoring network.}

\remiaaai{As illustrated in Table~\ref{table_embeddings_labels}, the integration of label information exhibits a positive influence, aligning with the hypothesis that such contextual data enhances the comprehension of human behavior patterns affecting ratings. However, the observed improvement is modest. This phenomenon can be attributed to several factors, including the extensive number of labels employed across the datasets (26), which is commensurate with the quantity of unique images. As a result, the inclusion of label information may introduce redundancy or lead to overfitting, thereby influencing the efficiency of the scoring mechanism.}

\begin{table}[t]
\centering
\caption{\remirebutal{\textbf{Impact of Adding Labels to the Encodings.} Each value is the average result on 5 runs with the standard deviation.}}
\label{table_embeddings_labels}
\remiaaai{
\begin{tabular}{lccc}
\toprule
\textit{Computation} & \textit{MSE} & \textit{QWK} & \textit{SCC} \\ \midrule
\textit{Embeddings} & \textbf{0.987 $\pm$ 0.098} & 0.469 $\pm$ 0.047 & 0.497 $\pm$ 0.045 \\
\textit{Embeddings + Labels} & 0.989 $\pm$ 0.113 & \textbf{0.471 $\pm$ 0.055} & \textbf{0.501 $\pm$ 0.052} \\
\bottomrule
\end{tabular}}
\end{table}

\paragraph{\remirebutal{Scoring functions}}

\remirebutal{In this section, we examine the influence of various scoring network architectures on performance. Specifically, we tested alternatives such as Ridge Regression, Lasso Regression, Support Vector Machines. The results of these experiments are presented in Table~\ref{table_scoring_function}.}

\remirebutal{By analyzing the performance of the different scoring functions, we observe that PASTA, implemented with \remiaaai{a multi linear perceptron} and leveraging the loss functions described in Section~\ref{scoring_net}, achieves superior results in terms of the Quadratic Weighted Kappa score and Spearman Correlation Coefficient. These outcomes highlight its effectiveness in accurately ranking labels. However, SVM exhibit lower Mean Square Error, suggesting better numerical precision in predicting label values. Our primary objective is to develop a robust metric for ranking XAI methods. As such, we place greater emphasis on metrics that assess ranking accuracy. Based on this criterion, the PASTA framework is favored over alternative scoring networks due to its superior performance in rank-oriented evaluations.}

\begin{table}[t]
\centering
\caption{\textbf{Impact of the Scoring Function.} Each value is the average result on 5 runs with the standard deviation.}
\label{table_scoring_function}
\remiaaai{
\begin{tabular}{lccc}
\toprule
\textit{Scoring Function} & \textit{MSE} & \textit{QWK} & \textit{SCC} \\ \midrule
\textit{PASTA} & 0.989 $\pm$ 0.113 & \textbf{0.471 $\pm$ 0.055} & \textbf{0.501 $\pm$ 0.052} \\
\textit{SVM} & \textbf{0.965 $\pm$ 0.077} & 0.463 $\pm$ 0.056 & \textbf{0.501 $\pm$ 0.051} \\
\textit{Ridge} & 1.169 $\pm$ 0.143 & 0.423 $\pm$ 0.046 & 0.444 $\pm$ 0.043 \\
\textit{Lasso} & 0.998 $\pm$ 0.075 & 0.395 $\pm$ 0.048 & 0.454 $\pm$ 0.052 \\
\bottomrule
\end{tabular}}
\end{table}

\subsection{\remi{Explanation embeddings}}

\remiaaai{This section presents several ablation studies that informed the development of our current methodology for embedding explanations. Throughout these experiments, the Siglip model was utilized as the VLM encoder.}

\paragraph{\remi{Saliency}} \label{appendix:embed_saliency}

\remi{Regarding saliency-based explanations, a key question arises about what should be considered as the image representing the explanation. Two variants were considered: using  the heatmap visualization that is presented to users, as shown in Equation~\ref{embed_saliency_bis}, or the input image as defined as: %
}
\remi{\begin{equation}
    \phi_{\text{image}}(\ve_i^j) = \text{VLM}_{\text{image}}(\vx_i \times \ve_i^j) 
    \label{embed_saliency}
\end{equation}}%
\remi{The element-wise multiplication of the input image with the saliency map selectively blurs the image, with regions corresponding to lower activation values being blurred, while areas with higher activation values remain clear.}

\remi{The results are presented in Table \ref{table_ablation_saliency}, where a slight improvement is observed in favor of using the image as a heatmap. This can be attributed to the fact that, despite being more computationally ambiguous, the heatmap display reveals the entire image. Additionally, this representation closely resembles the format of the samples provided to annotators.}

\begin{table}[t!]
\centering
\caption{\textbf{Influence of the saliency computation process.} \textit{Heatmap} refers to the process defined in Equation \ref{embed_saliency_bis} and \textit{Masked image} refers to the process defined in Equation \ref{embed_saliency}. Each value is the average result on 5 runs with the standard deviation.}
\label{table_ablation_saliency}
\remiaaai{
\begin{tabular}{lccc}
\toprule
\textit{Embedded Image} & \textit{MSE} & \textit{QWK} & \textit{SCC} \\ \midrule
\textit{Heatmap} & \textbf{0.989 $\pm$ 0.113} & \textbf{0.471 $\pm$ 0.055} & \textbf{0.501 $\pm$ 0.052} \\
\textit{Blur} & 0.992 $\pm$ 0.089 & 0.470 $\pm$ 0.048 & 0.498 $\pm$ 0.043 \\
\bottomrule
\end{tabular}}
\end{table}

\paragraph{\remi{CBM}} \label{appendix:embed_cbm}

\remi{Concerning concept bottleneck explanation, which basically can be interpreted as a dictionary attributing a scalar for each concept. One crucial step is converting CBM activations to VLM embeddings. We tested two ways do do so:}
\remi{\begin{itemize}
    \item By considering the raw text of concepts, ordered by importance (Equation \ref{embed_cbm})
    \item By using a sum of all the VLM embeddings of text, weightened by its activations:%
    \remi{
\begin{equation}
    \phi_{\text{text}}(\ve_i^j) = \frac{1}{|| \ve_i^j ||} \sum_k^{N_k} \ve_i^j[k] ~ \text{VLM}_{\text{text}}(\text{concept}_i[k])
    \label{embed_cbm_bis}
\end{equation}}
\end{itemize}}
If we use the first solution, there are questions about the number of concepts to keep, that we note as the parameters $N_{top}$. Table \ref{ablation_cbm_ntop} presents the influence of $N_{top}$ while \ref{ablation_cbm_computation} presents the influence of differents ways to compute the embeddings. \remiaaai{Analysis of the results indicates that an adequate number of concepts enhances ranking performance, which eventually plateaus, thereby informing the selection of $N_{top}=15$. Regarding the embedding procedure, optimal outcomes were achieved by encoding the top concepts as a sentence. The inclusion of a template demonstrated negligible influence on overall performance metrics.}

\begin{table}[t]
\centering
\caption{\textbf{Influence of the number of words selected as an input text $N_{top}$.} Each value is the average result on 5 runs with the standard deviation.}
\label{ablation_cbm_ntop}
\remiaaai{
\begin{tabular}{cccc}
\toprule
$N_{top}$ & \textit{MSE} & \textit{QWK} & \textit{SCC} \\ \midrule
5  & \textbf{0.988 $\pm$ 0.101} & 0.466 $\pm$ 0.047 & 0.495 $\pm$ 0.043 \\
10 & 0.992 $\pm$ 0.115 & 0.461 $\pm$ 0.066 & 0.493 $\pm$ 0.057 \\
15 & 0.989 $\pm$ 0.113 & \textbf{0.471 $\pm$ 0.055} & \textbf{0.501 $\pm$ 0.052} \\
20 & 0.994 $\pm$ 0.122 & 0.470 $\pm$ 0.059 & 0.498 $\pm$ 0.056 \\
\bottomrule
\end{tabular}}
\end{table}

\begin{table}[t]
\centering
\caption{\textbf{Influence of the CBM explanation embedding process.}
\protect\label{ablation_cbm_computation}
\textit{Weightened} refers to the process described in Equation \ref{embed_cbm_bis}, Sentence refers to the process described in Equation \ref{embed_cbm}, preceded with the template noted in \textit{Template}. Each value is the average result on 5 runs with the standard deviation.}
\remiaaai{
\begin{tabular}{lcccc}
\toprule
\textit{Computation} & \textit{MSE} & \textit{QWK} & \textit{SCC} & \textit{Template} \\ \midrule
\textit{Weighted} & 1.212 $\pm$ 0.288 & 0.392 $\pm$ 0.077 & 0.425 $\pm$ 0.082 & \thead{`` ''} \\ 
\textit{Sentence} & 0.989 $\pm$ 0.113 & \textbf{0.471 $\pm$ 0.055} & 0.501 $\pm$ 0.052 & \thead{`` ''} \\ 
\textit{Sentence} & \textbf{0.986 $\pm$ 0.119} & 0.469 $\pm$ 0.061 & \textbf{0.504 $\pm$ 0.055} & \thead{``Concepts in explanation:''} \\
\bottomrule
\end{tabular}}
\end{table}

\section{E. PASTA-score: Additional Examples}

\begin{table}[h]
\centering
\caption{\remiaaai{\textbf{PASTA scores corresponding to different explanations.} Dataset: catsdogscars. Label: dog}}
\label{transposed_table1}
\remiaaai{
\begin{tabular}{cccccccc}
\toprule
 & 
\includegraphics[width=2cm]{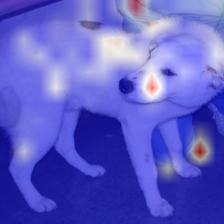} &
\includegraphics[width=2cm]{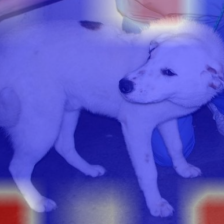} &
\includegraphics[width=2cm]{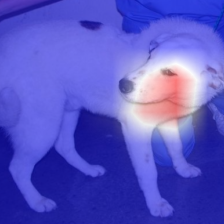} &
\includegraphics[width=2cm]{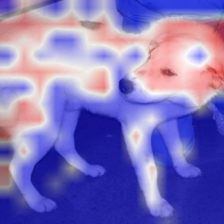} &
\includegraphics[width=2cm]{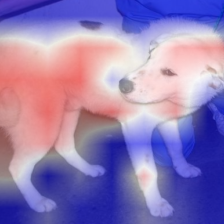} &
\includegraphics[width=2cm]{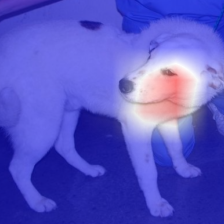} &
\includegraphics[width=2cm]{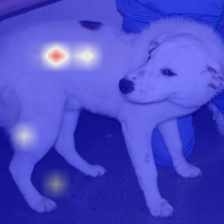} \\
\midrule
P (Q1)  & 2.01 & 2.04 & 3.94 & 3.01 & 3.81 & 3.97 & 2.14 \\
P (Q2)  & 1.90 & 1.82 & 3.31 & 2.80 & 3.57 & 3.37 & 1.70 \\
P (Q3) & 2.93 & 2.58 & 3.65 & 4.06 & 4.02 & 3.65 & 2.83 \\
P (Q4) & 3.04 & 2.95 & 3.84 & 4.09 & 4.14 & 3.84 & 2.75 \\
\bottomrule
\end{tabular}}
\end{table}

\begin{table}[h]
\centering
\caption{\remiaaai{\textbf{PASTA scores corresponding to different explanations.} Dataset: MonumAI. Label: Baroque}}
\label{transposed_table2}
\remiaaai{
\begin{tabular}{cccccccc}
\toprule
 &\includegraphics[width=2cm]{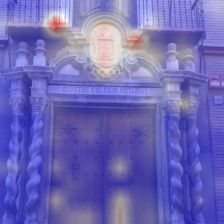} & \includegraphics[width=2cm]{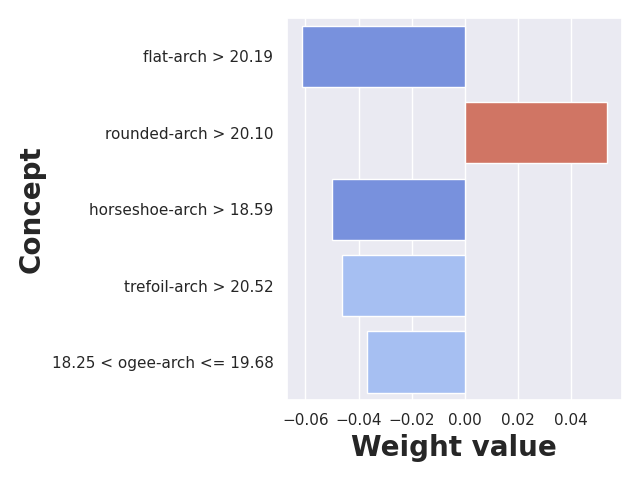} & \includegraphics[width=2cm]{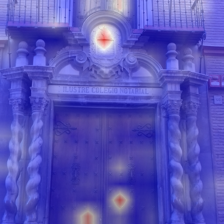} & \includegraphics[width=2cm]{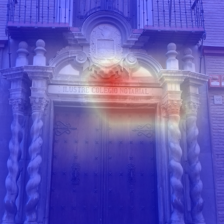} & \includegraphics[width=2cm]{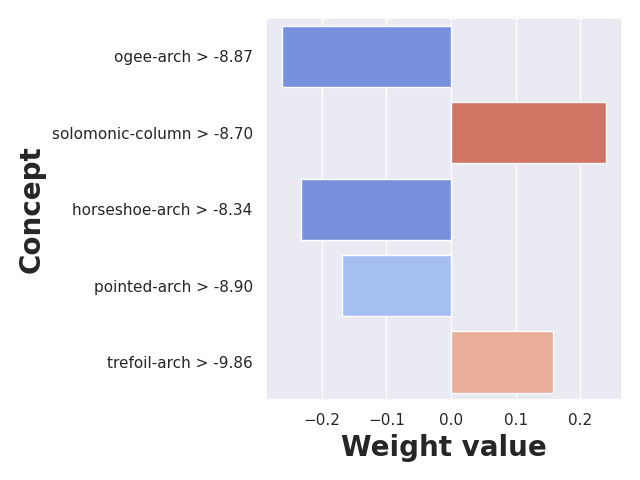} & \includegraphics[width=2cm]{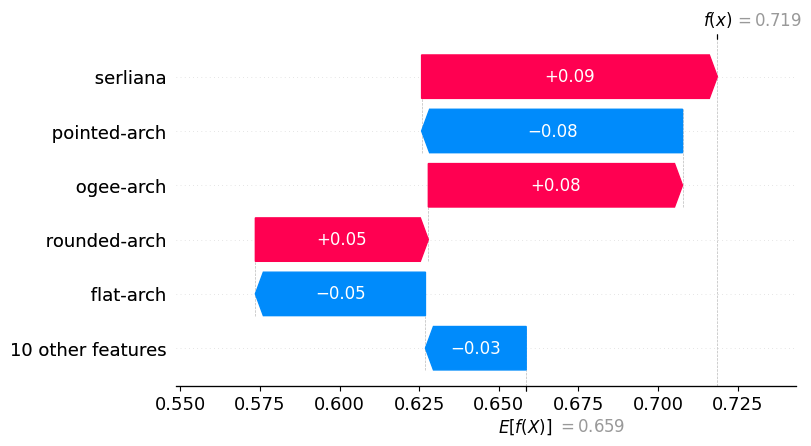} & \includegraphics[width=2cm]{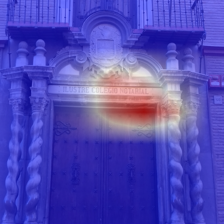}\\
\midrule
P (Q1) & 2.45 & 1.01 & 3.06 & 4.00 & 1.00 & 1.67 & 3.22 \\
P (Q2) & 2.51 & 1.26 & 2.99 & 3.48 & 1.57 & 1.52 & 3.51 \\
P (Q3) & 3.67 & 1.52 & 3.51 & 3.71 & 2.05 & 1.54 & 3.69 \\
P (Q4) & 3.35 & 1.33 & 3.45 & 3.89 & 1.33 & 1.59 & 3.79 \\
\bottomrule
\end{tabular}}
\end{table}

\begin{table}[h]
\centering
\caption{\remiaaai{\textbf{PASTA scores corresponding to different explanations.} Dataset: COCO. Label: transportation}}
\label{transposed_table_fixed_coco_enough}
\remiaaai{
\begin{tabular}{cccccccc}
\toprule
&\includegraphics[width=2cm]{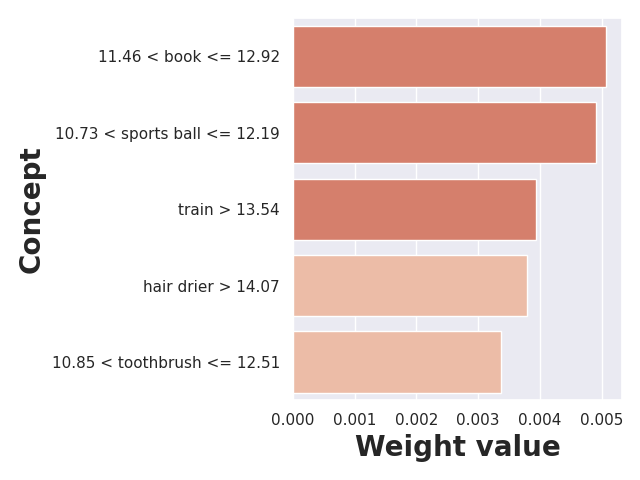} & \includegraphics[width=2cm]{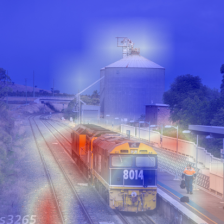} & \includegraphics[width=2cm]{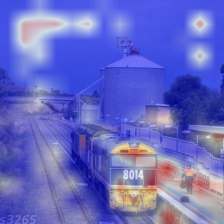} & \includegraphics[width=2cm]{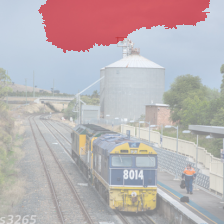} & \includegraphics[width=2cm]{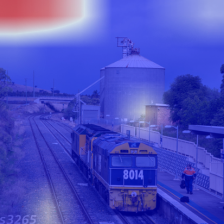} & \includegraphics[width=2cm]{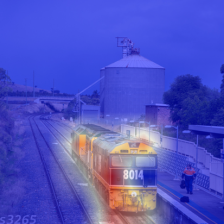} & \includegraphics[width=2cm]{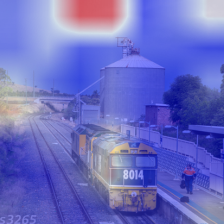}\\
\midrule
P (Q1) & 1.21 & 3.24 & 2.46 & 3.44 & 2.01 & 2.80 & 2.40 \\
P (Q2) & 1.61 & 3.22 & 2.45 & 3.36 & 2.23 & 2.75 & 2.37 \\
P (Q3) & 1.43 & 3.94 & 3.45 & 3.78 & 3.01 & 3.69 & 3.41 \\
P (Q4) & 1.59 & 3.65 & 3.20 & 3.50 & 2.96 & 3.25 & 3.28 \\
\bottomrule
\end{tabular}}
\end{table}

\begin{table}[h]
\centering
\caption{\remiaaai{\textbf{PASTA scores corresponding to different explanations.} Dataset: PascalPart. Label: cat}}
\label{transposed_table_pascalpart}
\remiaaai{
\begin{tabular}{cccccccc}
\toprule
&\includegraphics[width=2cm]{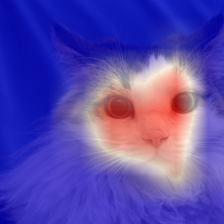} & \includegraphics[width=2cm]{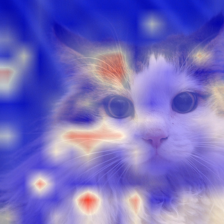} & \includegraphics[width=2cm]{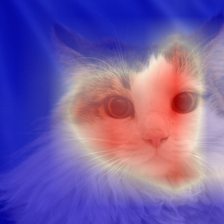} & \includegraphics[width=2cm]{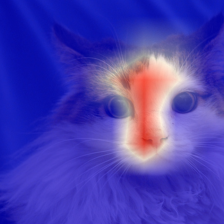} & \includegraphics[width=2cm]{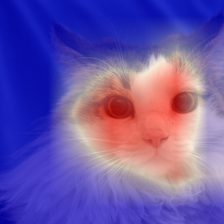} & \includegraphics[width=2cm]{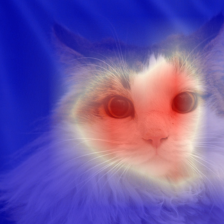} & \includegraphics[width=2cm]{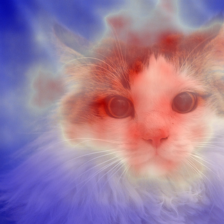}\\
\midrule
P (Q1) & 2.02 & 1.68 & 3.90 & 2.51 & 3.93 & 2.50 & 3.30 \\
P (Q2) & 2.53 & 1.71 & 3.91 & 2.55 & 3.90 & 2.60 & 3.42 \\
P (Q3) & 2.88 & 2.77 & 3.78 & 3.22 & 3.94 & 3.30 & 3.80 \\
P (Q4) & 3.06 & 2.82 & 3.78 & 3.27 & 3.82 & 3.26 & 3.82 \\
\bottomrule
\end{tabular}}
\end{table}

\begin{table}[t]
\centering
\caption{\remiaaai{\textbf{PASTA scores corresponding to different explanations.} Dataset: PascalPart. Label: person}}
\label{transposed_table_pascalpart_second}
\remiaaai{
\begin{tabular}{cccccccc}
\toprule
&\includegraphics[width=2cm]{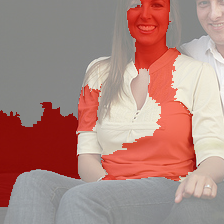} & \includegraphics[width=2cm]{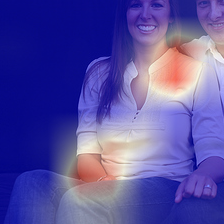} & \includegraphics[width=2cm]{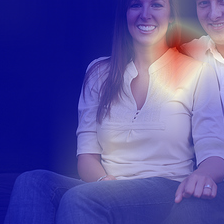} & \includegraphics[width=2cm]{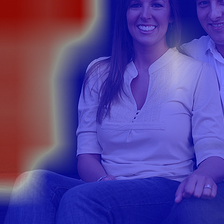} & \includegraphics[width=2cm]{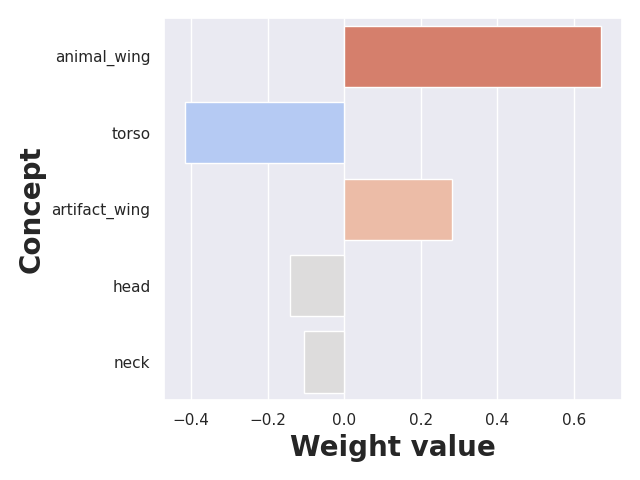} & \includegraphics[width=2cm]{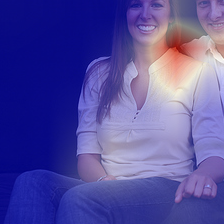} & \includegraphics[width=2cm]{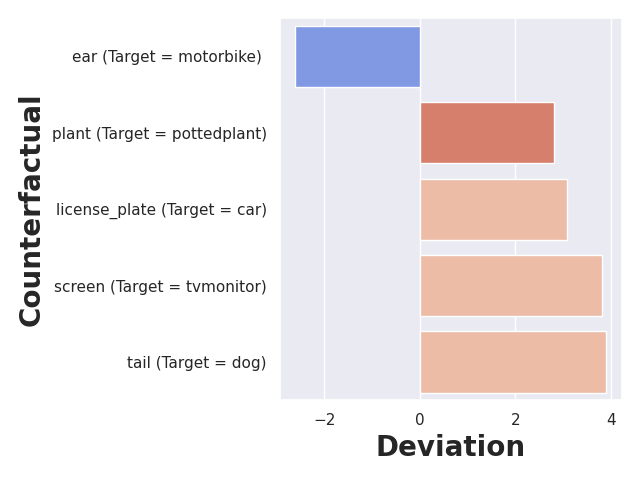}\\
\midrule
P (Q1) & 3.68 & 3.55 & 3.69 & 3.12 & 1.35 & 3.71 & 1.93 \\
P (Q2) & 3.46 & 3.19 & 3.16 & 3.07 & 1.74 & 3.16 & 1.81 \\
P (Q3) & 3.50 & 3.64 & 3.64 & 3.02 & 1.77 & 3.59 & 2.56 \\
P (Q4) & 4.23 & 4.09 & 4.05 & 3.82 & 1.29 & 4.04 & 1.91 \\
\bottomrule
\end{tabular}}
\end{table}

\begin{table}[t]
\centering
\caption{\remiaaai{\textbf{PASTA scores corresponding to different explanations.} Dataset: COCO. Label: cultural}}
\label{transposed_table_fixed_coco_second}
\remiaaai{
\begin{tabular}{cccccccc}
\toprule
&\includegraphics[width=2cm]{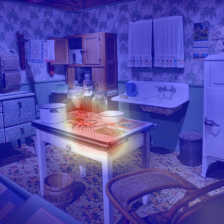} & \includegraphics[width=2cm]{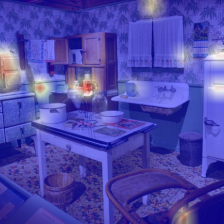} & \includegraphics[width=2cm]{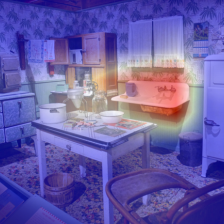} & \includegraphics[width=2cm]{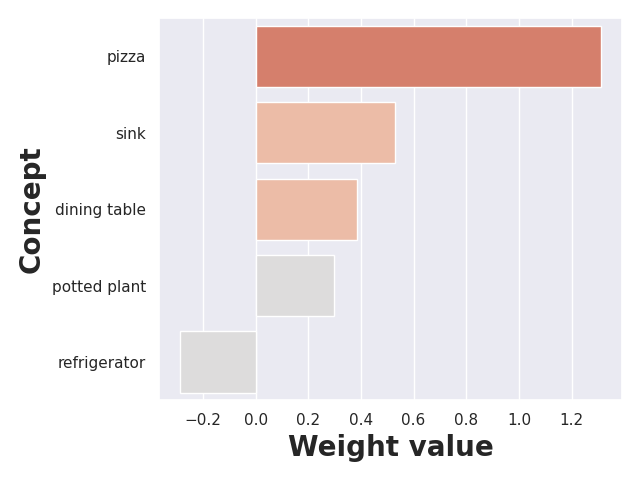} & \includegraphics[width=2cm]{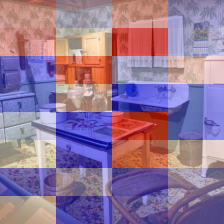} & \includegraphics[width=2cm]{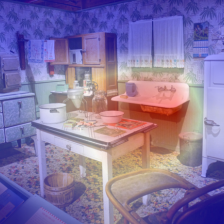} & \includegraphics[width=2cm]{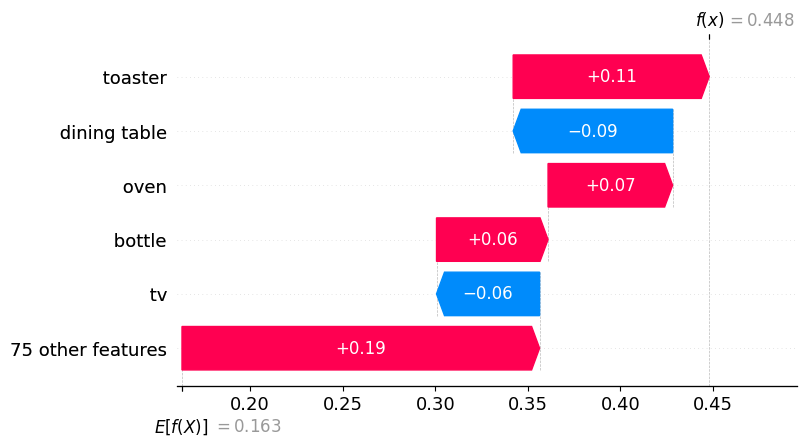}\\
\midrule
P (Q1) & 2.65 & 2.27 & 2.78 & 1.21 & 2.96 & 2.75 & 1.45 \\
P (Q2) & 2.21 & 1.73 & 2.31 & 1.64 & 2.82 & 2.14 & 1.83 \\
P (Q3) & 2.40 & 2.17 & 3.05 & 1.37 & 3.25 & 2.49 & 1.72 \\
P (Q4) & 2.49 & 2.20 & 2.84 & 1.30 & 3.06 & 2.75 & 1.92 \\
\bottomrule
\end{tabular}}
\end{table}

\begin{table}[t]
\centering
\caption{\remiaaai{\textbf{PASTA scores corresponding to different explanations.} Dataset: catsdogscars. Label: car}}
\label{transposed_table_catsdogscars_second}
\remiaaai{
\begin{tabular}{cccccccc}
\toprule
&\includegraphics[width=2cm]{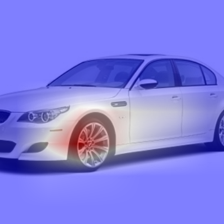} & \includegraphics[width=2cm]{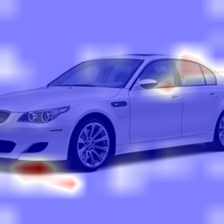} & \includegraphics[width=2cm]{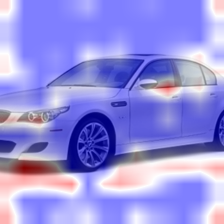} & \includegraphics[width=2cm]{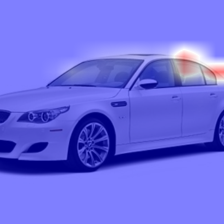} & \includegraphics[width=2cm]{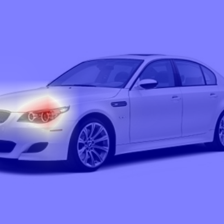} & \includegraphics[width=2cm]{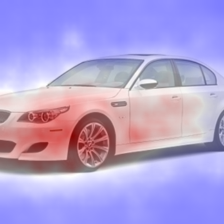} & \includegraphics[width=2cm]{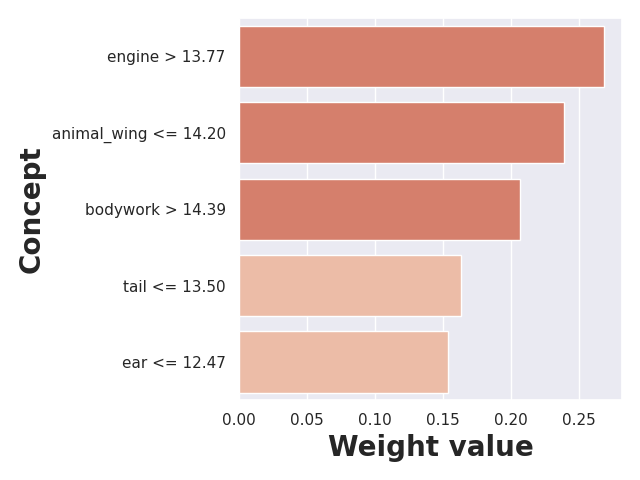}\\
\midrule
P (Q1) & 3.20 & 2.67 & 4.80 & 2.10 & 2.74 & 4.80 & 1.50 \\
P (Q2) & 3.18 & 2.60 & 4.69 & 1.53 & 2.37 & 4.69 & 1.64 \\
P (Q3) & 2.98 & 2.54 & 4.30 & 2.55 & 3.17 & 4.30 & 2.16 \\
P (Q4) & 3.34 & 2.47 & 4.06 & 2.51 & 3.01 & 4.06 & 1.72 \\
\bottomrule
\end{tabular}}
\end{table}

\begin{table}[t]
\centering
\caption{\remiaaai{\textbf{PASTA scores corresponding to different explanations.} Dataset: PascalPart. Label: motorbike}}
\label{transposed_table_pascalpart_second2}
\remiaaai{
\begin{tabular}{cccccccc}
\toprule
&\includegraphics[width=2cm]{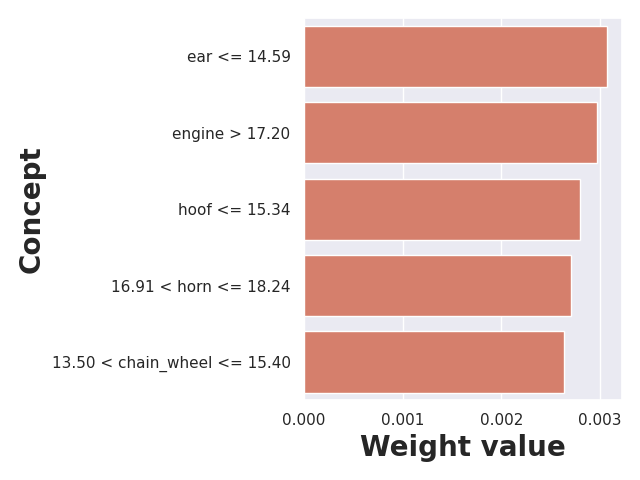} & \includegraphics[width=2cm]{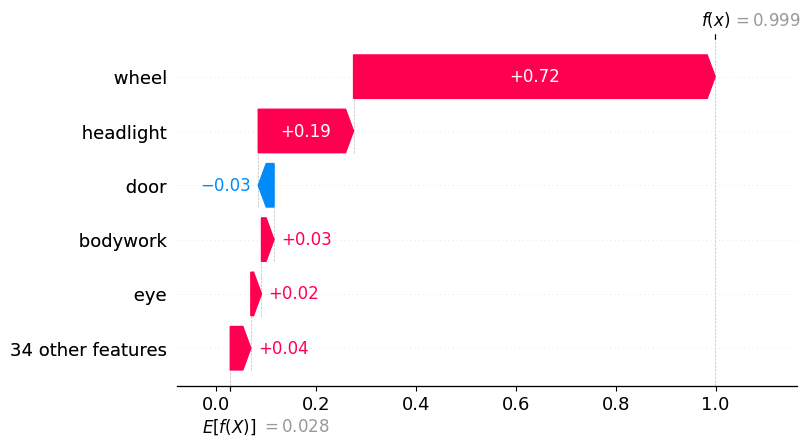} & \includegraphics[width=2cm]{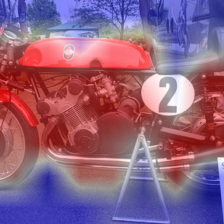} & \includegraphics[width=2cm]{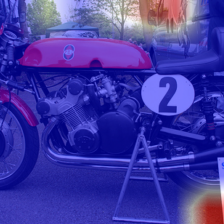} & \includegraphics[width=2cm]{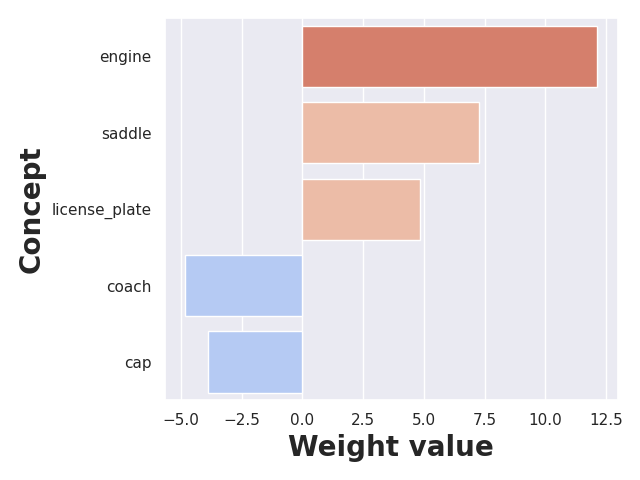} & \includegraphics[width=2cm]{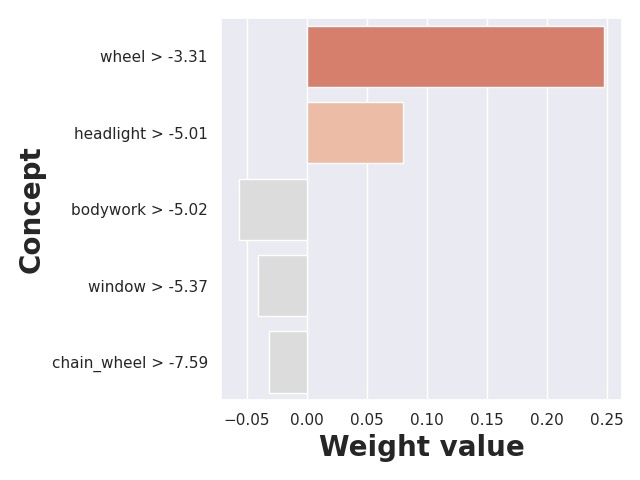} & \includegraphics[width=2cm]{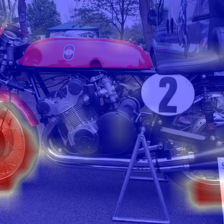}\\
\midrule
P (Q1) & 1.26 & 1.42 & 3.38 & 2.43 & 1.10 & 1.00 & 3.26 \\
P (Q2) & 1.59 & 1.61 & 3.35 & 2.22 & 1.46 & 1.52 & 3.07 \\
P (Q3) & 2.18 & 1.43 & 4.41 & 3.31 & 1.96 & 1.77 & 4.18 \\
P (Q4) & 2.15 & 2.10 & 4.22 & 3.47 & 1.45 & 2.24 & 4.08 \\
\bottomrule
\end{tabular}}
\end{table}

\begin{table}[t]
\centering
\caption{\remiaaai{\textbf{PASTA scores corresponding to different explanations.} Dataset: MonumAI. Label: Gothic}}
\label{transposed_table_pascalpart_second3}
\remiaaai{
\begin{tabular}{cccccccc}
\toprule
&\includegraphics[width=2cm]{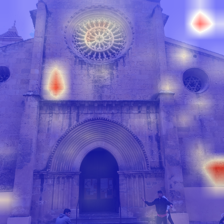} & \includegraphics[width=2cm]{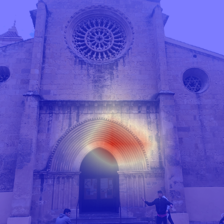} & \includegraphics[width=2cm]{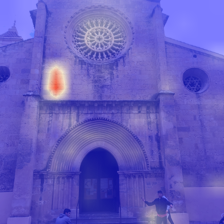} & \includegraphics[width=2cm]{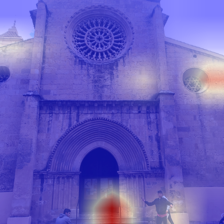} & \includegraphics[width=2cm]{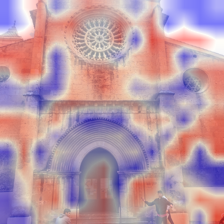} & \includegraphics[width=2cm]{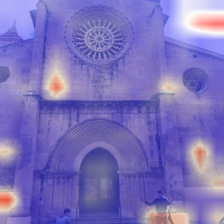} & \includegraphics[width=2cm]{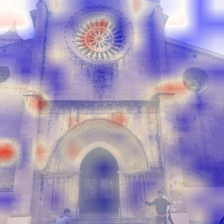}\\
\midrule
P (Q1) & 2.74 & 3.75 & 3.38 & 3.00 & 3.97 & 2.88 & 3.69 \\
P (Q2) & 2.51 & 3.43 & 3.13 & 2.86 & 3.91 & 2.79 & 3.44 \\
P (Q3) & 3.21 & 3.90 & 3.58 & 3.43 & 4.37 & 3.70 & 3.98 \\
P (Q4) & 3.18 & 3.85 & 3.51 & 3.38 & 4.25 & 3.52 & 3.84 \\
\bottomrule
\end{tabular}}
\end{table}

\end{document}